%% file: PaperForReview.tex
\documentclass[10pt,twocolumn,letterpaper]{article}

\usepackage[pagenumbers]{cvpr} % To force page numbers, e.g. for an arXiv version

\usepackage{graphicx}
\usepackage{amsmath}
\usepackage{amssymb}
\usepackage{booktabs}
\usepackage{multirow}
\usepackage{multicol}
\usepackage[title]{appendix}
\usepackage{xcolor}

\usepackage[pagebackref,breaklinks,colorlinks]{hyperref}

\usepackage[capitalize]{cleveref}
\crefname{section}{Sec.}{Secs.}
\Crefname{section}{Section}{Sections}
\Crefname{table}{Table}{Tables}
\crefname{table}{Tab.}{Tabs.}

\def\cvprPaperID{9398} % *** Enter the CVPR Paper ID here
\def\confName{CVPR}
\def\confYear{2022}

\begin{document}

\title{A Comprehensive Study of Image Classification Model Sensitivity to Foregrounds, Backgrounds, and Visual Attributes}

%\title{Assessing Model Sensitivity to Foregrounds, Backgrounds, and Attributes using \underline{Ri}ch \underline{V}isual \underline{A}ttribute \underline{S}egmentations (RIVAS10)}

\author{Mazda Moayeri$^{1}$\\
{\tt\small mmoayeri@umd.edu}
\and 
Phillip Pope$^{1}$\\
{\tt\small pepope@umd.edu}\\
\and 
Yogesh Balaji$^{2}$ \\
{\tt\small ybalaji@nvidia.com }\\
\and
Soheil Feizi$^{1}$ \\
{\tt\small sfeizi@cs.umd.edu} \\
\and
$^{1}$ Department of Computer Science \\
University of Maryland\\
\and
$^{2}$ NVIDIA \\
}
% \and
% Phillip Pope\\
% Department of Computer Science \\
% University of Maryland\\
% {\tt\small pepope@umd.edu}
% \and
% Yogesh Balaji\\
% Department of Computer Science \\
% University of Maryland\\
% {\tt\small yogesh@cs.umd.edu}
% \and
% Soheil Feizi\\
% Department of Computer Science \\
% University of Maryland\\
% {\tt\small sfeizi@cs.umd.edu}
% }
\maketitle
%%%%%%%%% ABSTRACT
\begin{abstract}
   While datasets with single-label supervision have propelled rapid advances in image classification, additional annotations are necessary in order to quantitatively assess how models make predictions. To this end, for a subset of ImageNet samples, we collect segmentation masks for the entire object and $18$ informative attributes. We call this dataset RIVAL10 ({\bf RI}ch {\bf V}isual {\bf A}ttributes with {\bf L}ocalization), consisting of roughly $26k$ instances over $10$ classes. Using RIVAL10, we evaluate the sensitivity of a broad set of models to noise corruptions in foregrounds, backgrounds and attributes. In our analysis, we consider diverse state-of-the-art architectures (ResNets, Transformers) and training procedures (CLIP, SimCLR, DeiT, Adversarial Training). We find that, somewhat surprisingly, in ResNets, adversarial training makes models more sensitive to the background compared to foreground than standard training. Similarly, contrastively-trained models also have lower relative foreground sensitivity in both transformers and ResNets. Lastly, we observe intriguing adaptive abilities of transformers to increase relative foreground sensitivity as corruption level increases. Using saliency methods, we automatically discover spurious features that drive the background sensitivity of models and assess alignment of saliency maps with foregrounds. Finally, we quantitatively study the attribution problem for neural features by comparing feature saliency with ground-truth localization of semantic attributes. %\SF{We find that neural nodes with strong overlap on top images in the training set do not necessarily generalize well to held-out data.}\SF{This statement is very strong and contradicts several prior works. We need to be very careful that inferences are accurate }

\end{abstract}

%%%%%%%%% BODY TEXT
% \begin{figure}
%     \centering
%     \newsubfloat{
%         \includegraphics[width=0.9\linewidth]{latex/figures/deit_small_60_equine.jpg}
%         }
%     \newsubfloat{
%       \includegraphics[width=0.9\linewidth]{latex/figures/robust_resnet50_60_frog.jpg}
%       }
%     \newsubfloat{
%       \includegraphics[width=0.9\linewidth]{latex/figures/simclr_60_deer.jpg}
%     }
%     \caption{Examples where background noise degrades performance of highly accurate models more than foreground noise. Gaussian $L_\infty$ noise with standard deviation $\sigma = 0.24$ shown. Probabilities are averaged over ten trials. While these examples are cherry picked, we observe that they are surprisingly prevalent, and model design can effect the degree to which such cases arise.}
%     \label{fig:examples}
% \end{figure}

\begin{figure}[t!]
    \centering
    \begin{subfigure}[t]{0.4\textwidth}
        \centering
        \includegraphics[width=0.85\textwidth]{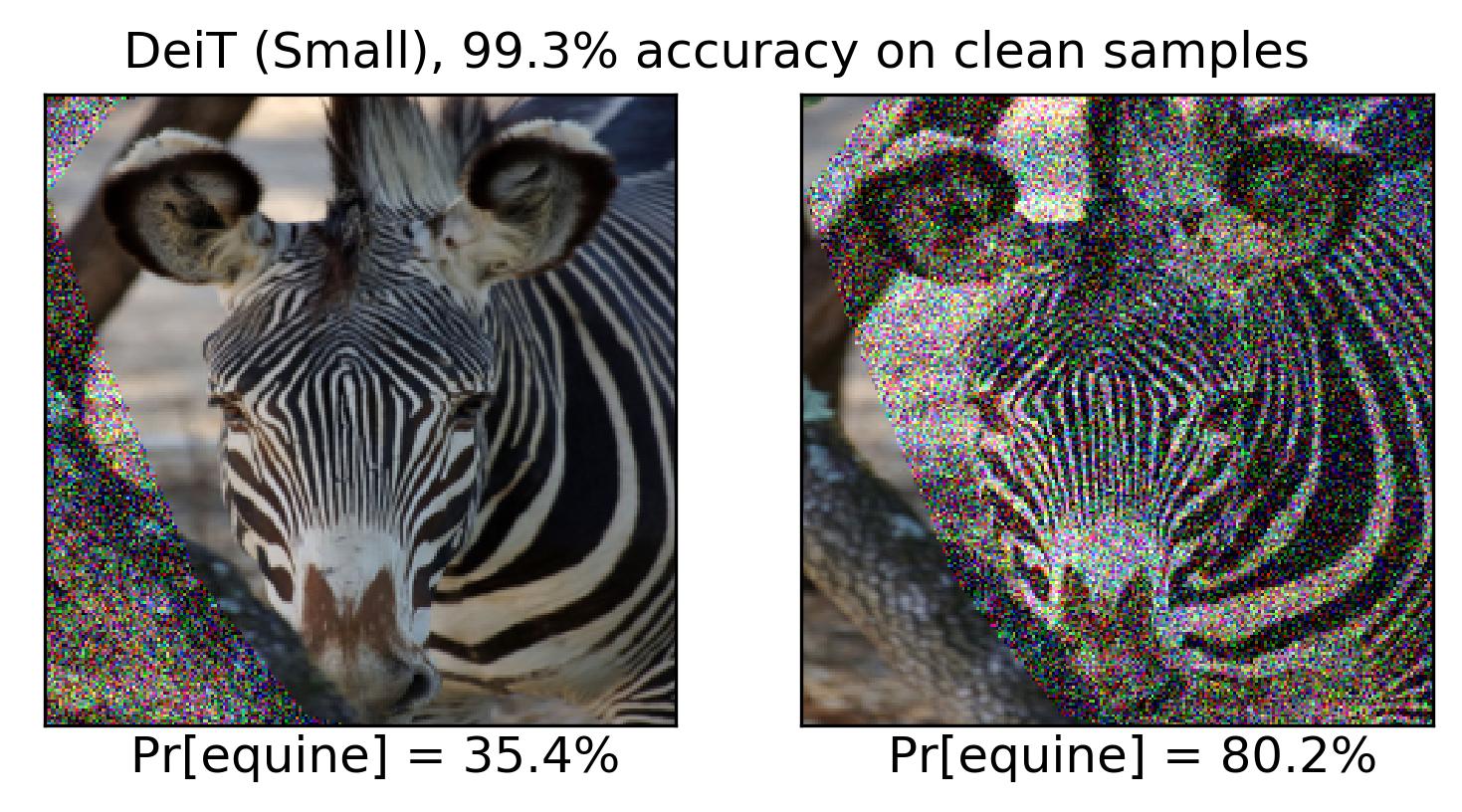}
    \end{subfigure}\\
    \begin{subfigure}[t]{0.4\textwidth}
        \centering
        \includegraphics[width=0.85\textwidth]{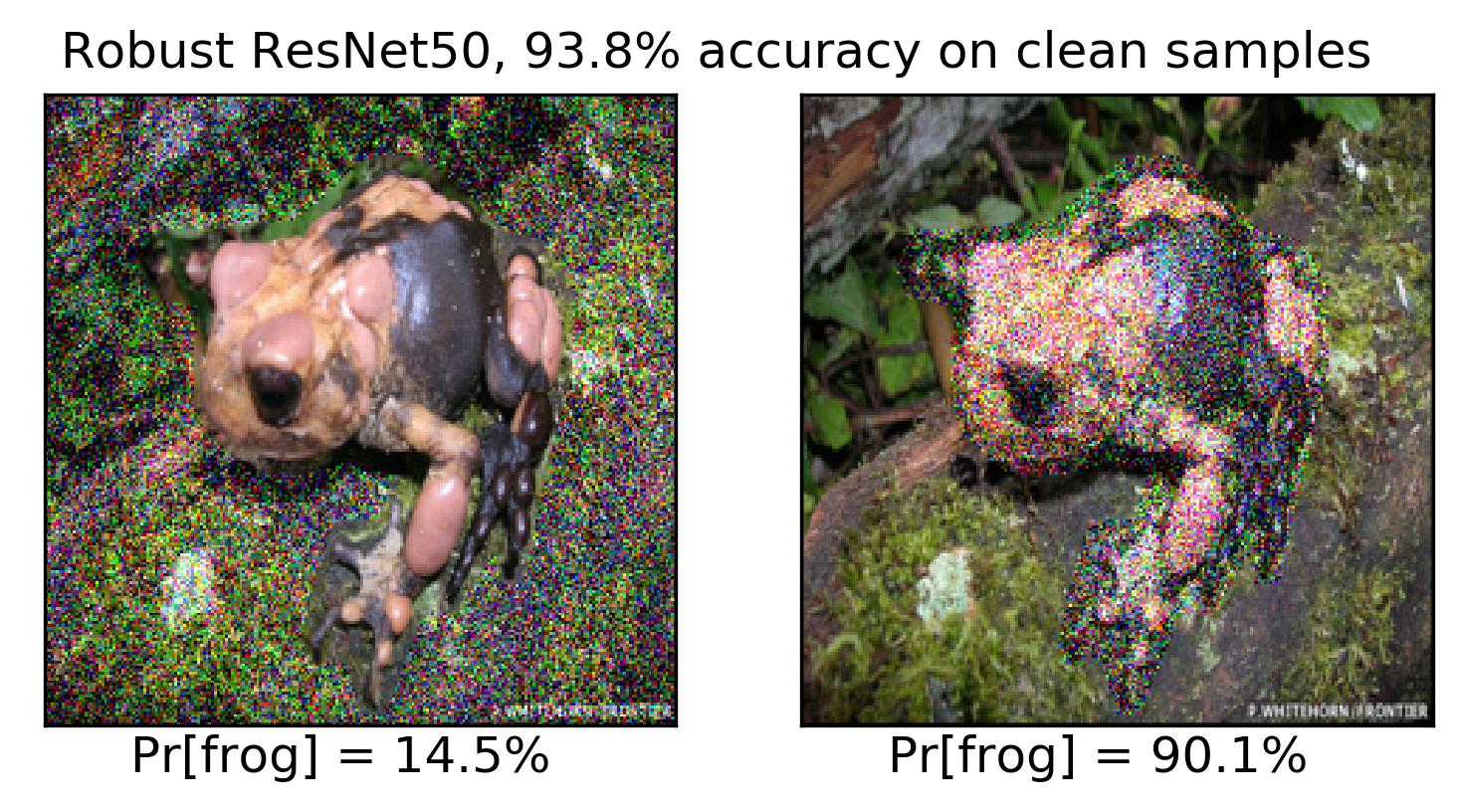}
    \end{subfigure}\\
    \begin{subfigure}[t]{0.4\textwidth}
        \centering
        \includegraphics[width=0.85\textwidth]{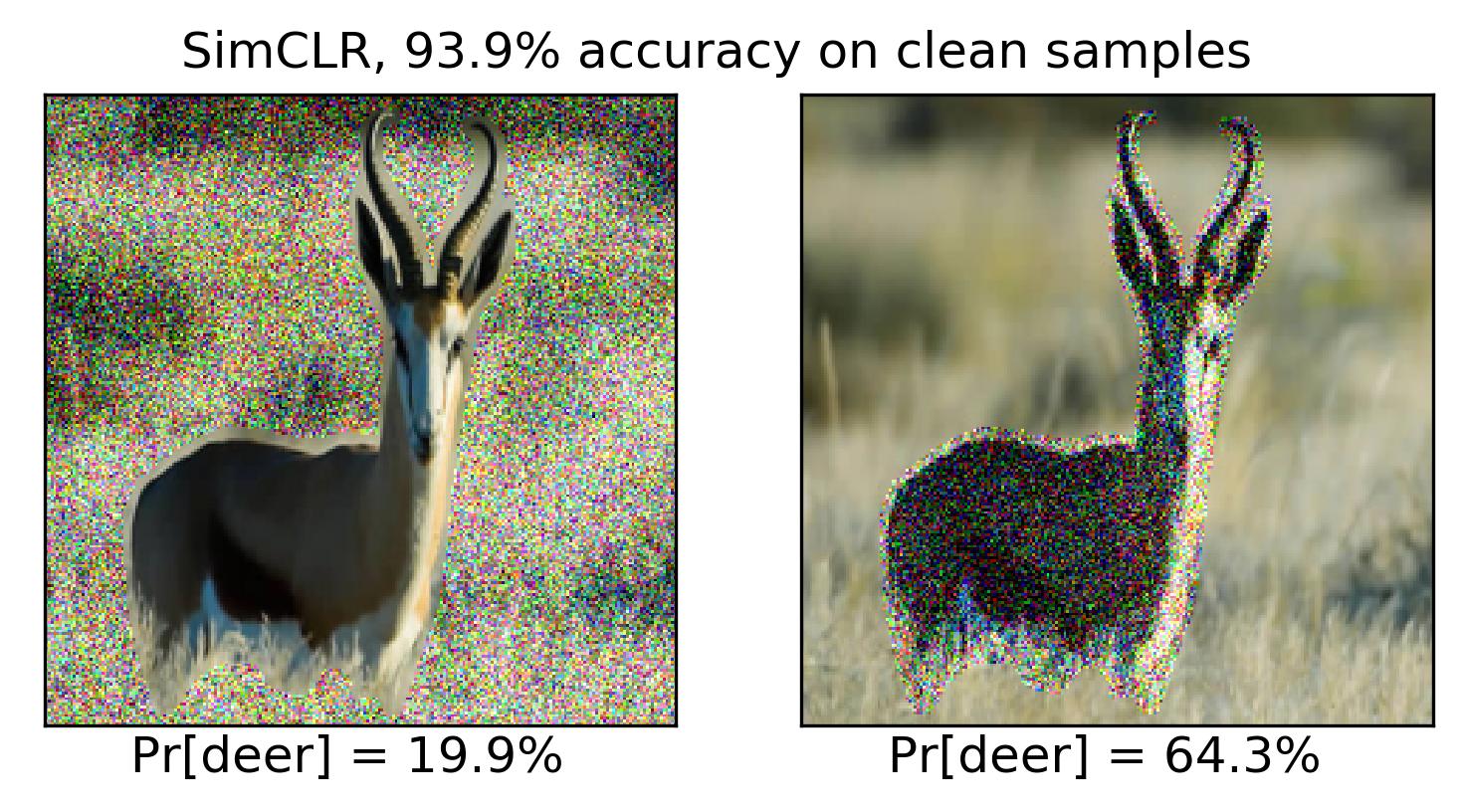}
    \end{subfigure}
    \caption{Examples where background noise degrades performance of highly accurate models more than foreground noise. Gaussian $\ell_\infty$ noise with standard deviation $\sigma = 0.24$ shown. Probabilities are averaged over ten trials. While these examples are cherry picked, we observe that they are surprisingly prevalent, and model design can affect the degree to which such cases arise.}
    \label{fig:examples}
\end{figure}

\begin{figure}[ht]
    \centering
    \includegraphics[width=\linewidth]{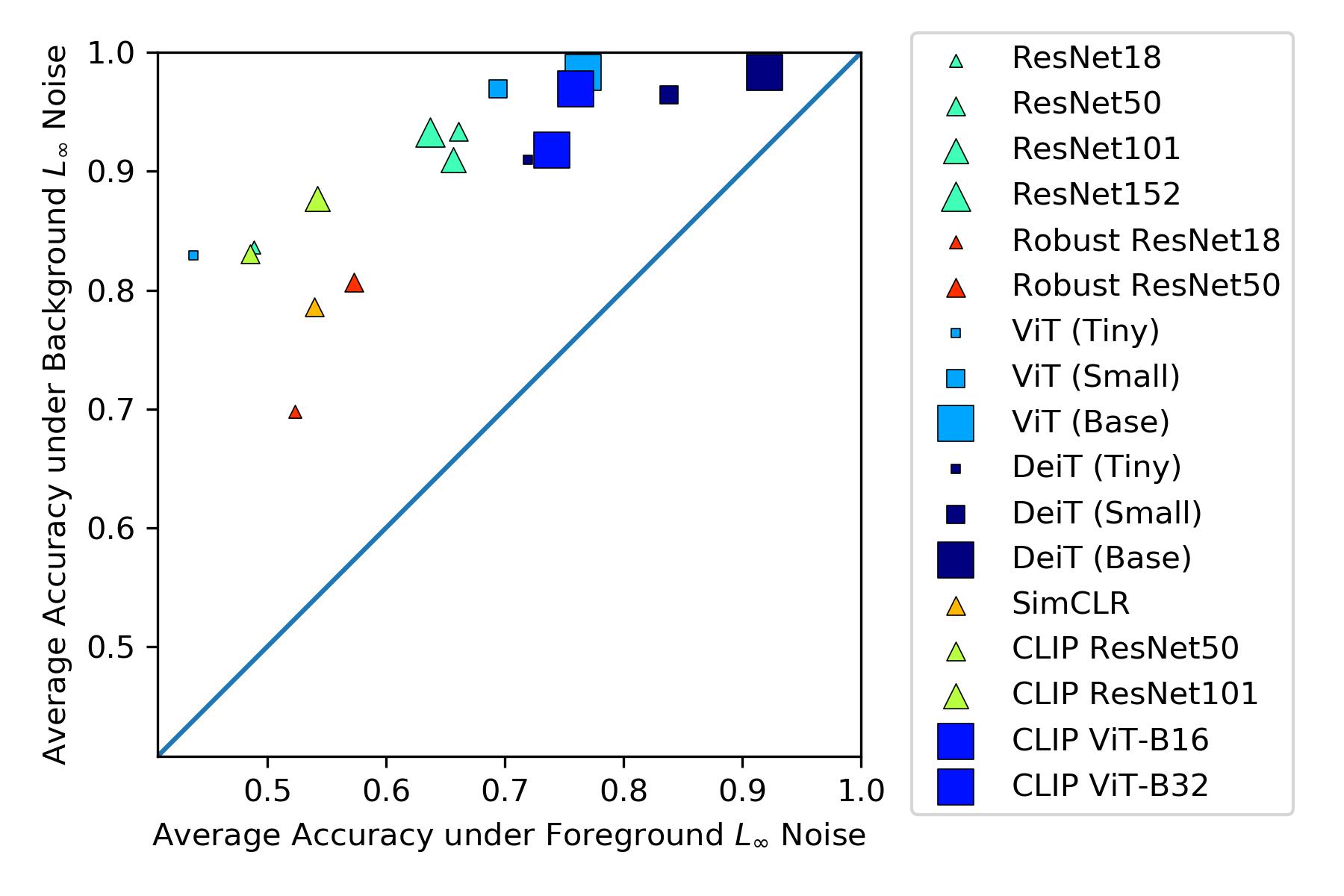}
    \caption{Accuracy under noise averaged over multiple noise levels. Marker size is proportional to parameter count. Models with higher relative foreground sensitivity lie further from the diagonal.}
    \label{fig:scatter_fg_bg}
\end{figure}

\section{Introduction}

Large scale benchmark datasets like ImageNet\cite{deng2009imagenet} that were constructed with single class label annotation have propelled rapid advances in the image classification task\cite{resnets, huang2017densely, tan2019efficientnet, zhai2021scaling}. Over the last decade, several network architectures and training procedures were proposed to yield very high classification accuracies \cite{simonyan2014very, resnets, tan2019efficientnet, vits}. However, methods to interpret these model predictions and to diagnose undesirable behaviors are fairly limited. One of the most popular class of approaches are saliency methods\cite{gradcam,cam,smoothgrad, integrated-grads} that use model gradients to produce a saliency map corresponding to the most influential input regions that yielded the resulting prediction. However, these methods are qualitative, require human supervision, and can be noisy, thus making their judgements potentially unreliable when made in isolation of other supporting analysis.

% To this end, we introduce a novel dataset that includes  {\bf RI}ch {\bf V}isual {\bf A}ttributions with {\bf L}ocalization (RIVAL10). Our dataset consists of images from 20 categories of ImageNet-1k \cite{}, with a total of roughly $26k$ high resolution images organized into 10 classes, matching those of CIFAR10 \cite{}. The main contribution of our dataset is instance wise labels for 18 informative visual attributes, as well as segmentation masks for each attribute and the entire object. We present our dataset as a tool for understanding models trained on ImageNet, and provide a study of the sensitivity of a diverse set of models  to foregrounds, backgrounds, and attributes. We hope to inspire readers to leverage RIVAL10's rich annotations to draw new insights in to the behavior of deep neural classifiers. 

In this paper, we argue that to obtain a proper understanding of how specific input regions impact the prediction, we need additional ground truth annotations beyond a single class label. To this end, we introduce a novel dataset, RIVAL10, whose samples include {\bf RI}ch {\bf V}isual {\bf A}ttributions with {\bf L}ocalization. RIVAL10 consists of images from $20$ categories of ImageNet-1k \cite{deng2009imagenet}, with a total of $26k$ high resolution images organized into $10$ classes, matching those of CIFAR10 \cite{cifar10}. The main contribution of our dataset is instance wise labels for $18$ informative visual attributes, as well as segmentation masks for each attribute and the entire object. We present our dataset as a general resource for understanding models trained on ImageNet. We then provide a study of the sensitivity of a diverse set of models to foregrounds, backgrounds, and attributes. 

% Our study couples a noise analysis with saliency methods. We first argue that model sensitivity can be measured by the degree to which adding noise to a region causes prediction performance to degrade. While all models are more sensitive to foreground corruption, we find that transformers and contrastively trained models are less relatively sensitive to foreground noise. Vision Transformers trained only on ImageNet (DeiTs) have the lowest foreground sensitivity, with the largest model being more sensitive to background noise nearly {\it half} the time. However, an ablation study reveals that only Transformer features can accurately classify images with backgrounds removed using a linear layer. Thus, we conclude transformers uniquely can choose to attend to multiple image regions, leading them to make more use of the backgrounds when they are informative, and to disregard them when they are not. 

Our study of background and foreground model sensitivity is motivated by some counter-intuitive model behaviors on images whose background and foreground regions were corrupted with Gaussian noise: Figure \ref{fig:examples} shows instances where highly accurate models have performance degraded much more due to the background noise than the foreground noise. While this is not the norm (i.e. models are more sensitive to foregrounds on average), the existence of these examples warrants greater investigation, as they expose a stark difference in how deep models and humans perform object recognition. Quantifying the degree to which different architectures and training procedures admit these examples can shed new insight on how models incorporate foreground and background information. 

To this end, we conduct a \emph{noise analysis} that leverages object segmentation masks to quantitatively assess model sensitivity to foregrounds relative to backgrounds. We proxy sensitivity to a region by observing model performance under corruption of that region. We propose a normalized metric, {\it relative foreground 
sensitivity} ($RFS$), to compare models with various general noise robustness. A high $RFS$ value indicates that the model uses foreground features in its inferences more than background ones since corrupting them result in higher performance degradation.

In Figure \ref{fig:scatter_fg_bg}, we see different architectures and training procedures lead to variations in both general noise robustness (projection onto the main diagonal) and relative foreground sensitivity (normalized distance orthogonal to the diagonal). Notably, we find that adversarially training ResNets significantly reduces $RFS$, surprisingly suggesting that robust ResNet models make greater use of background information. We also observe contrastive training to reduce $RFS$, and transformers to uniquely be able to adjust $RFS$ across noise levels, reducing their sensitivity to backgrounds as corruption level increases. Lastly, we find object classes strongly affect $RFS$ across models. %For example, across various models, ships and cats are often more sensitive to background noise suggesting models learn to utilize background content more than foreground in classifying these labels.

We couple our noise analysis with saliency methods to add a second perspective of model sensitivity to different input regions. Using RIVAL10 segmentations, we can quantitatively assess the alignment of saliency maps to foregrounds. We also show how we can discover spurious background features by sorting images based on the saliency alignment scores. We observe that performance trends that our noise analysis reveals are not captured using qualitative saliency methods alone, suggesting our noise analysis can provide new insights on model sensitivity to foregrounds and backgrounds.

%The saliency analysis provides qualitative understanding, but larger trends that the noise analysis reveals are not captured using saliency methods alone, suggesting our method can provide new insight on model sensitivity to foregrounds and backgrounds.

Lastly, we utilize RIVAL10 attribute segmentations to systematically investigate the generalizability of neural feature attribution: for a neural feature (i.e., a neuron in the penultimate layer of the network) that achieves the highest intersection-over-union (IOU) score with a specific attribute mask on top-$k$ images within a class, how the IOU scores of that neural feature behave on other samples in that class. For some class-attribute pairs (e.g. dog, floppy-ears), we indeed observe generalizability of neural feature attributions, in the sense that test set IOUs are also high.

In summary, we present a novel dataset with rich annotations of object and attribute segmentation masks that can be used for a myriad of applications including model interpretability. We then present a study involving three quantitative methods to analyze the sensitivity of models to different regions in inputs. We hope that our richly annotated RIVAL10 dataset helps studying failure modes of current deep classifiers and paves the way for building more reliable models in the future. 

%\SF{you have fig 1 in the intro but there is no explanation of it.}

%to inspire other researchers to make use of RIVAL10 dataset to gather further insights on the behavior of deep classifiers. 

%We hope to inspire other researchers to make use of RIVAL10 dataset to gather further insights on the behavior of deep classifiers. 

% Main findings:
% \begin{itemize}
%     \item Relative sensitivity to fg varies across architectures + object classes
%     \item Specifically, DeiTs are surprisingly not much more sensitive to fg than bg noise.
%     \item Robust ResNets are less relatively sensitive to fg than non robust resnets
%     \item Saliency in foreground does not translate to greater relative sensitivity to fg corruption
%     \item Attribute nodes exist; do they generalize ?
%     \item Models are pretty robust to attribute ablation
%     \item DCR: hybrid model has interpretability advantages; augmentation can improve generalization + saliency alignment of fine grained attr classification, but surprisingly hurts adv robustness
% \end{itemize}
%Need to review other attributed datasets and briefly explain collection procedure
%Figures: clustered correlations, attribute prevalences

\section{Review of Literature}
\subsection{Related Datasets}

Prior to the rise of deep learning, a number of works studied attribute classification, leading to the construction of datasets such as "Animals with Attributes" \cite{lampert2009learning} and aPASCAL VOC 2008 \cite{farhadi2009describing} (adding annotations to  \cite{Everingham10}). \cite{WelinderEtal2010} published the Caltech-UCSD Birds 200 (CUB), a fine-grained classification datasets of bird species with object segmentations and part {\it localizations} in the form of single coordinates as opposed to segmentation masks like in RIVAL10. Finally,  \cite{russakovsky2010attribute} collected object attributes on a small-scale subset of ImageNet. More recently, \cite{Ouyang_2015_ICCV} publish a large-scale object attribute dataset on a subset of ImageNet. The Celeb-A dataset from \cite{liu2015faceattributes} contains attribution and has applications to generative modeling, but limited utility for general representation learning since it only contains face images. A broader dataset is Visual Attributes in the Wild (VAW)\cite{pham2021learning}, which provides large-scale in the wild attribute annotations for ~250k object instances.

Many datasets aim to stress test models to reveal limitations. \cite{hendrycks2019robustness} introduces ImageNet variants under diverse corruption types, including Gaussian noise. \cite{hendrycks2021nae} adds two more ImageNet variants that include challenging natural samples and out of distribution samples, on which top models see massive accuracy drops. Models evaluated on \cite{objectnet} similarly see large drops, though this dataset differs in that it is strictly a test set. Other works introduce datasets related to the task of assessing background reliance of classifiers, such as \cite{madry_noise_or_signal} and \cite{waterbirds}, which perform some variation of swapping or altering foregrounds and backgrounds. Though similar, these works differ in objective and technical contribution to ours. \cite{waterbirds} focuses on developing a novel distributionally robust optimization procedure.  \cite{madry_noise_or_signal} emphasizes designing a multitude of test datasets through creative editing of foreground and background regions to serve as a general benchmark to evaluate models. In contrast, our work presents a novel method to analyze relative foreground sensitivity, and demonstrates its utility by applying it to a breadth of diverse, cutting edge models, engaging different architectures and training paradigms in a comprehensive fashion, leading to {\it model-specific} observations. Further, our RIVAL10 dataset is significantly larger and richer in annotation. 

Recently, \cite{singla2021salient} uses saliency maps and feature visualization in a semi-automated process to identify deep neural nodes corresponding to core or spurious features for an object of a given class, resulting in a large-scale dataset with segmentations corresponding to salient features. However, annotation of the segmented regions are limited to just labeling them as `core' or `spurious'. 

\subsection{Interpretability Methods}
A number of methods have been proposed to interpret model predictions, such as saliency or class activation maps\cite{gradcam}, influence functions \cite{koh2020understanding}, and surrogate white box models\cite{debuggable, lime}. However, saliency maps have been found to be noisy and influence functions are fragile \cite{fragile_interpretability, basu_fragile}. Certain methods seek to interpret the functions of neural nodes via synthesizing inputs that maximize the activation of the node \cite{olah2018_buildingblocks, ftr_viz,barlow}, though these methods are limited when non-adversarially robust models are used\cite{olah2017feature}, and largely offer qualitative insights. 

A motivation behind the development of interpretability methods is to work towards addressing the `shortcut learning' issue, where models rely on easy-to-learn features that lead to high performance on training sets, but poor generalization in other settings. \cite{shortcuts} discusses this at length, recommending the development and usage of challenging datasets whose inputs are out-of-distribution with respect to standard benchmarks. RIVAL10's rich annotations open the door to the construction of many challenge datasets, in which shortcuts are broken via swapping backgrounds, foregrounds, and {\it attributes}. We show examples of such crafted images in the appendix.  

Other constructive works aimed to reduce the reliance of deep models on spurious features appeal to counterfactual data generation \cite{chang2021towards, counterfactual_gan, adv_mix}, often appealing to disentangled representations or explicit annotations to break correlations of texture, shapes, colors, and backgrounds. Further, \cite{fereshte} found that removing spurious features can in fact hurt accuracy and disproportionately affect groups. Thus, the notion that spurious features are always harmful is incomplete, and a closer look is required to ground discussions regarding the shortcut learning issue. Lastly, \cite{counterfactual_theory} provides theoretical context for stress testing models to discern causal factors.

\section{RIVAL10}
\begin{figure*}
\vspace{-0.2cm}
\centering
\begin{minipage}{.495\textwidth}
  \centering
  \includegraphics[width=\linewidth]{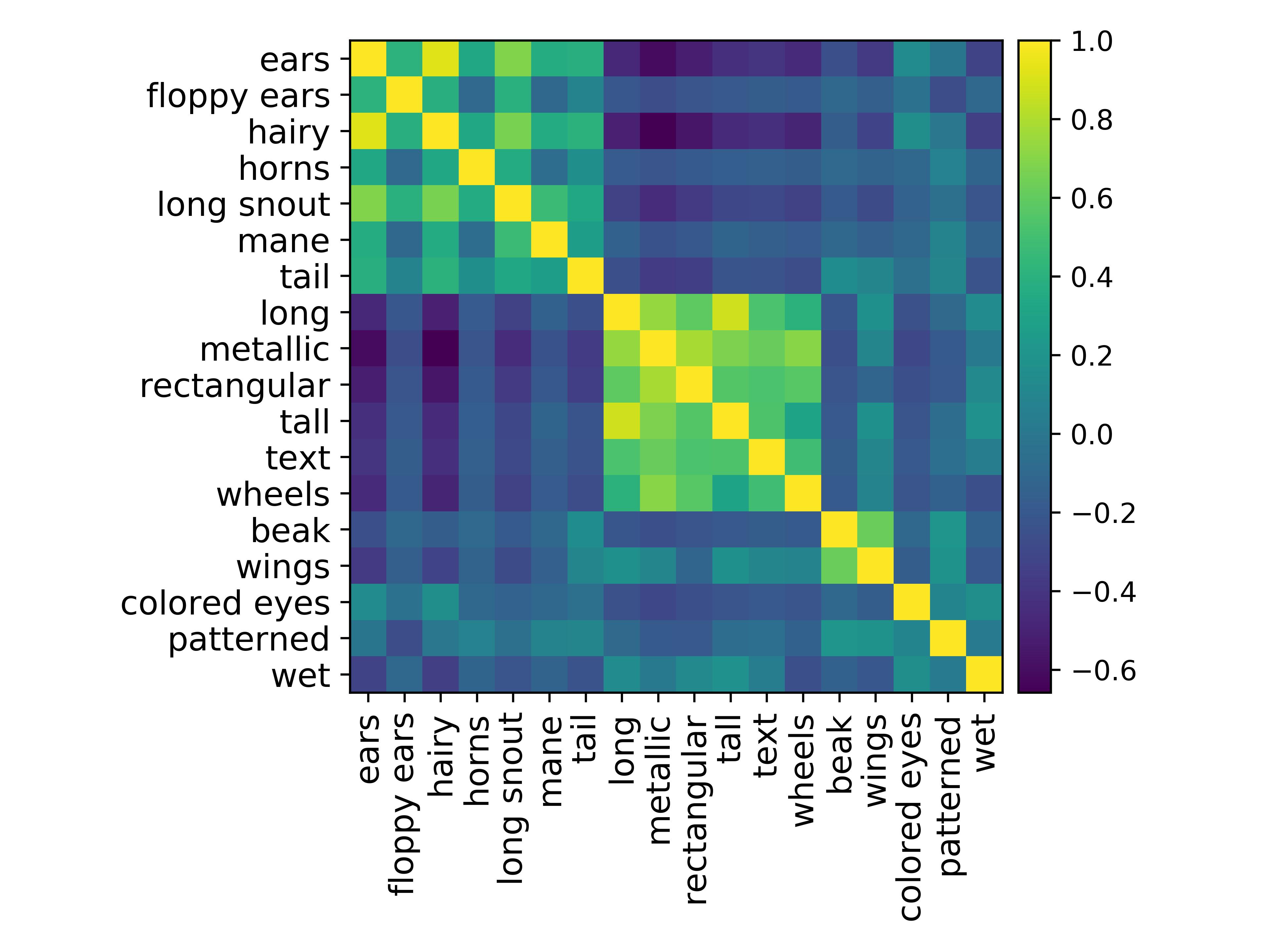}
\end{minipage}
\begin{minipage}{.495\textwidth}
  \centering
  \includegraphics[width=\linewidth]{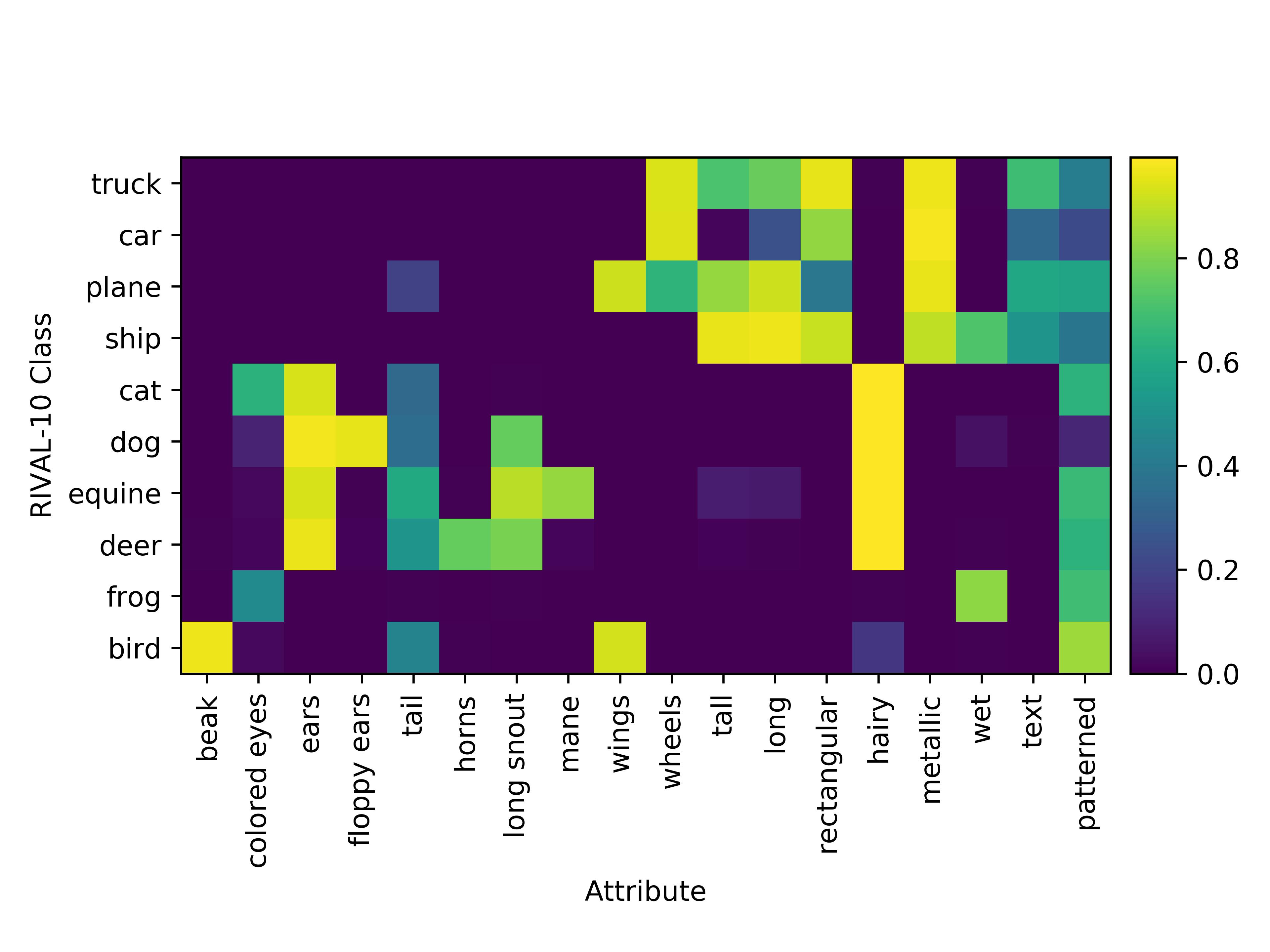}
\end{minipage}%
\caption{(\textbf{Left}): Correlations between attributes in the training split.
  (\textbf{Right}): Class-wise means of attribute vectors in the training split.}
\label{fig:attr-corr-and-class-attr-means}
\vspace{-0.4cm}
\end{figure*}
\subsection{Overview}

RIVAL10 differs from previous attributed datasets in that it provides \textit{attribute-specific} localizations. That is, for every positive instance of an attribute, a polygonal segmentation mask (possibly multiple parts) is provided to identify the image region in which the attribute occurs.

\looseness=-1
Perhaps, the most similar dataset in this regard is the recent Fashionpedia \cite{fashionpedia}, a dataset providing attributes and localizations of $27$ apparel categories. However, the dataset is proposed for the fashion domain which limits its utility for general purpose object recognition task. To the best of our knowledge, RIVAL10 is the first \textit{general domain} dataset to provide both rich semantic attributes and localization, the combination of which we envision to aid in analyzing the robustness and interpretability of deep networks. While other datasets used for semantic segmentation and object detection go beyond single label annotations \cite{coco, Everingham15, Cordts2016Cityscapes}, they are not designed with classifiers specifically in mind, like RIVAL10. 

Classes were chosen to be aligned with CIFAR-10 to enable analyzing the existing architectures and training techniques developed for the object recognition task. In particular, the classes we provide are: \textit{bird, car, cat,  deer, dog, equine, frog, plane, ship, truck}. We collected the following attributes for these object categories: \textit{beak, colored-eyes, ears, floppy-ears, hairy, horns, long, long-snout, mane, metallic, patterned, rectangular, tail, tall, text, wet, wheels, wings}. Some attributes were inspired from \cite{russakovsky2010attribute}. 
%We visualize samples from RIVAL-10 in Figure \ref{fig:samples}.

We chose attributes to be intuitively informative, capturing semantic concepts that humans may allude to in classifying RIVAL10 objects. While the attributes contain some redundant information, they are nonetheless discriminative in the sense that a linear classifier on attributes achieves $93.3\%$ test accuracy. We visualize attribute correlations and class-wise averages in Figure \ref{fig:attr-corr-and-class-attr-means}. 
%Notice that class-wise means differ significantly in one or more attributes for any pair of classes. 

%Attributes were designed to be discriminative between classes but not one-to-one with classes. Given any attribute, its presence or absence does not imply a class. Attributes are discriminative in the sense that they \textit{approximately linearly separate classes}. For instance, our dataset achieves around $93.3\%$ test accuracy with a linear classifier on attributes. We visualize this separability in Figure \ref{fig:attr-corr-and-class-attr-means} (right), where the mean class-wise  attribute vectors are show in a matrix and cells are colored by the empirical frequency of that attribute. We can see that classes on average are distinguishable by attributes. We choose attributes to be intuitively informative, capturing semantic concepts that humans may allude to in classifying RIVAL10 objects. 

\subsection{Data Collection}

All images were sourced from ImageNet \cite{deng2009imagenet}. The images used in each RIVAL10 class were derived from pairs of related ImageNet classes. In other words, $20$ classes from Imagenet were used to build the $10$ RIVAL10 classes (details in appendix). 
%Naturally some correlation structure exists between attributes. We visualize these correlations in Figure \ref{fig:attr-corr-and-class-attr-means} (left).
To collect our attributes and localizations, we hired workers from Amazon Mechanical Turk (AMT). Data collected through AMT without careful control may be of low quality. To encourage quality annotations, we utilize strategies recommended by the HCI community \cite{hci-amt}: providing detailed instructions, \textit{screening} workers for aptitude, and monitoring worker performance with attention checks.

Binary attributions were collected first. Workers were required to pass a qualification test of 20 images with known ground truth attributes: only workers who achieved a minimal overall precision and recall of $0.75$ were hired for full data collection. Because the task of segmentation is more involved than indicating whether or not an attribute is present, we required a second qualification test, assessing annotation quality by computing intersection-over-union (IOU) of the submitted attribute masks with ground truth masks. Workers were required to complete five segmentations with an average IOU of at least $0.7$. This strategy encourages selection of workers who have demonstratively read and understood the instructions.

To ensure that quality is maintained in both the attribution and segmentation phases, roughly 5\% of images provided to workers to annotate already had ground truth labels. These so-called attention checks allowed for the monitoring of annotation quality during the collection process. In the first stage of collecting binary attribute labels, the average precision and recall scores were $0.81$ and $0.84$ respectively. For each positive instance of an attribute marked in the first phase of data collection, an attribute segmentation was collected in the second phase. Completeing attribute segmentations in a second pass allowed for the review of the binary attributions and the removal of any false positives. Average IOU of attention checks completed during the second phase of data collection was $0.745$. 

Further details about our data collection pipeline, including images of our instructions shown to workers, histograms of scores on the qualifying exam and attention checks, selection process of the AMT workers, payments and other details can be found in the appendix.

\section{Models}
\begin{figure*}[h!]
    \centering
    \includegraphics[width=0.9\textwidth]{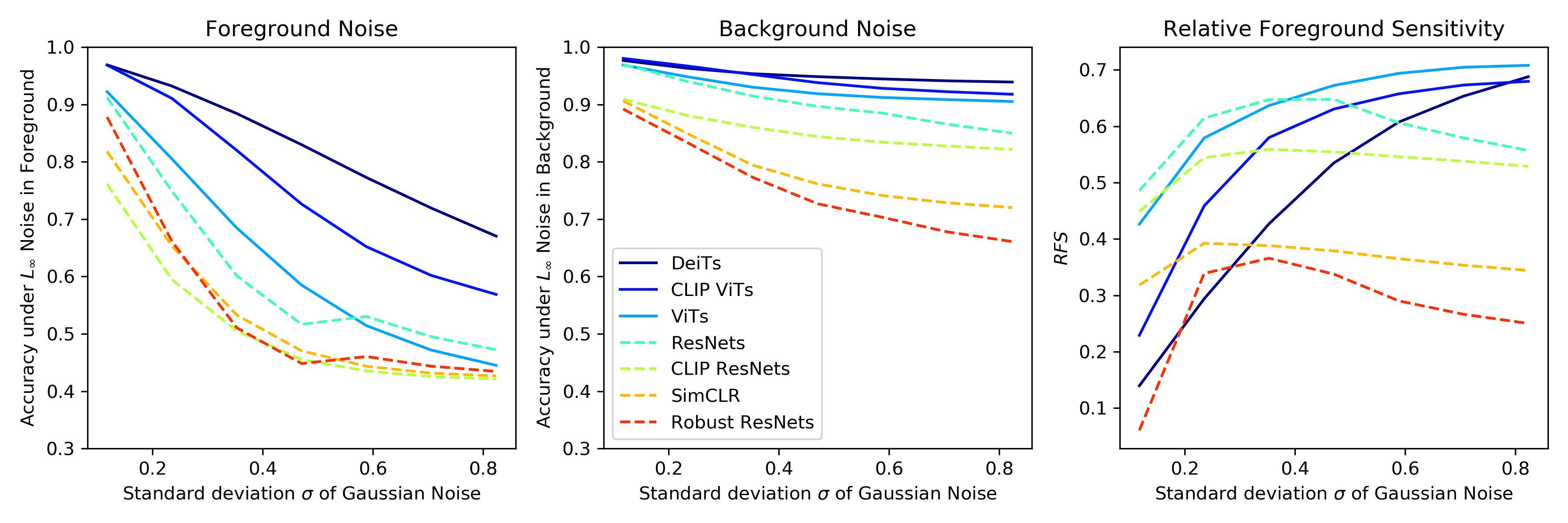}
    \caption{Accuracy under noise in foreground ({\bf left}) and background ({\bf middle}) at various noise levels. Models are grouped by architecture and training procedure, with a curve corresponding to the average over all models in a group. ({\bf Right}): $RFS$ by group.}
    \label{fig:accs_over_varying_noise}
\end{figure*}

In our analyses, we focus on ResNets and Vision Transformers\cite{resnets, vits}. We inspect ResNets trained (i) in a standard supervised fashion, (ii) adversarially via $\ell_2$ projected gradient descent\cite{pgd}, and (iii) contrastively (i.e. no direct label supervision), with SimCLR and CLIP\cite{SimCLR, clip}. We also consider CLIP Vision Transformers, as well as standard Vision Transformers (ViT) and Data efficient Image Transformers (DeiT)\cite{deit}. DeiTs differ from ViTs primarily in their training set, solely using ImageNet-1k while  ViTs used ImageNet-21k. %JFT is Google's internal dataset.
To make up for not having the inductive biases of ResNets, ViTs increased the amount of training data, while DeiTs instead rely upon extensive augmentation. All other models, with the exception of those trained with CLIP, use ImageNet-1k as the pretraining set. CLIP, on the other hand, uses a much larger dataset of images and associated text. A full discussion on models is offered in the appendix. 

To perform classification on RIVAL10 dataset, we attach a linear head to the features extracted from the pen-ultimate layer of the base models. All base models have their weights frozen (from pretraining) and are not updated during fine-tuning. This preserves the feature space learned in the original pretraining. All models achieve upwards of 90\% accuracy on the RIVAL10 test set, essentially controlling for classification ability. We note that while there is leakage between ImageNet-1k and the RIVAL10 test set, the purpose of this study is not to improve model's predictive accuracy directly, but instead to better understand the information used in making predictions. 

Recently, a number of works compare the robustness of ViTs to ResNets. %\cite{vits_at_least_as_robust} conclude that with sufficient data and training time, ViTs are at least as robust as ResNets on a wide range of input and model perturbations. \cite{vits_robust_attn_is_key} find improved robustness to commoncorruptions, distribution shifts, and natural adversarial examples, claiming attention is key to the gains. On adversarial robustness, \cite{vits_maybe_more_robust} claim ViTs possess greater adversarial robustness than convolutional neural networks (CNNs), though \cite{vits_about_same} push back, claiming CNNs can be as adversarially robust if they adopt the same training recipes as ViTs. 
While there are mixed findings on adversarial robustness \cite{vits_maybe_more_robust, vits_about_same}, there is agreement that ViTs have stronger out-of-distribution generalization, likely due to self attention \cite{vits_at_least_as_robust, vits_robust_attn_is_key}. In contrast, our work focuses on relative robustness to noise in foreground and background regions. 
%\cite{vits_maybe_more_robust} suggest self-attention may be key to improved robustness, though \cite{vits_about_same} pushes back, claiming increasing training time and data to match ViTs can improve \cite{vits_at_least_as_robust, vits_about_same, vits_maybe_more_robust} \SF{what mixed results? explain.}. In contrast, our work does not focus on general robustness \SF{what is general robustness?}, but instead uses robustness to noise in specific regions as a tool to discern model sensitivities. 

\section{Foreground and Background Sensitivity}
% \begin{figure*}[h!]
%     \centering
%     \includegraphics[width=0.9\textwidth]{latex/figures/accs_under_noise_linf_normed_diff.jpg}
%     \caption{Accuracy under noise in foreground ({\bf left}) and background ({\bf middle}) at various noise levels. Models are grouped by architecture and training procedure, with a curve corresponding to the average over all models in a group. ({\bf Right}): $RFS$ by group.}
%     \label{fig:accs_over_varying_noise}
% \end{figure*}
%Motivating question: to what degree do models rely on backgrounds in making their predictions?

\subsection{Noise Analysis}
We add noise to the foreground and background separately to see how corrupting each region degrades model performance. Consider a sample $\mathbf{x}$ with a binary object mask $\mathbf{m}$ where $\mathbf{m}_{i,j}=1$ if the pixel $\mathbf{x}_{i,j}$ is a part of the object. We first construct a noise tensor $\mathbf{n}$ that has pixel values drawn i.i.d. from $\mathcal{N}(0,\sigma^2)$, where $\sigma$ is a parameter controlling the noise level. Then, we obtain noisy-background $\mathbf{\tilde{x}}_{bg}$ and noisy-foreground $\mathbf{\tilde{x}}_{fg}$ samples as:
$$\mathbf{\tilde{x}}_{fg} = \text{clip}(\mathbf{x} + \mathbf{n} \odot \mathbf{m}), \; \mathbf{\tilde{x}}_{bg} = \text{clip}(\mathbf{x} + \mathbf{n} \odot (\mathbf{1}-\mathbf{m}))$$
\looseness=-1
where $\odot$ is the hadamard product, and `clip' refers to clipping all pixel values to the $[0,1]$ range. We add Gaussian noise so to preserve the image content. Note that additive pixel-wise noise leads to the same magnitude of perturbation in the foreground and background under the $\ell_\infty$ norm. We also repeat our analysis with $\ell_2$ normalized noise (presented in the appendix) to avoid a bias against larger regions and obtain similar results.

%to avoid a bias against larger regions. However, the relative size of foregrounds and backgrounds varies across samples, so the direction of this bias is unclear. 
%This norm introduces a bias against larger regions, which under the $L_2$ norm, are corrupted more by our noise addition. The relative size of the foreground and background regions can vary greatly across data points, so the direction of this bias is unclear. For completeness, we repeat experiments with the $L_2$ normalized noise and obtain similar results, presented in the appendix. 

We seek to quantify the sensitivity of a model to foregrounds relative to its sensitivity to backgrounds. To this end, we introduce {\it relative foreground sensitivity} ($RFS$). Let $a_{fg}$ and $a_{bg}$ denote accuracy under noise in the foreground and background, respectively, and $\bar{a} := (a_{fg}+a_{bg})/2$ denote their mean (referred to as the general noise robustness). We then define $RFS$ for a model $\mathbf{F}$ as 
$$RFS(\mathbf{F}) = \frac{a_{bg}-a_{fg}}{2 \min(\bar{a}, 1-\bar{a})}.$$

Essentially, $RFS$ normalizes the gap in model performance under foreground and background noise by the total possible gap, given the general noise robustness of the model. In Figure \ref{fig:scatter_fg_bg}, $RFS$ takes on the geometric meaning of the ratio between the distance of $(a_{fg}, a_{bg})$ to $(\bar{a}, \bar{a})$, to the largest possible distance from the diagonal in the unit square for a point with general noise robustness $\bar{a}$. The scale factor in the denominator gives $RFS$ a range of $[-1,1]$, with larger values corresponding to greater relative foreground sensitivity.%Intuitively, models with very high (low) noise robustness will have $(a_{fg}, a_{bg})$ lie in the top right (bottom left) of the unit square. Thus, our normalization scales up the gap between $a_{fg}$ and $a_{bg}$ to reflect how general noise robustness limits the size of the gap. 
\begin{figure*}[h!]
    % \newsubfloat{
    % \begin{subfloat}
        \includegraphics[width=\linewidth]{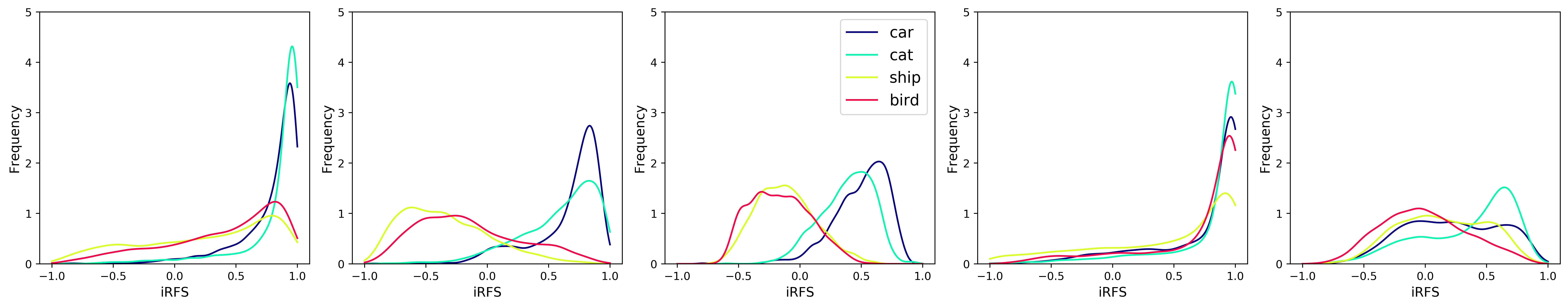}
        % }
    % \end{subfloat}
    % \begin{sufloat}
    % \newsubfloat{
      \includegraphics[width=\linewidth]{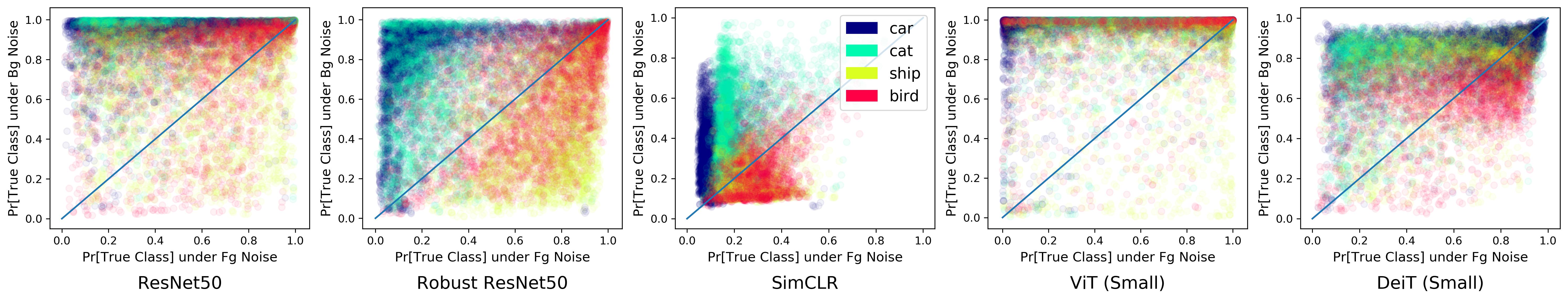}
    %   }
    % \end{sufloat}
    \centering
    \vspace{-0.5cm}
    \caption{Relative foreground sensitivity per instance for four classes and five models of roughly equal size. ({\bf Top}): Histogram of $iRFS$; positive denotes greater foreground sensitivity. ({\bf Bottom}): Scatter; top left indicates high relative foreground sensitivity. Across models, ships and birds have low foreground sensitivity, often being more sensitive to noise in the background than the foreground.} 
    \label{fig:noise_hist_and_scatter}
\end{figure*}

We also consider an instance-wise version, $iRFS$, defined for a model $\mathbf{F}$ {\it and} a sample $\mathbf{x}$. Here, we use the probability that model $\mathbf{F}$ predicts sample $\mathbf{x}$ to belong to its true class as the measure of model performance instead of accuracy. Let $p_{fg}$ and $p_{bg}$ denote this probability for $\tilde{\mathbf{x}}_{fg}$ and $\tilde{\mathbf{x}}_{bg}$, respectively. Thus, with $\bar{p} := (p_{fg}+p_{bg})/2$, 
$$iRFS(\mathbf{F}, \mathbf{x}) = \frac{p_{bg}-p_{fg}}{2 \min(\bar{p}, 1-\bar{p})}.$$

In our experiments, we consider seven equally spaced noise levels from $\sigma=30/255$ to $210/255$. For each sample in the test set, we take ten trials of adding noise to the foreground and background separately {\it per noise level}. RIVAL10's test set consists of roughly $5k$ images, so for each model type, we assess $5k \times 7 \times 10 = 350,000$ trials in total.

\subsection{Empirical Observations}
Fig. \ref{fig:scatter_fg_bg} shows different models have vastly different performance in terms of both general noise robustness and relative foreground sensitivity.
%First, we note that across models, there is variance in both general noise robustness and relative foreground sensitivity. 
In Figure \ref{fig:scatter_fg_bg}, transformers generally lie further up the main diagonal than ResNets, corroborating observations that transformers are more robust to common corruptions. Increasing model size improves general robustness, though it does more so for transformers than ResNets. Models lie at different distances orthogonal to the diagonal as well, indicating architecture and training procedure affect relative foreground sensitivity.

In Figure \ref{fig:accs_over_varying_noise}, we categorize model types based on architecture and training procedure, averaging $RFS$ over groups to reveal general trends. Robust ResNets have the lowest $RFS$, much lower than standard ResNets, a somewhat surprising result given that background reliance has been thought to be linked to increased adversarial vulnerability in the past \cite{nlp_spur, madry_noise_or_signal}. SimCLR has the next lowest $RFS$ overall, and generally, contrastive training procedures (CLIP, SimCLR) seem to reduce $RFS$ in both ResNets and ViTs. 

In comparing transformers to ResNets overall, we see at low noise levels, transformers sometimes have lower $RFS$ than ResNets. Interestingly, as noise level rises, $RFS$ in transformers increases as well, while $RFS$ is mostly stable for ResNets. This suggests that transformers can adaptively alter the attention paid to different image regions based on the level of corruption. Comparing between transformers, we see DeiTs with much lower $RFS$ than ViTs, suggesting that the heavy augmentations DeiTs leveraged to achieve increased data efficiency may have also made the models much more sensitive to backgrounds. 
\begin{figure}[h!]
\vspace{-0.2cm}
\centering
  \centering
  \includegraphics[width=\linewidth]{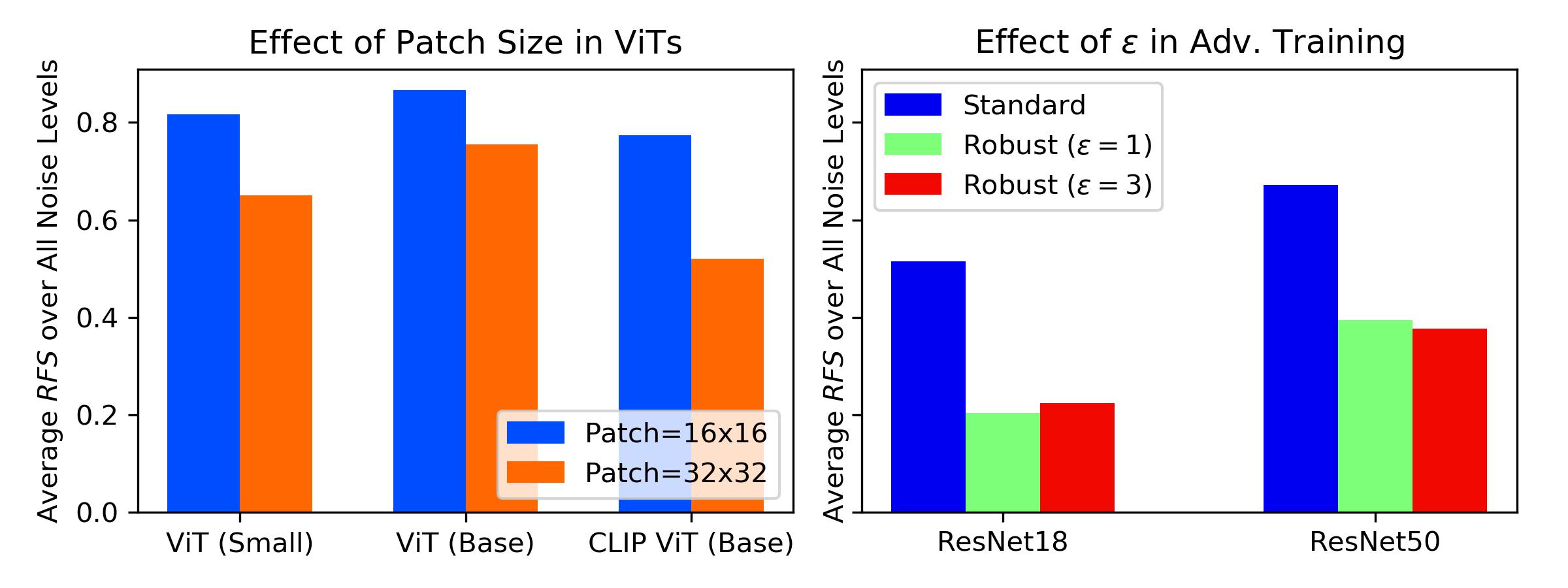}
\caption{Controlled ablation studies. Average $RFS$ over all noise levels presented for brevity. ({\bf Left}): Increasing patch size in ViTs decreases relative foreground sensitivity. ({\bf Right}) Robust models are much less relatively sensitive to foregrounds, but $\epsilon$ used in adversarial training does not affect $RFS$ much.}
\label{fig:ablations_patch_robustness}
\vspace{-0.4cm}
\end{figure}
\begin{figure*}[h!]
\vspace{-0.2cm}
\centering
\begin{minipage}{.496\textwidth}
  \centering
  \includegraphics[width=\linewidth]{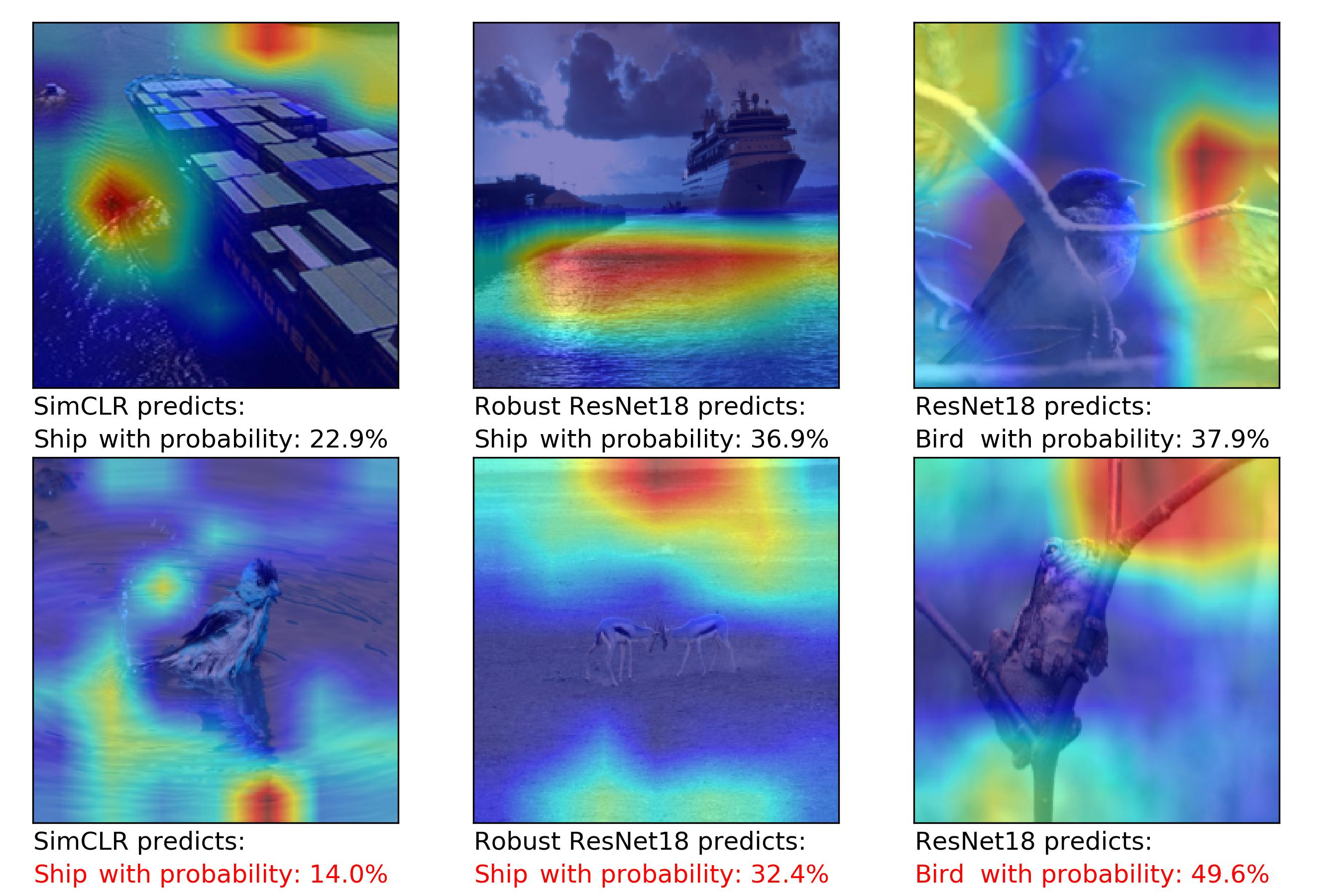}
\end{minipage}
\begin{minipage}{.496\textwidth}
  \centering
  \includegraphics[width=\linewidth]{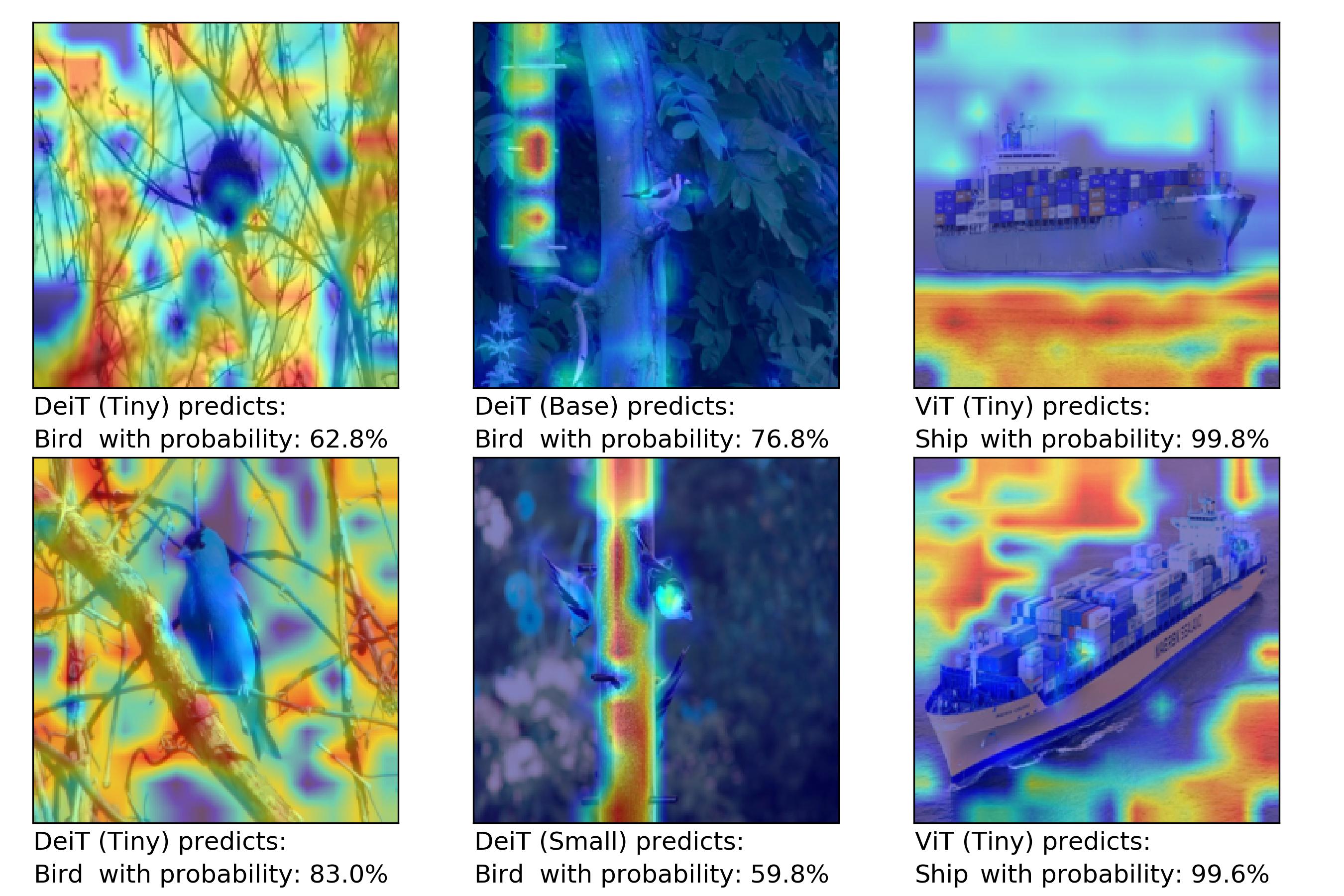}
\end{minipage}%
\caption{Instances of images with low saliency alignment, highlighting spurious features of water for ships, and branches and bird feeders for birds. (\textbf{Left}): Spurious features leading to misclassification (in red). (\textbf{Right}): Other instances of spurious features. }
\label{fig:spurious_birds_ships}
\vspace{-0.4cm}
\end{figure*}

In Figure \ref{fig:ablations_patch_robustness}, we more closely inspect the effect of patch sizes in ViTs and the attack budget $\epsilon$ used in adversarial training (which affects accuracy-robustness trade-off). We find that increasing the patch size in ViTs from $16\times 16$ to $32 \times 32$ reduces $RFS$ when averaged over all noise levels. The robustness ablation affirms that robust ResNets are much less relatively sensitive to foregrounds than standard ResNets, though the attack size seen in training does not seem to significantly affect $RFS$. 
\begin{figure}[ht]
    \centering
    \includegraphics[width=0.47\textwidth]{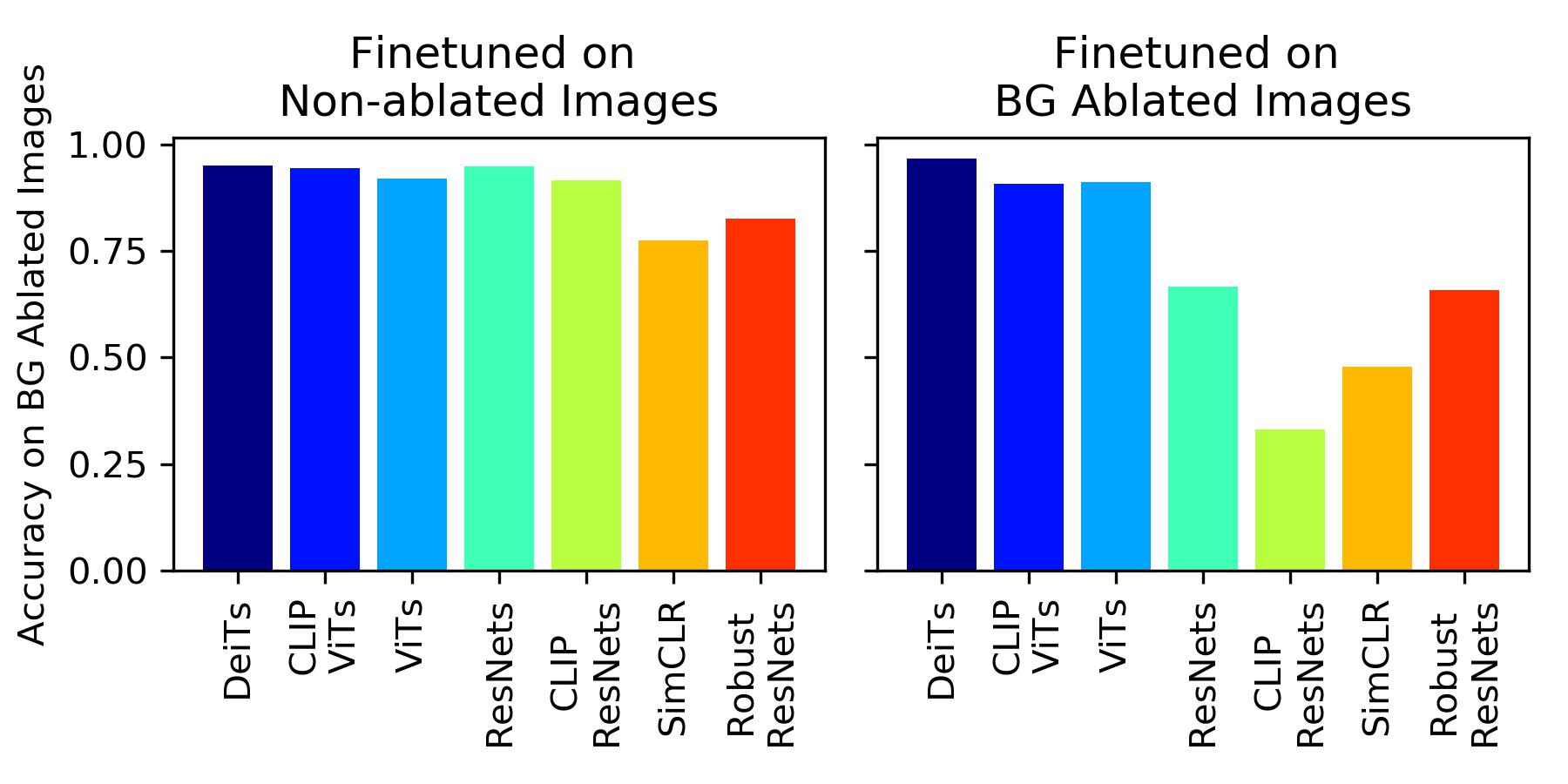}
    \caption{Model accuracy on images with backgrounds ablated via graying. The right plot shows accuracies for models finetuned on the images with backgrounds ablated. Only transformers can fit a linear layer on the features of background ablated images without compromising performance.}
    \label{fig:bg_ablated}
\end{figure}
% \begin{figure*}[h!]
%     \newsubfloat{
%         \includegraphics[width=\linewidth]{figures/hist_prob_diffs_linf_normalize.jpg}
%         }
%     \newsubfloat{
%       \includegraphics[width=\linewidth]{figures/scatter_probs_under_noise_linf_subset.jpg}
%       }
%     \centering
%     \vspace{-0.5}
%     \caption{Relative foreground sensitivity per instance for four classes and five models of roughly equal size. ({\bf Top}): Histogram of $iRFS$; positive denotes greater foreground sensitivity. ({\bf Bottom}): Scatter; top left indicates high relative foreground sensitivity. Across models, ships and birds have low foreground sensitivity, often being more sensitive to noise in the background than the foreground.} 
%     \label{fig:noise_hist_and_scatter}
% \end{figure*}
% \begin{figure*}[h!]
% \vspace{-0.2cm}
% \centering
% \begin{minipage}{.496\textwidth}
%   \centering
%   \includegraphics[width=\linewidth]{figures/mistakes.jpg}
% \end{minipage}
% \begin{minipage}{.496\textwidth}
%   \centering
%   \includegraphics[width=\linewidth]{figures/other.jpg}
% \end{minipage}%
% \caption{Instances of images with low saliency alignment, highlighting spurious features of water for ships, and branches and bird feeders for birds. (\textbf{Left}): Spurious features leading to misclassification (in red). (\textbf{Right}): Other instances of spurious features. }
% \label{fig:spurious_birds_ships}
% \vspace{-0.4cm}
% \end{figure*}

Moving away from comparing models, in Figure \ref{fig:noise_hist_and_scatter}, we see {\bf foreground sensitivity is largely affected by class}. In particular, across models of roughly equal size, ships and cats are often more sensitive to background noise, suggesting models learn to utilize background content more than foreground content in recognizing them. The class distinction is less pronounced in DeiTs and ViTs, with ViTs assigning high foreground sensitivity for all classes, and DeiTS having mixed sensitivity across classes, with many $iRFS$ scores larger than $0$.

% \begin{figure*}[h!]
% \vspace{-0.2cm}
% \centering
% \begin{minipage}{.496\textwidth}
%   \centering
%   \includegraphics[width=\linewidth]{figures/mistakes.jpg}
% \end{minipage}
% \begin{minipage}{.496\textwidth}
%   \centering
%   \includegraphics[width=\linewidth]{figures/other.jpg}
% \end{minipage}%
% \caption{Instances of images with low saliency alignment, highlighting spurious features of water for ships, and branches and bird feeders for birds. (\textbf{Left}): Spurious features leading to misclassification (in red). (\textbf{Right}): Other instances of spurious features. }
% \label{fig:spurious_birds_ships}
% \vspace{-0.4cm}
% \end{figure*}
\begin{figure}
    \centering
    \includegraphics[width=0.47\textwidth]{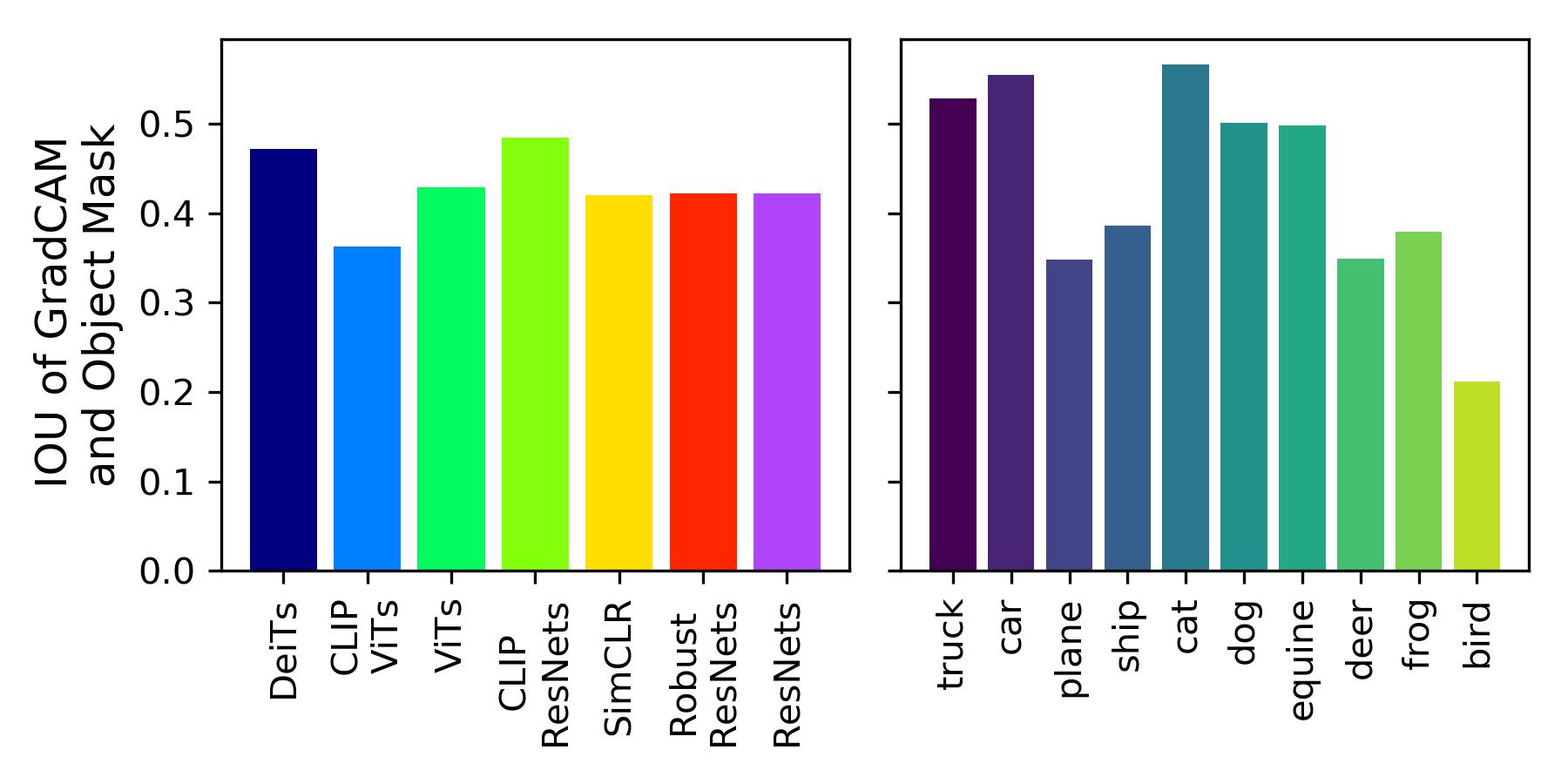}
    \caption{Alignment of binarized saliency maps with object segmentation masks, measured by intersection over union (IOU). Averaged over models ({\bf left}) and object classes ({\bf right}).}
    \label{fig:ious_sal_alignment}
\end{figure}
\subsection{Removing Backgrounds Entirely}
We also inspect the accuracy of models on images with backgrounds grayed out, similar to \cite{madry_noise_or_signal}, though now considering ViTs, CLIP, and SimCLR, which had not been developed at the time of their study. Also, the rich annotations of RIVAL10 allow for going beyond foreground or background ablation (see the appendix for a discussion of attribute removal). Ablation via graying can be thought of as another kind of noise, where all pixels are smoothed to $0.5$. In Figure \ref{fig:bg_ablated}, the left plot reveals that Robust ResNets and SimCLR see the largest drops in accuracy when evaluated on images with grayed backgrounds. Transformers do well on ablated images, consistent with the observation that transformers had high $RFS$ at the largest noise levels. Furthermore, when we attempt to fit a linear layer to classify background-ablated images, only the features from transformer models are sufficiently informative to have high linear classification accuracy. Thus, while transformers make use of backgrounds, they still retain significant foreground information in their feature space. This result suggests transformers are much more robust to localized distribution shift. That is, distribution shift in one region (the background) may affect model perception of other unperturbed regions much less in transformers than ResNets. 

% {\bf Figures (1.5 pages)} 
% \begin{itemize}
% \item Two subplots: (left) Accuracy vs. Noise level, separate lines for fg and bg noise. (right) Fg Noise Accuracy as a fraction of BG Noise Accuracy (higher means the relative sensitivity to fg noise over bg noise is low)

% \item Scatter of Acc under BG Noise vs. Acc under FG Noise; reflects general noise robustness (position along y=x) and fg sensitivity (position orthogonal to y=x). Marker size denoting parameter count, marker type denoting RN vs. Transformer

% \item Smoothed histograms showing difference in Pr[True Class] for representative sample of models of equal size. Highlighting classes that have high fg sensitivity (car, cat) and low fg sensitivity (ship, bird)

% \item Qualitative examples of surprising cases: where fg noise improves classification confidence ? (perhaps not necessary; might be cool to look into)

% \end{itemize}

\subsection{Saliency Alignment}
%%One pair, merged
\begin{figure}[ht]
\vspace{-0.3cm}
\centering
\includegraphics[width=0.45\textwidth]{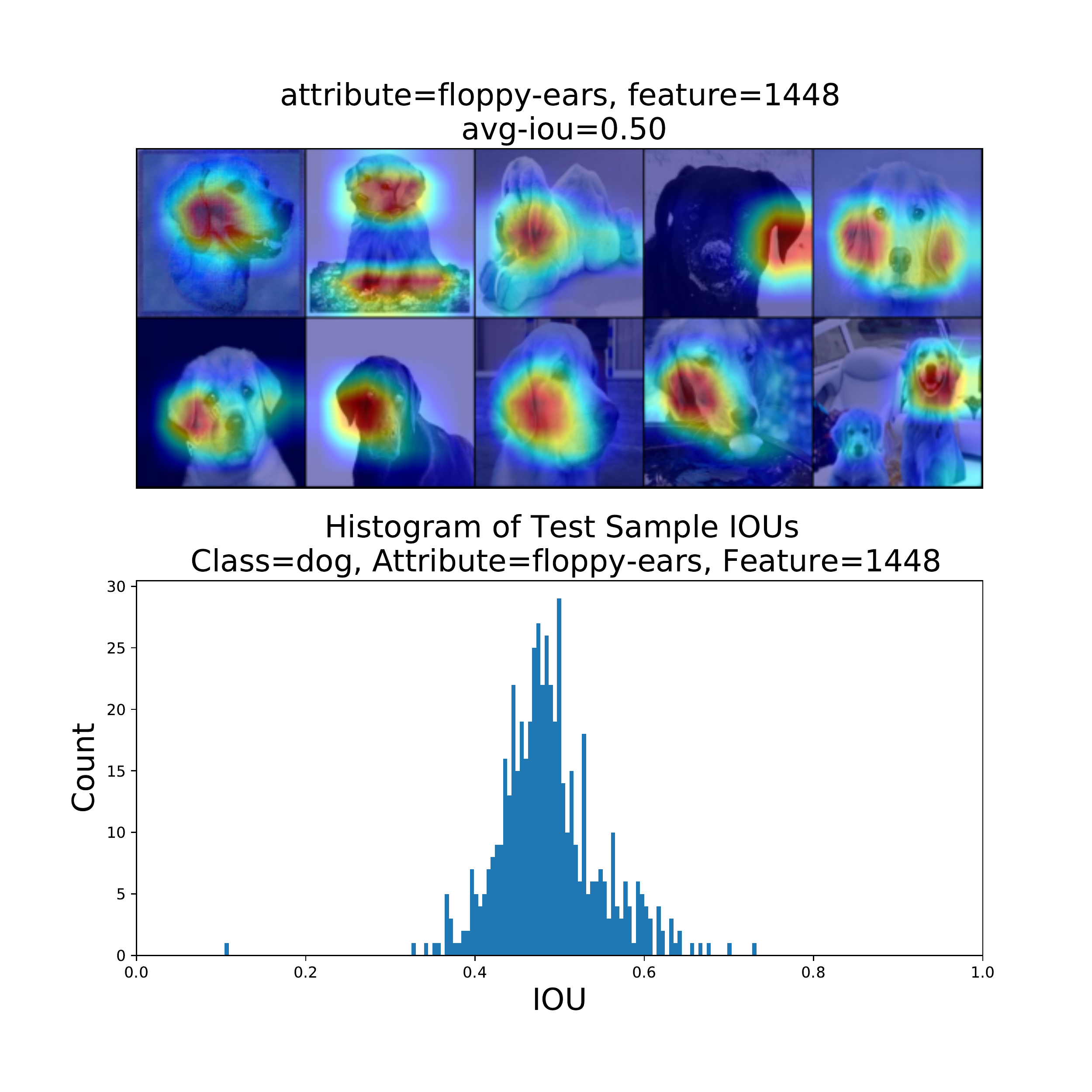}
\caption{(\textbf{Top}): Example GradCAMs on test images with respect to the top feature identified by IOU in training set.
(\textbf{Bottom}): Histograms of IOUs corresponding to this feature, attribute pair.}
\label{fig:top-gradcam-and-hist}
\vspace{-0.4cm}
\end{figure}

%\begin{figure}[h]
%\vspace{-0.2cm}
%\centering
%\includegraphics[width=0.45\textwidth]{figures/gradcam-and-hist_feat-vals_car_wheels_542_test.pdf}
%\caption{(\textbf{Top}): Example GradCAMs on test images with respect to the top feature identified by IOU in training set.
%(\textbf{Bottom}): Histograms of IOUs corresponding to this feature, attribute pair.}
%\label{fig:top-gradcam-and-hist-2}
%\vspace{-0.4cm}
%\end{figure}

To complement the noise analysis, we use GradCAM \cite{gradcam} to assess the amount of saliency that models place on foreground pixels. RIVAL10's object segmentations allow us to automatically quantify saliency alignment with foregrounds, removing the need for human inspections. We also can easily detect failure modes, where models deem background regions as highly salient, by sorting samples based on saliency alignment. We present several metrics to assess saliency alignment in the appendix. We find that extracting samples with the lowest difference in average pixel saliency in foreground and background yield the most interesting failure modes. We present examples selected this way in Figure \ref{fig:spurious_birds_ships}, highlighting spurious background features that contribute to the low $RFS$ of ships and birds observed across models in Figure \ref{fig:noise_hist_and_scatter}. Specifically, models look for water and coasts when classifying ships, and twigs and branches when classifying birds.
%In this section, we focus on $\delta$ Densities, which we define for a saliency map $\mathbf{s} \in [0,1]^d$ and binary object mask $\mathbf{m} \in \{0,1\}^d$ as 
%$$\Delta\text{-Density} = \frac{\mathbf{s} \cdot \mathbf{m}}{\mathbf{m} \cdot \mathbf{m}} - \frac{\mathbf{s} \cdot (\mathbf{1} - \mathbf{m})}{d - (\mathbf{m} \cdot \mathbf{m})}$$

In Figure \ref{fig:ious_sal_alignment}, we appeal to the standard metric intersection-over-union (IOU). Saliency maps are binarized using a threshold of $0.5$ before being compared with object segmentation masks. On average, saliency alignment is similar across models, despite their being large differences in $RFS$ identified in the noise analysis, suggesting saliency maps may give an incomplete picture of model sensitivity. In comparing saliency alignment across classes, we see much larger differences, emphasizing the result that {\bf class matters} when it comes to background reliance. 

\section{Neural Node Attribution Analysis}

The attribution of features in a neural network is a fundamental problem in modern machine learning work. Saliency, when computed with respect to a given feature, is a prominent approach for doing so \cite{gradcam, integrated-grads, erion2021, kim2018interpretability}. Although many works make claims of attribution based on saliency, to the best of our knowledge, quantitative validation is rarely given \cite{zhou-2021}. Here we propose to quantitatively evaluate node attribution via saliency through comparison with the ground-truth attribute localization in RIVAL10. 

We propose the following procedure. Given a pretrained robust ResNet50 feature extractor and a class label, we identify the top $10$ training images by activation with that label for each component in the feature layer (the penultimate layer). We then compute saliency using GradCAM at each neural feature on these top-10 images, and compare them against ground truth attribute localization. Saliencies are binarized at max-normalized threshold of $\tau=0.5$. The intersection-over-union (IOU) with the ground truth attribute localization is then computed for each sample, and finally averaged. This obtains a score, which we interpret as measuring the quality of neural feature attribution based on saliency alignment to the attribute segmentations of the top-10 images. We then select the neural feature with highest alignment per attribute, identifying these features as the best candidates for node attribution. Note that searching by top IOU is only possible with ground truth attributes and localizations, as is the case with RIVAL10.

\looseness=-1
Next, we check if these neural features generalize to held-out data not used in the analysis, namely the test set of RIVAL10. Here we analyze one class-attribute pair and show additional results in the appendix. We visualize the GradCAMs of top testing samples with respect to the top features identified in the training set in Figure \ref{fig:top-gradcam-and-hist}. We observe visually that the saliencies align well with the given attribute on these samples. We then compute the IOU scores on {\it all} images in the test set with the given class and attribute labels. We plot this histogram in Figure \ref{fig:top-gradcam-and-hist}.  We observe that IOU values are on average high ($>0.5$) indicating that the neural features generalize well to held-out data for considered cases. We note that this analysis is just one approach for quantitatively evaluating feature attribution. We stress the importance of quantitative measurements rather than relying on just visualization, and envision that our RIVAL10 dataset may help refine the discourse around feature attribution.

\section{Conclusion}
% \looseness=-1
We present {\bf RI}ch {\bf V}isual {\bf A}ttributes with {\bf L}ocalization (RIVAL10), and quantitatively assess sensitivities of state-of-the-art models under noise corruption. Specifically, we find adversarially or contrastively training ResNets leads to reduced relative foreground sensitivity. Further, we observe transformers to adaptively raise foreground sensitivity as noise level increases, while ResNets do not. Applying automated alignment metrics to saliency maps reveals instances of spurious background features used by models. Lastly, we observe promising evidence that neural node attributions based on top activating images generalize to instances unseen during attribution. We hope RIVAL10's rich annotations lead future studies to gain new quantifiable insights on the behavior of deep image classifiers. 

\section{Acknowledgements}
This project was supported in part by NSF CAREER AWARD 1942230, ONR grant 13370299, HR001119S0026, ARL grant W911NF2120076, and  AWS Machine Learning Research Award.

%%%%%%%%% REFERENCES

%\clearpage
% \newpage

{\small
\bibliographystyle{ieee_fullname}
\bibliography{PaperForReview}
}

\newpage
\appendix
\
\include{supplementary}

\end{document}

%% file: supplementary.tex
\def\cvprPaperID{9398} % *** Enter the CVPR Paper ID here
\def\confName{CVPR}
\def\confYear{2022}

% \begin{document}
\begin{appendix}
% \title{Supplementary: A Comprehensive Study of Image Classification Model Sensitivity to Foregrounds, Backgrounds, and Visual Attributes}

% \and
\maketitle
\begin{table*}[h!]
    \centering
    \begin{tabular}{|c|c|cc|cc|c|} \toprule
        RIVAL10 & Number of  & \multicolumn{4}{|c|}{ImageNet-1k Classes comprising RIVAL10 Class} & Positive\\
        Class & Instances & Class Name \#1 & WordNet ID \#1 & Class Name \#2 & WordNet ID \#2 & Attributions \\ \midrule
        Truck & $2523$& Moving Van & n03796401 & Semi & n04467665 & $13577$\\
        Car & $2665$&Waggon & n02814533 & Convertible & n03100240 & $9415$\\
        Plane & $2655$& Airliner & n02690373 & Military plane & n04552348 & $15277$\\
        Ship & $2660$& Ocean liner & n03673027 & Container vessel & n03095699 & $14122$\\
        Cat & $2667$& Persian cat & n02123394 & Egyptian cat & n02124075 & $9309$\\
        Dog & $2660$& Labrador retriever & n02099712 & Golden retriever & n02099601 & $11251$\\
        Equine & $2663$& Sorrel & n02389026 & Zebra & n02391049 & $13343$\\
        Deer & $2657$& Gazelle & n02423022 & Impala & n02422699 & $12274$\\
        Frog & $2667$& Tailed Frog & n01644900 & Tree-frog & n01644373 & $5317$\\
        Bird & $2667$& Goldfinch & n01531178 & Housefinch & n01532829& $8822$ \\ \midrule
        \multicolumn{7}{|c|}{Total: $\mathbf{26,484}$ instances ($21,178$ train, $5,308$ validation) with $\mathbf{112,707}$ positive attributions ($\sim 4.26$ per image)} \\
        \bottomrule
    \end{tabular}
    \caption{Breakdown of RIVAL10 dataset. Corresponding ImageNet-1k classes listed. }
    \label{tab:rival10_breakdown}
\end{table*}

\section{Additional Details on RIVAL10}
We present a full breakdown of the RIVAL10 dataset in this section. RIVAL10 consists of ImageNet-1k samples organized into the classes of CIFAR10. Each RIVAL10 class is comprised of the training and validation samples drawn from two ImageNet-1k classes. In table \ref{tab:rival10_breakdown}, we present the ten classes of RIVAL10, along with the two corresponding ImageNet-1k classes per class. 

\begin{figure*}[h!]
    % \newsubfloat{
        \includegraphics[width=0.85\linewidth]{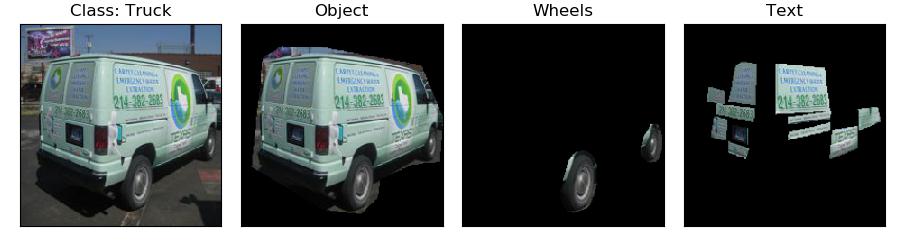}
        % }
    % \newsubfloat{
      \includegraphics[width=0.85\linewidth]{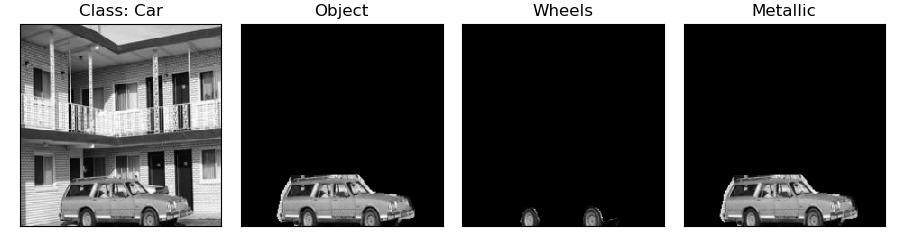}
    %   }
    % \newsubfloat{
      \includegraphics[width=0.85\linewidth]{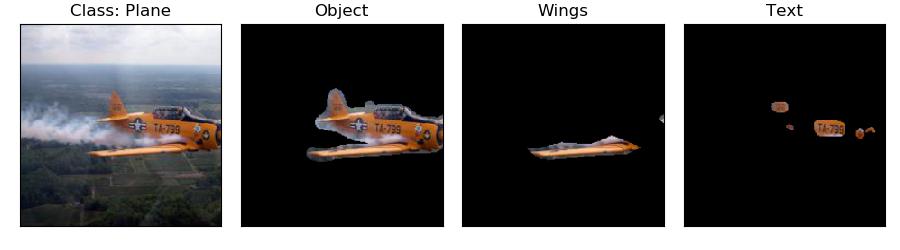}
    %   }
    % \newsubfloat{
        \includegraphics[width=0.85\linewidth]{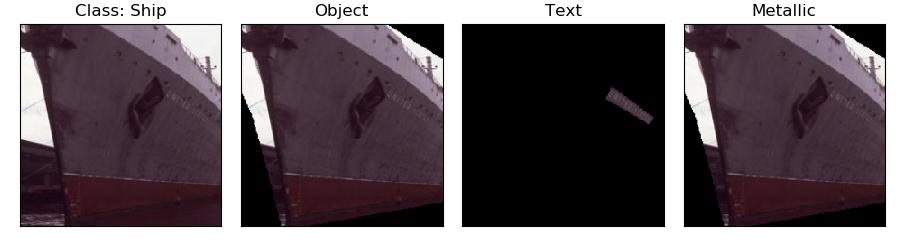}
        % }
    % \newsubfloat{
      \includegraphics[width=0.85\linewidth]{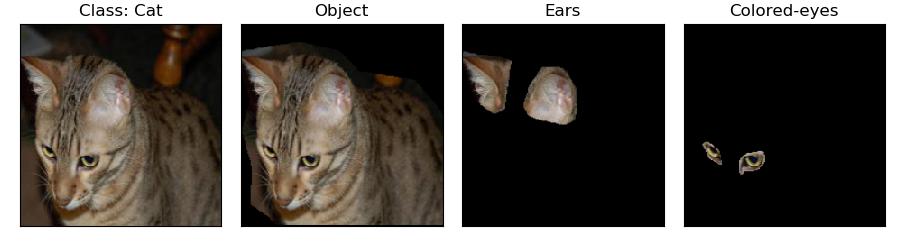}
    %   }
    \centering
    \caption{RIVAL10 examples. Left column has original image. Next column shows object mask applied onto the original image. The following two columns show attribute masks applied onto the original image.} 
    \label{fig:rival10_egs1}
\end{figure*}

\begin{figure*}[h!]
    % \newsubfloat{
      \includegraphics[width=0.85\linewidth]{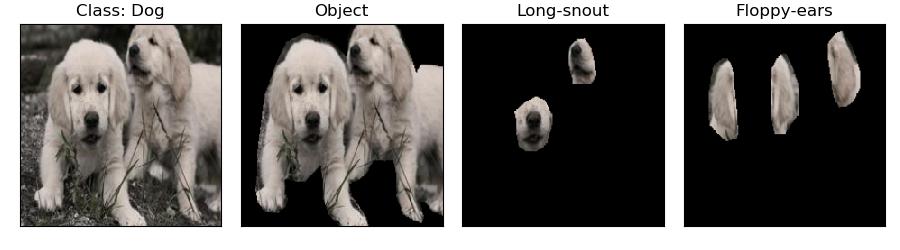}
    %   }
    % \newsubfloat{
        \includegraphics[width=0.85\linewidth]{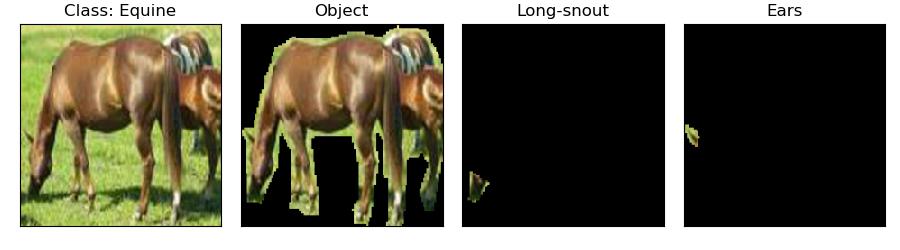}
        % }
    % \newsubfloat{
      \includegraphics[width=0.85\linewidth]{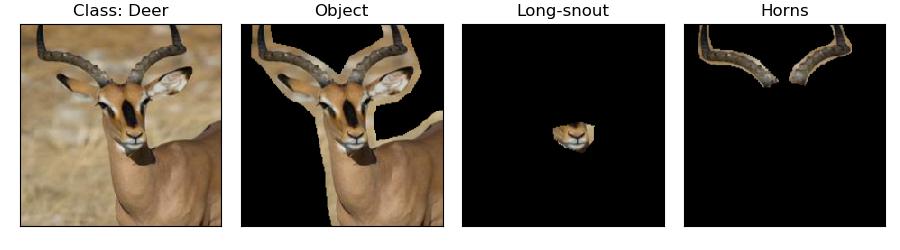}
    %   }
    % \newsubfloat{
      \includegraphics[width=0.85\linewidth]{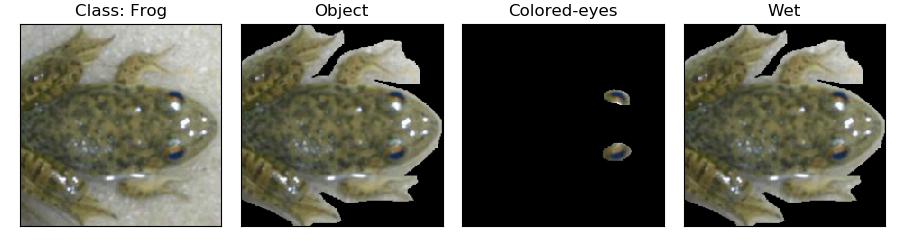}
    %   }
    % \newsubfloat{
      \includegraphics[width=0.85\linewidth]{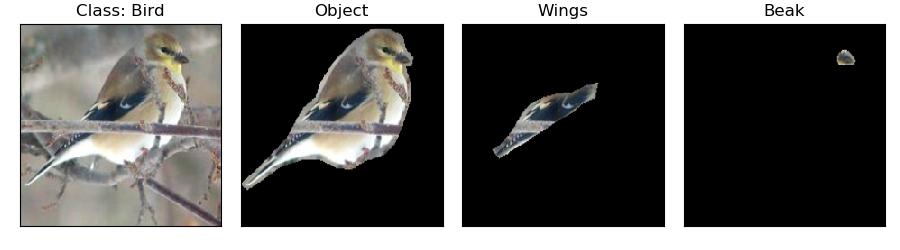}
    %   }
    \centering
    \caption{RIVAL10 examples. Left column has original image. Next column shows object mask applied onto the original image. The following two columns show attribute masks applied onto the original image.} 
    \label{fig:rival10_egs2}
\end{figure*}

In Figures \ref{fig:rival10_egs1} and \ref{fig:rival10_egs2}, we present representative examples drawn at random from the dataset, along with localized attribution. Every sample has a class label and complete binary labels for 18 attributes. That is, all positive instances of attributes are marked. This differs from the partial-label setting which is common in attribute learning. Further, for every positive instance of an attribute, a segmentation mask is provided, as well as a segmentation mask for the entire object for every sample. The figures show the object mask and two positive attribute masks per image via applying the mask to the image; that is, taking the elementwise product of the segmentation mask and the image, so to black out any pixels outside of the segmentation mask. 

\begin{figure}
    \centering
    \includegraphics[width=0.9\linewidth]{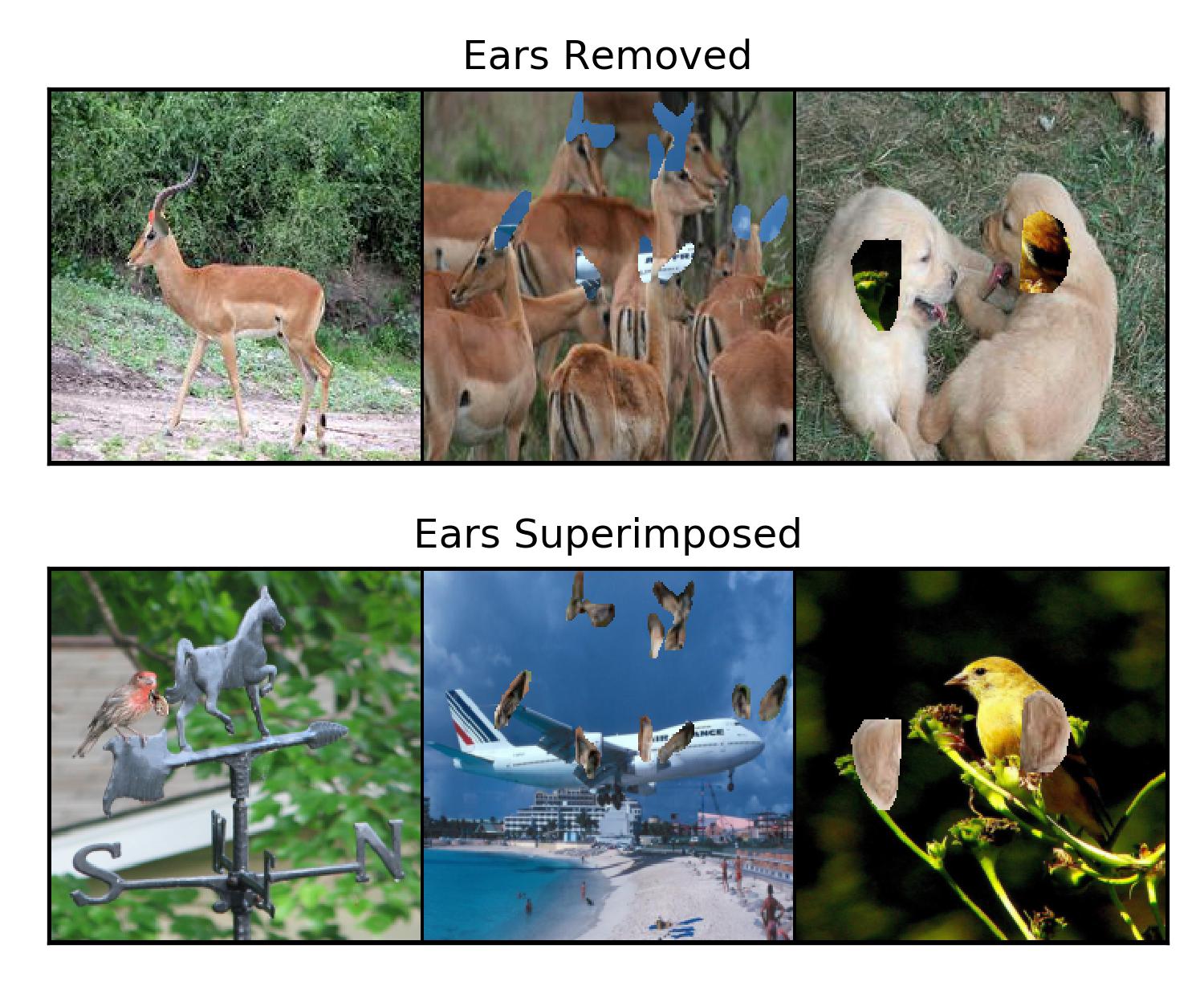}
    \caption{Examples of attribute-swapped inputs.}
    \label{fig:attr_swapped}
\end{figure}

We note that for the attributes {\it metallic, hairy, wet, tall, long, rectangular}, and {\it patterned}, we use the entire-object mask as the attribute segmentation, as these attributes pertain to the entire object. Segmentation masks can be leveraged to create many variants of RIVAL10. In Figure \ref{fig:attr_swapped}, we display examples of challenging inputs yielded via attribute removal and insertion.  

\section{Additional Details on Data Collection}
\begin{figure}
    \centering
    \includegraphics[width = \linewidth]{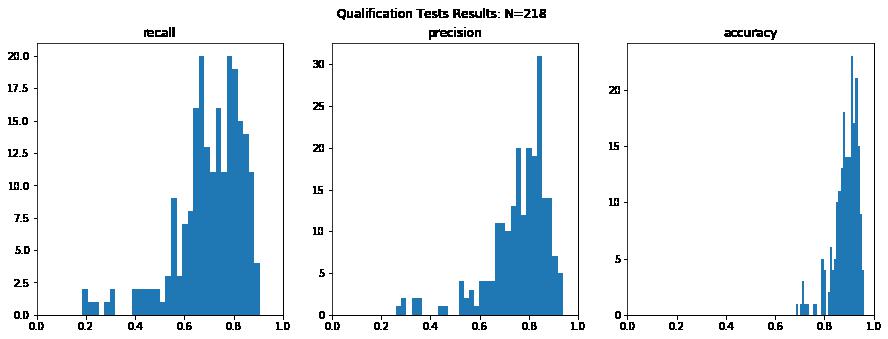}
    \caption{Histograms of Worker recall, precision, and accuracy scores on the qualification exam. }
    \label{fig:qual_metrics}
\end{figure}
\begin{figure}
    \centering
    \includegraphics[width = \linewidth]{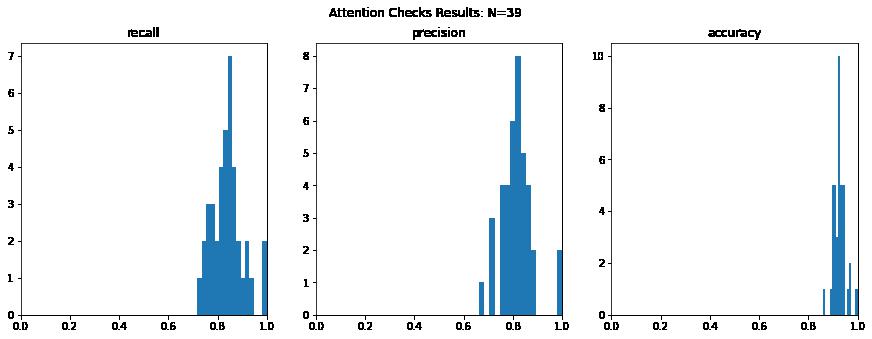}
    \caption{Histograms of per-Worker recall and precision on randomly placed attention checks during the main phase of data collection.}
    \label{fig:attention_check_metrics}
\end{figure}

\textbf{Worker Pool}: We selected workers from the US to promote English fluency, which is necessary for reading the instructions.  We also selected workers who have completed $>95\%$ of their tasks to further promote successful task completion. 

\textbf{Worker Payment}: Each task of 20 images was estimated to take 10 minutes.  We set a rate of \$1.50 per task, which amounts to \$9.00 / hour, which is ~25\% above the US Federal Minimum Wage (\$7.25) at time of this writing. In the second phase of collection, workers were compensated at a rate of \$0.1 per segmentation. We estimate one segmentation to take 30-45 seconds on average, which amounts to a wage of \$9-12 an hour.

\textbf{Qualification Exam}: As discussed in the main text we required workers to pass a qualification exam for access to the main phase of data collection. The qualification exam consisted of $20$ images with ground-truth annotations which we defined. Workers were asked to read the instructions carefully and complete the exam. We then computed precision, recall, and accuracy metrics on these questions. A total of $218$ workers took the exam. All workers were paid a $\$1.50$ for the exam, regardless if they passed or not. We report the distribution of worker scores in Figure \ref{fig:qual_metrics}.  

We use these distributions to inform a chose of threshold for passing the exam, where the two relevant decision factors are (1) high bar for metrics to promote annotation quality (2) a large pool of workers for higher rate of data collection.  We note that since attributes are sparse, accuracy is not a good metric for distinguishing worker performance. This can be seen in the concentration of values in Figure \ref{fig:qual_metrics} (right). We found that $90$ workers scored greater than or equal to $0.75$ in precision and recall {\it jointly}, and decided to use this as our threshold. Of the workers who completed the qualifying exam, $N=39$ contributed to the main phase. The number of annotations completed by each worker varied (min: $20$, max: $1000$).

\textbf{Attention Checks}: We additionally measure worker performance during the main phase of data collection through attention checks. Overall, $4\%$ of samples to annotate had ground truth annotations completed by the authors. This allows us to estimate worker quality during the main phase, and ensure that worker attention is maintained. The ground truths for these attention checks were collected from a pool of trusted CS graduate students. 

Overall metrics on these attention checks were similar to threshold set for the qualification exam: the average precision and recall {\it across workers} were $0.81$ and $0.84$ respectively. We report these per-worker metrics in Figure \ref{fig:attention_check_metrics}. 

\begin{figure*}
\centering
\includegraphics[width=\linewidth]{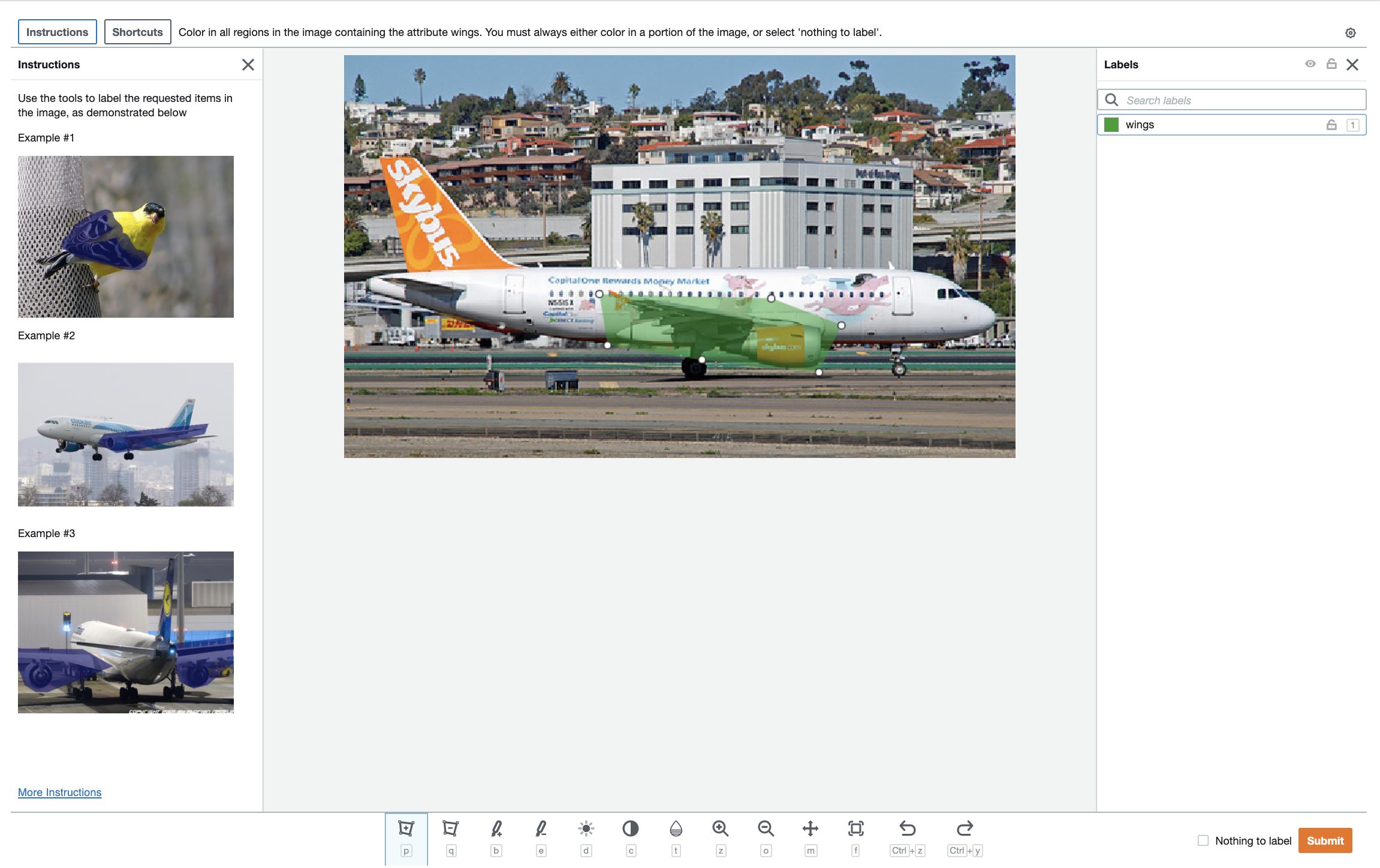}
\caption{Screenshot of annotation form and tools for completing segmentations.}
\label{fig:mturk_seg}
\end{figure*}

\textbf{Collection of Segmentations}: In a second pass, workers submitted segmentation masks for any attribute positively annotated previously. Workers had access to many tools to complete segmentations, including zooming, a polygon tool, and a brush. Detailed example segmentations were provided per attribute. Figure \ref{fig:mturk_seg} shows a screen shot of the segmentation platform.  A similar qualification check was administered before the second phase of data collection, with a minimum average IOU of $0.7$ required on at least five segmentations. Also, an average IOU of $0.745$ was achieved on attention checks.

\begin{figure}
    \centering
    \includegraphics[width = \linewidth]{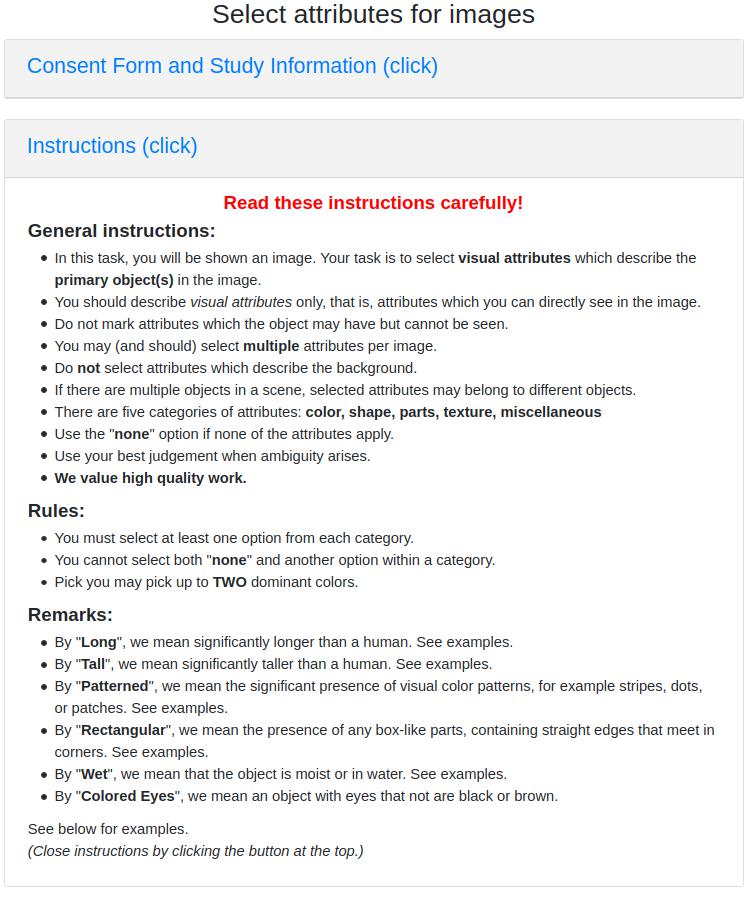}
    \caption{Screenshot of instructions page shown to workers}
    \label{fig:screenshot_instructions}
\end{figure}

\begin{figure}
    \centering
    \includegraphics[width = \linewidth]{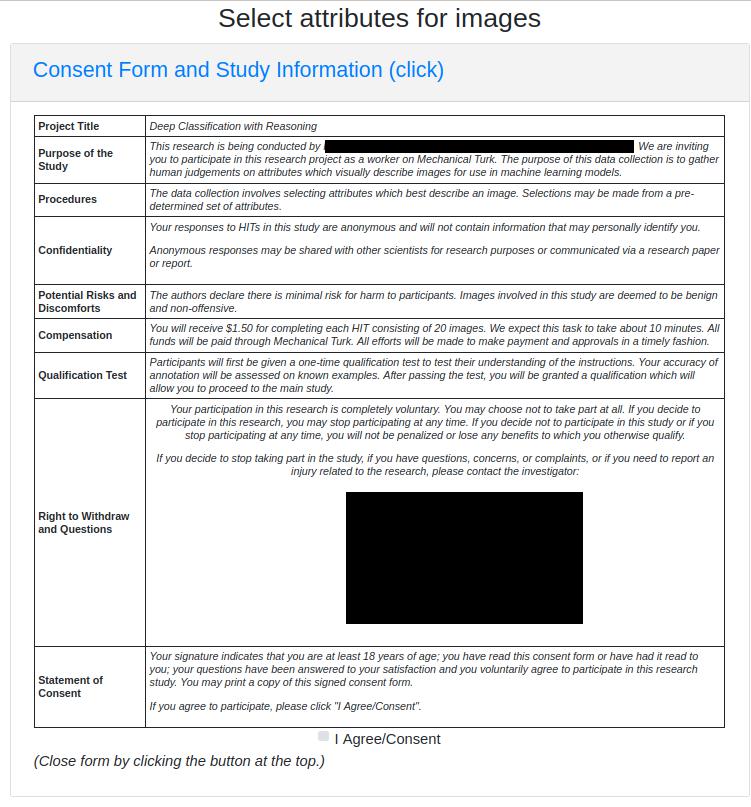}
    \caption{Screenshot of consent page shown to workers}
    \label{fig:screenshot_consent}
\end{figure}

\begin{figure}
    \centering
    \includegraphics[width = \linewidth]{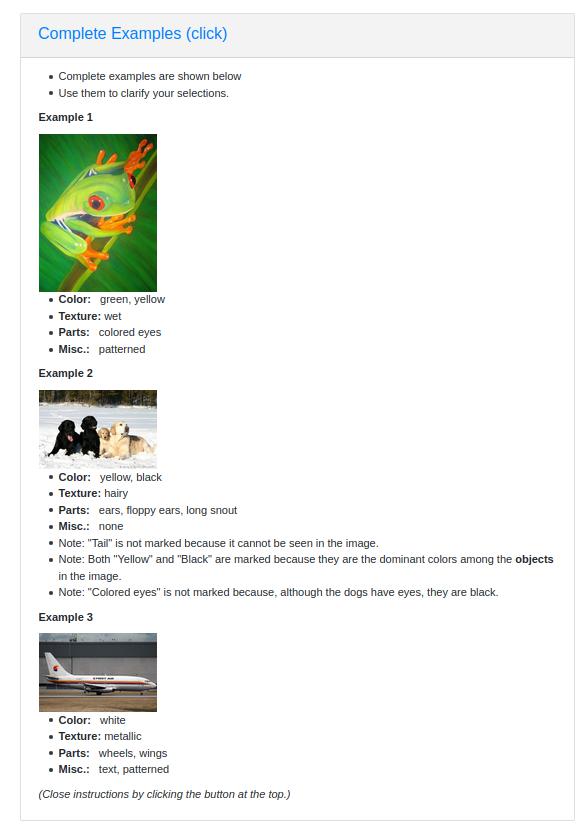}
    \caption{Screenshot of examples page shown to workers}
    \label{fig:screenshot_examples}
\end{figure}

\begin{figure}
    \centering
    \includegraphics[width = \linewidth]{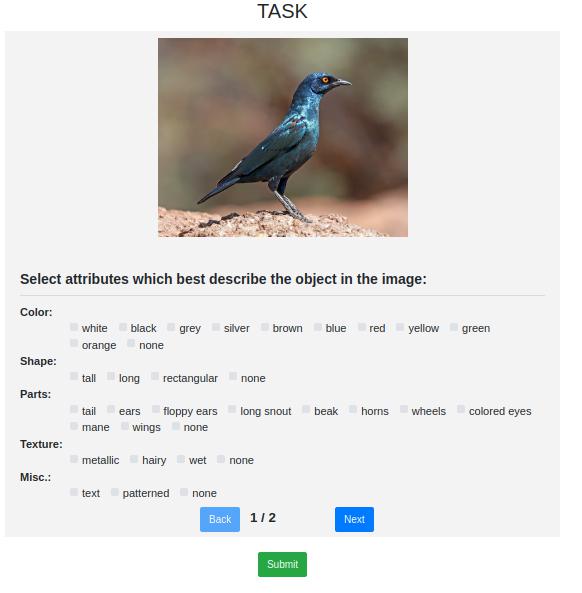}
    \caption{Screenshot of annotation form shown to workers}
    \label{fig:screenshot_anno}
\end{figure}

\textbf{Screenshots of Instructions Given to Workers}: We show screenshots of the instructions, consent form,  examples, and annotation form in Figures \ref{fig:screenshot_instructions}, \ref{fig:screenshot_consent}, \ref{fig:screenshot_examples}, and \ref{fig:screenshot_anno} respectively. We have redacted identifying information of the authors appropriately.

\section{Model Details}

\begin{table*}[ht]
    \centering
    \begin{tabular}{|c|cccccc|}\toprule
        \multirow{2}{*}{Model} & Pretraining & Parameter & RIVAL10 & Source of & Original & \multirow{2}{*}{Notes} \\
        & Set & Count & Accuracy  & Weights& Paper & \\
        \midrule
         ResNet18  &  \multirow{4}{*}{IN-1k} &11.4M  &  95.48  &  \multirow{4}{*}{\cite{NEURIPS2019_9015}}  & \multirow{4}{*}{\cite{resnets}}&\\
         ResNet50  &&  23.9M  &  99.10  & & &\\
         ResNet101  &&  42.8M  &  99.21  & &&\\
         ResNet152  &&  58.5M  &  99.43  & &&\\ \hline
         Robust ResNet18  & \multirow{4}{*}{IN-1k}& 11.4M  &  91.80  &   \multirow{4}{*}{\cite{robustness}} & \multirow{4}{*}{\cite{pgd}} & $\ell_2$-PGD, $\epsilon=3.0$\\
         Robust ResNet50  &&  23.9M  &  93.82  &&& $\ell_2$-PGD, $\epsilon=3.0$\\ 
         Robust ResNet18$^\dagger$  &&  11.4M  &  93.69  &&& $\ell_2$-PGD, $\epsilon=1.0$\\
         Robust ResNet50$^\dagger$  &&  23.9M  &  97.29  & & & $\ell_2$-PGD, $\epsilon=1.0$\\ \hline
         SimCLR  & IN-1k&  23.9M  &  93.87  & \cite{bolts} & \cite{SimCLR} & RN50 backbone\\ \hline
         CLIP ResNet50  & \multirow{4}{*}{YFCC100M} &  23.9M  &  96.34  &  \multirow{4}{*}{\cite{clip}} &\multirow{4}{*}{\cite{clip}} &\\
         CLIP ResNet101  &&  42.8M  &  96.27  & &&\\ 
         CLIP ViT-B/16  &&  86M  &  99.17  & &&Patch=$16\times16$\\
         CLIP ViT-B/32  &&  87M  &  98.44  & & &Patch=$32\times32$\\ \hline
         ViT (Tiny)  &  \multirow{5}{*}{IN-21k + IN-1k}&5M  &  94.82  &   \multirow{5}{*}{\cite{timm_library}} & \multirow{5}{*}{\cite{vits}} & Patch=$16\times16$\\
         ViT (Small)  &&  22M  &  98.96  &    & & Patch=$16\times16$\\
         ViT (Base)  &&  86M  &  99.64  &    &  & Patch=$16\times16$\\
         ViT (Small)$^\dagger$  &&  23M  &  97.86  &    & & Patch=$32\times32$\\
         ViT (Base)$^\dagger$  &&  87M  &  99.26  &    &  & Patch=$32\times32$\\ \hline
         DeiT (Tiny)  &  \multirow{3}{*}{IN-1k}&5M  &  96.42  &  \multirow{3}{*}{\cite{timm_library}}  & \multirow{3}{*}{\cite{deit}}& Patch=$16\times16$ \\
         DeiT (Small)  &&  22M  &  99.30  &    & & Patch=$16\times16$\\
         DeiT (Base)  &&  86M  &  99.74  &    & & Patch=$16\times16$ \\ \bottomrule
    \end{tabular}
    \caption{Details on all models analyzed. $^{\dagger}$ denotes models that were only considered in specific ablations (i.e. not present in main figures). IN refers to ImageNet. }
    \label{tab:models}
\end{table*}

Our experiments included a diverse set of model architectures and training paradigms. A primary challenge of our work was facilitating fair comparisons across models that operate very differently from one another at train and test time. In this section, we provide greater discussion on the differences among the models and their affect on our analysis.

\subsection{Architectures and Training Procedures}

Architecturally, we focus on ResNets\cite{resnets} and Transformers\cite{vits}. Both architectures are deep, consisting of many layers, though the nature of layers are markedly different. ResNets rely on convolutions, which introduce the spatial inductive biases such as translational invariance. Transformers, on the other hand, view an image as a collection of patches, an apply attention layers to allow distant patches to effect one another. Thus, images are processed significantly differently across the two architectures. However, seeing as both architectures are used in image classification, comparisons are warranted and necessary. Other works also compare transformers and ResNets, as mentioned in the main text. 

Among training procedures, most models seek to minimize cross entropy loss, using single class-label supervision on clean training samples. Robust ResNets instead undergo adversarial training\cite{pgd}, which replaces clean training samples with adversarially attacked ones. These models are then robust in the sense that they admit far fewer adversarial examples, where imperceptible perturbations cause models with high clean accuracy to badly misclassify attacked inputs. 

We also consider contrastively trained models, which differ dramatically in that they do no use class-labels during training. The contrastive loss refers to training encoders to draw representations of similar inputs close to one another, while simultaneously pushing representations of different inputs apart. In SimCLR\cite{SimCLR}, two views of a single input are created via data augmentation. In CLIP\cite{clip}, the representation of an image is contrastively drawn to the representation of a corresponding {\it text} caption, obtained using two separate encoders (image and text) that share a latent space, remarkably extending contrastive learning to multiple encoders operating on different mediums. 
Notice that neither SimCLR nor CLIP has the exclusive objective of image classification, like the other supervised models we study. Instead, they seek to learn informative representations, which can then be used for a variety of downstream tasks. However, object recognition is one of the main downstream task considered, and it is by no means abnormal to finetune SimCLR or CLIP encoders to perform image classification. We note that CLIP models have also been shown to have impressive zero-shot classification abilities. We leave investigation of CLIP's zero-shot classification to future work. 

\subsection{A Single Test Environment}

Given that models differ in their training algorithms and settings, we seek to create a single testing environment that preserves feature spaces learned in pretraining. Simply, we isolate feature extractors, usually by removing the final classifying layer (if present). We then fit a linear layer atop the fixed features via supervised training on RIVAL10. Specifically, we use an Adam optimizer with learning rate of $1e^{-4}$, betas of $0.9, 0.999$, and weight decay of $1e^{-5}$, for ten epochs. When finetuning on background ablated images, we allow for an additional ten epochs. As seen in table \ref{tab:models}, all models achieve over $90\%$ test accuracy using our simple finetuning process. We do not wish to compare model accuracies, though we argue that high accuracies across the board show that no model is significantly disadvantaged with respect to its classification ability. 

\subsection{Other Factors of Variation}
\begin{figure*}
\begin{minipage}{0.33\textwidth}
\centering
\includegraphics[width=0.9\linewidth]{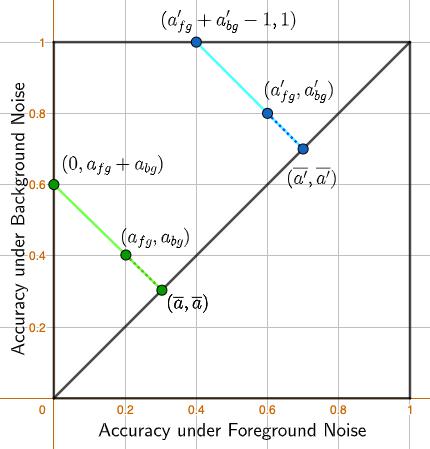}
\end{minipage}
\begin{minipage}{0.65\textwidth}
\centering
\includegraphics[width=\linewidth]{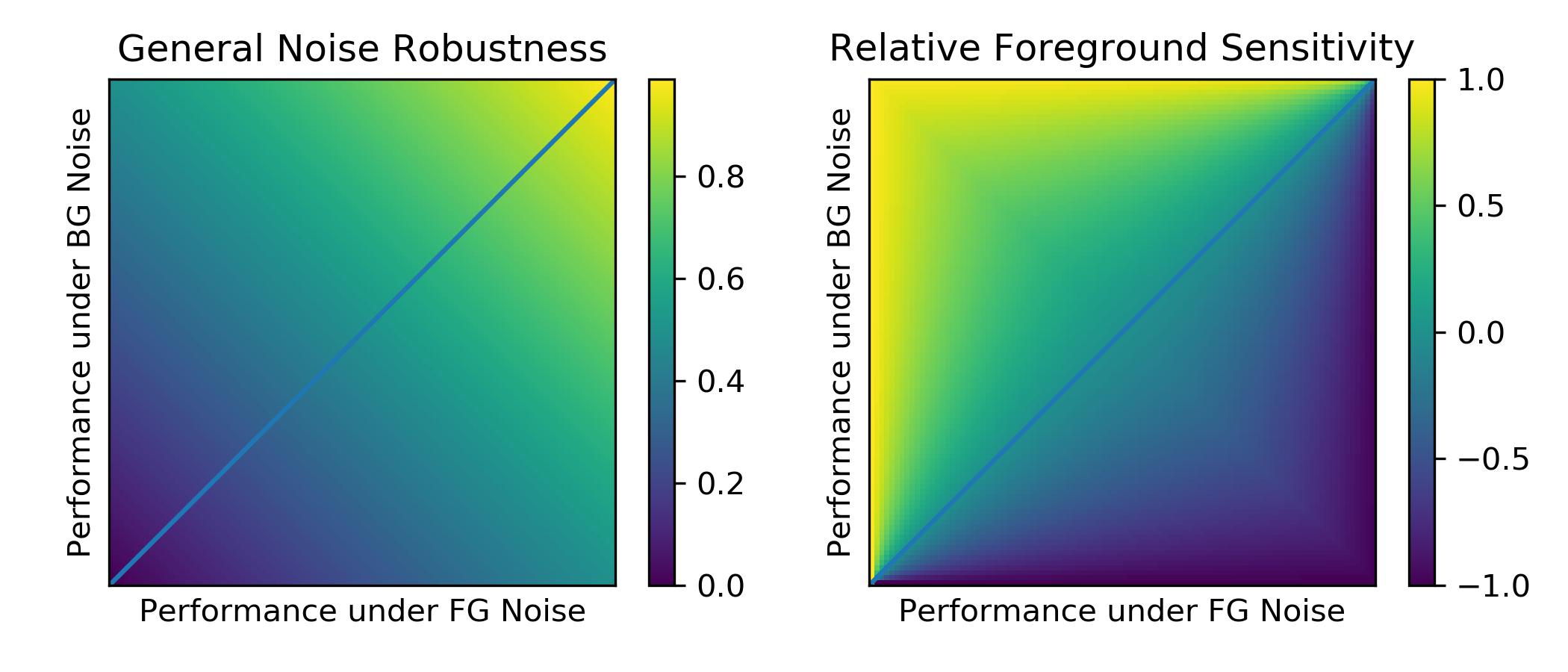}
\end{minipage}
    % \centering
    % \includegraphics[width=0.9\linewidth]{latexfigures/rfs_derivation.jpg}
    \caption{({\bf Left}) We demonstrate how $RFS$ is derived as a ratio of the distance of $(a_{fg}, a_{bg})$ from the diagonal over the maximum distance to the diagonal for a point with fixed noise robustness $\overline{a} = 1/2 (a_{fg} + a_{bg})$. ({\bf Right}) Visualization of general noise robustness and relative foreground sensitivity for all points in the unit square. Moving along the main diagonal increases general noise robustness, and moving away (above) increases relative foreground sensitivity.}
    \label{fig:rfs_deriv}
\end{figure*}

Differences in network size and pretraining set, listed in table \ref{tab:models}, are two other significant factors of variation across the models we compare. Most models only use ImageNet-1k as the pretraining set. ViTs and CLIP models use larger datasets. While this is not ideal, differences are unavoidable in any comparison, and we argue that the pretraining sets fundamentally inform the models themselves, similar to how architecture and training procedure do. In the case of ViTs, we also consider DeiTs, which are only trained on Imagenet-1k, allowing for direct inspection of the effect of the larger pretraining set on transformer behavior.   

As for varying network sizes, we take multiple measures to paint a full picture. First, we take models of varying size within each category of interest. We find that across model types, larger networks achieve higher accuracies for clean and noisy samples. Our primary metric ($RFS$), however, normalizes for general noise robustness. Secondly, for all model types aside form CLIP ViTs, we include an instance with roughly 23M parameters. When only comparing these models, the same trends emerge. 

\section{$RFS$ and other Normalizations}
We propose relative foreground sensitivity ($RFS$) as a normalized measure to directly compare the sensitivities of models with varying general noise robustness. In this section, we expand on the derivation of $RFS$, and present results using $L_2$ normalized noise.

\subsection{Geometric Derivation of $RFS$}

Recall that the founding logic of our sensitivity analysis is that a model's sensitivity to a region can be measured by the degradation in performance due to noise corruption of that region. However, models with greater general robustness to noise will see lesser degradation due to noise in either region. Similarly, models with low noise robustness may see severe degradation due to noise in both regions. $RFS$ is designed to normalize against variance in general noise robustness, yielding a single measure to compare various models across.

In figure \ref{fig:rfs_deriv} (left), we consider a point with accuracies $a_{fg}, a_{bg}$ under foreground and background noise respectively. Further, we assume $\overline{a} = 1/2(a_{fg}+a_{bg}) \leq 0.5$ and $a_{fg} < a_{bg}$. Now, the distance from $(a_{fg}, a_{bg})$ to the diagonal (dashed green) is equal to the distance to $(\overline{a}, \overline{a})$, which amounts to $$\text{Distance to Diagonal} = \sqrt{2}(\overline{a} - a_{fg}) = \frac{\sqrt{2}(a_{bg} - a_{fg})}{2}$$
The maximum distance from the diagonal for a point with general noise robustness $\overline{a}$ then corresponds to the length of the green segment (solid and dashed). Here, the limiting factor is that $a_{fg} \geq 0$. This distance is 
$$\text{Max Distance to Diagonal} = \sqrt{2}(a_{fg}+a_{bg} - \overline{a}) = \sqrt{2}\overline{a}$$
Thus, $RFS = \frac{\sqrt{2}/2(a_{bg} - a_{fg})}{\sqrt{2}\overline{a}} = \frac{a_{bg}-a_{fg}}{2\overline{a}}$ when $\overline{a} \leq 0.5$. 

Now, we consider a point $(a_{fg}', a_{bg}')$ with $\overline{a'} = 1/2(a_{fg}'+a_{bg}') > 0.5$. The distance to the diagonal (dashed blue) is identical to the first case. Here, the maximum distance from the diagonal (full blue segment) is limited by the fact that $a_{bg}' \leq 1$. This yields
$$\text{Max Distance to Diagonal} = \sqrt{2}(1-\overline{a'})$$
leading to a final $RFS$ of $\frac{a_{bg}'-a_{fg}'}{2(1-\overline{a'})}$ when $\overline{a'} > 0.5$. Combining these cases gives the general formula for $RFS$.
$$RFS = \frac{a_{bg} - a_{fg}}{2\min(\overline{a}, 1-\overline{a})}$$

Intuitively, $RFS$ measures the gap in accuracy under background and foreground noise under a normalization. The normalization is designed to account for the fact that models with very high or very low noise robustness will be limited in the maximum gap attainable. In Figure \ref{fig:rfs_deriv}, we visualize both general noise robustness and $RFS$ for all accuracies under foreground and background noise to add further context. 
\begin{figure}[ht]
    \centering
    \includegraphics[width=\linewidth]{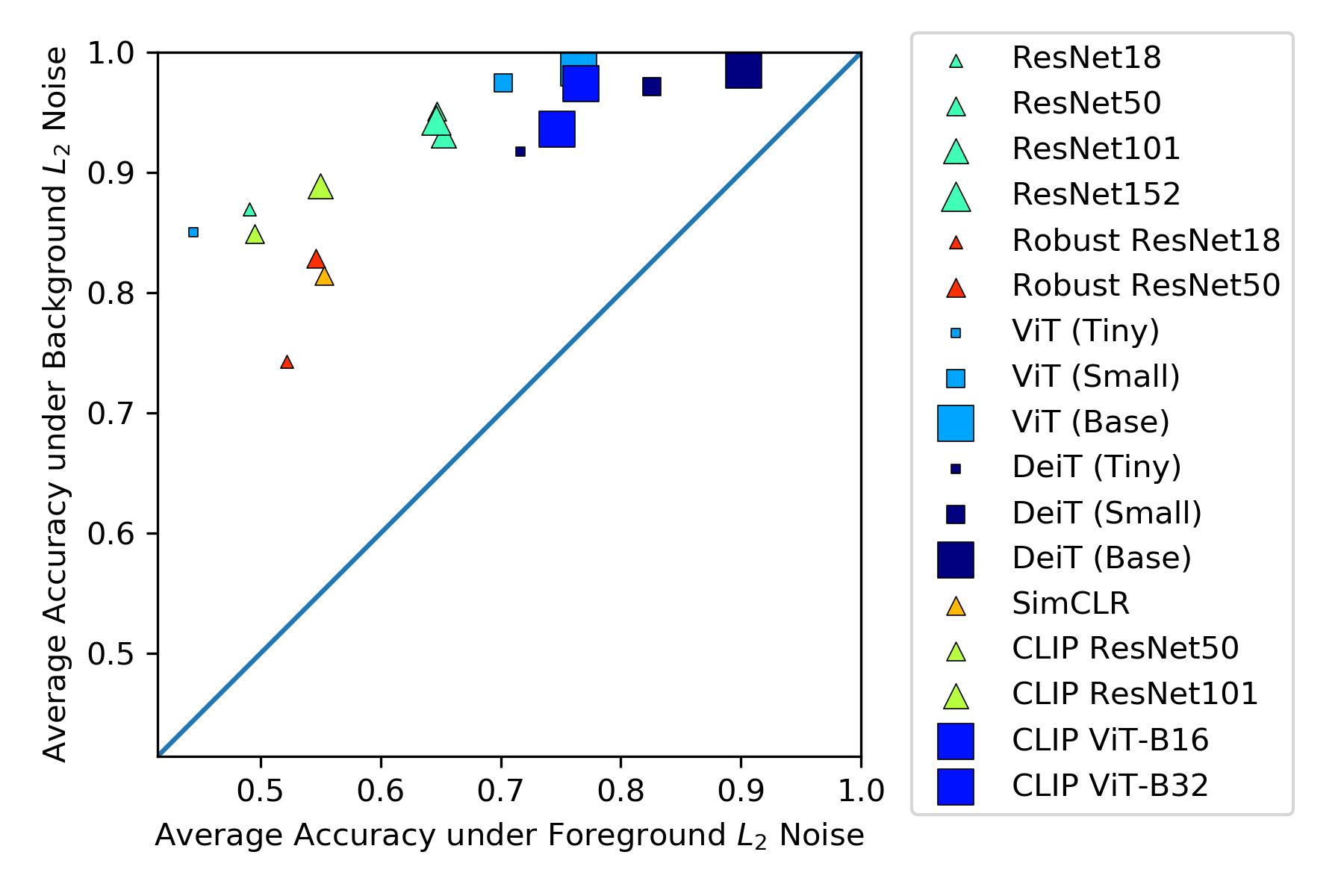}
    \caption{Accuracy under $\mathbf{L_2}$ normalized noise averaged over multiple noise levels. Marker size is proportional to parameter count. Models with higher relative foreground sensitivity lie further from the diagonal.}
    \label{fig:l2_scatter_fg_bg}
\end{figure}

\begin{figure*}[h!]
    \centering
    \includegraphics[width=0.9\textwidth]{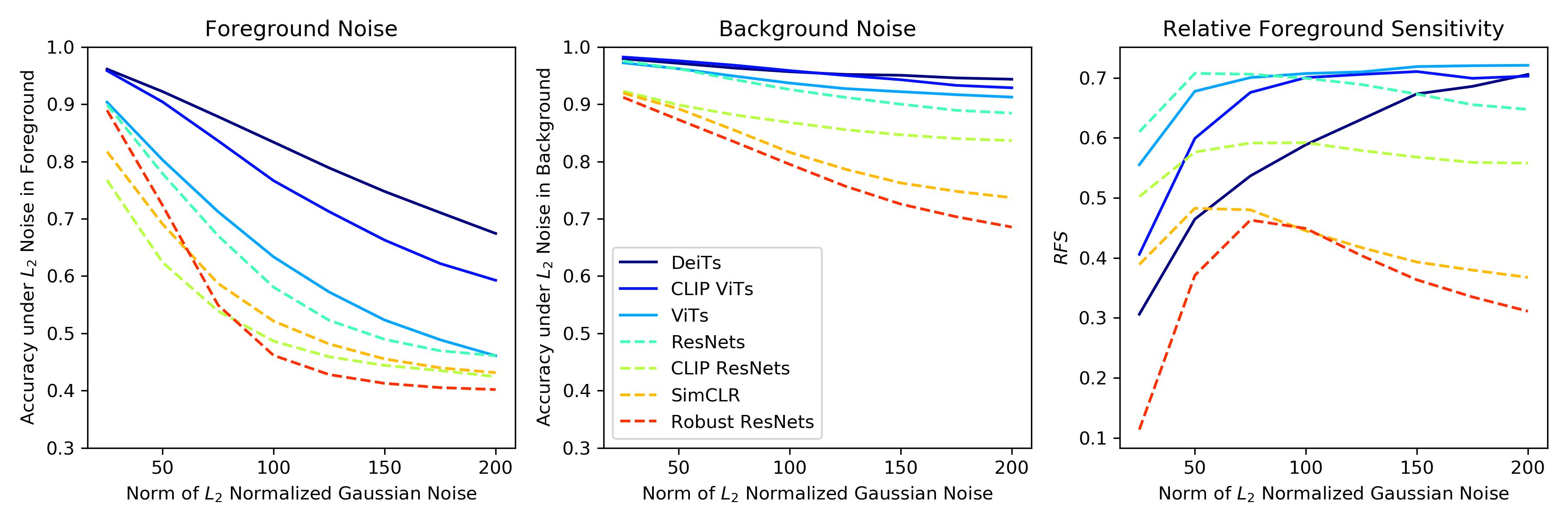}
    \caption{Accuracy under $\mathbf{L_2}$ normalized  noise in foreground ({\bf left}) and background ({\bf middle}) at various noise levels. Models are grouped by architecture and training procedure, with a curve corresponding to the average over all models in a group. ({\bf Right}): $RFS$ by group.}
    \label{fig:l2_accs_over_varying_noise}
\end{figure*}

\begin{figure*}[h!]
    % \newsubfloat{
    % \begin{subfigure}[]
        \includegraphics[width=\linewidth]{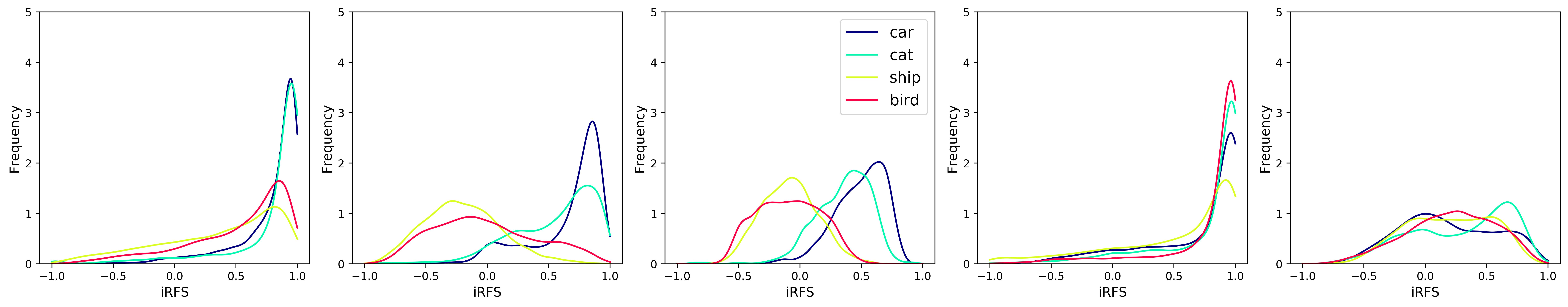}
        % }
    % \end{subfigure}
    % \newsubfloat{
    % \begin{subfigure}[]
      \includegraphics[width=\linewidth]{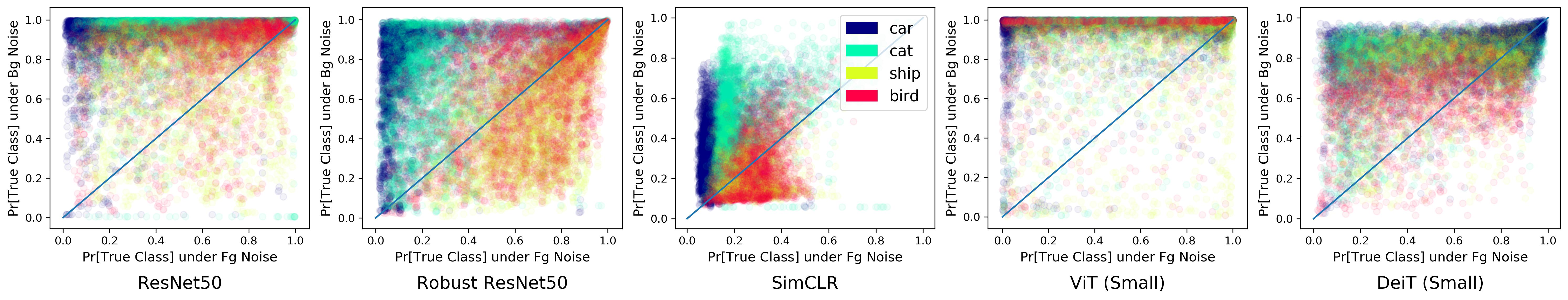}
    %   }
    % \end{subfigure}
    \centering
    \vspace{-0.5cm}
    \caption{Relative foreground sensitivity per instance for four classes and five models of roughly equal size, computed using $\mathbf{L_2}$ normalized noise corruption. ({\bf Top}): Histogram of $iRFS$; positive denotes greater foreground sensitivity. ({\bf Bottom}): Scatter; top left indicates high relative foreground sensitivity. Class distinction is slightly less pronounced than with $L_\infty$ noise, but still substantial.}
    \label{fig:l2_noise_hist_and_scatter}
\end{figure*}
\subsection{Results under $L_2$ Normalization of Noise}

We now reproduce the major figures from our noise analysis under $L_2$ normalized noise. We consider eight equally spaced noise levels, with $L_2$ norms ranging from 25 to 200. We find that the trends are near identical. The one small difference is that the class distinctions in Figure \ref{fig:l2_noise_hist_and_scatter} are slightly less severe. In particular, for DeiTs, the distribution of $iRFS$ scores on birds is roughly the same as that on ships. Recall that applying equal $L_\infty$ noise to two regions will incur a greater perturbation to the larger region when measuring under the $L_2$ norm. Thus, $L_\infty$ noise could introduce a bias where larger regions are corrupted more. The direction of this bias is unclear though, as relative sizes of foregrounds and backgrounds vary. Our corroborated results under $L_2$ normalized noise suggest that the aforementioned bias has little effect on our conclusions. 

\section{Saliency Alignment}

\begin{figure*}
    \centering
    \begin{minipage}{.195\textwidth}
    \centering
    \includegraphics[width=\linewidth]{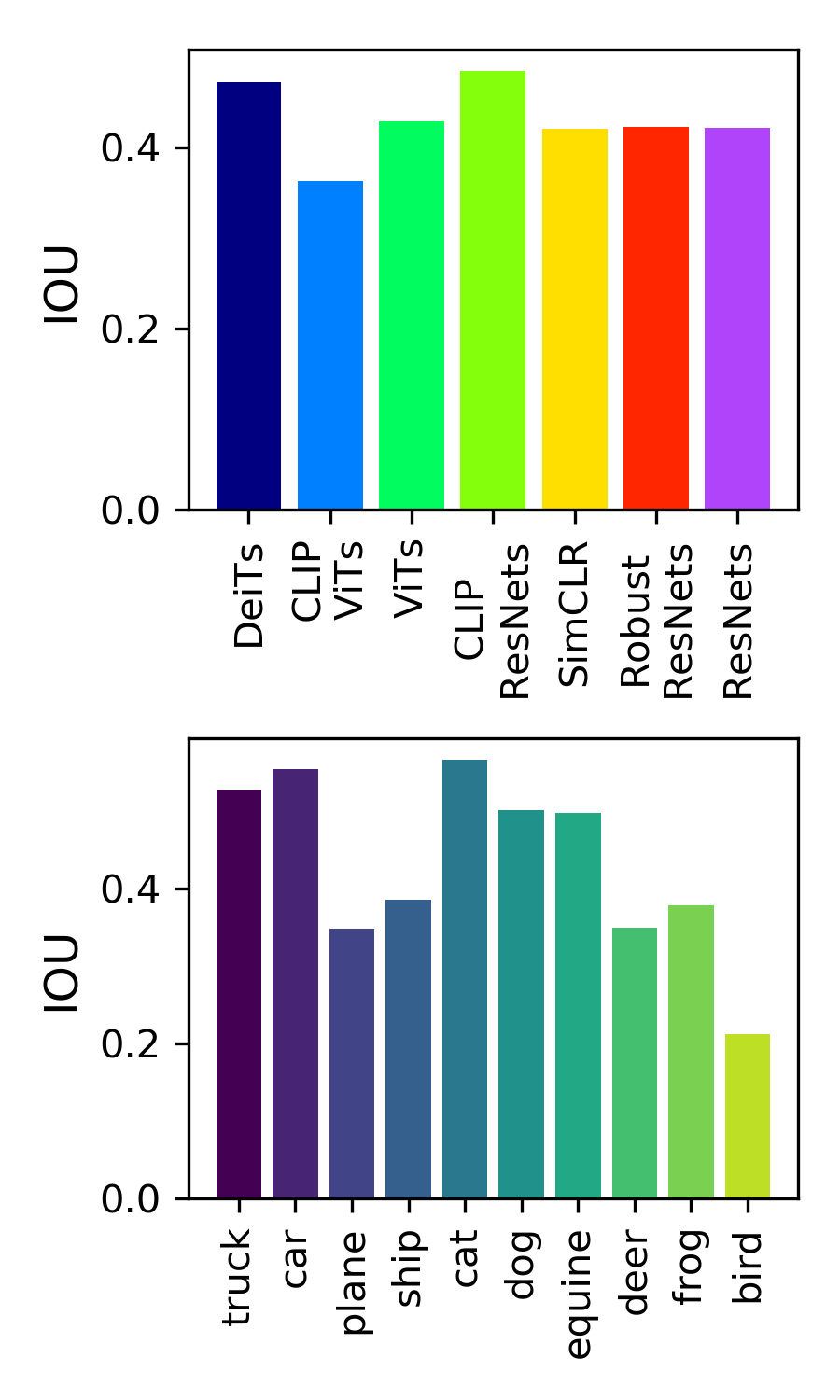}
    \end{minipage}
    \begin{minipage}{.195\textwidth}
    \centering
    \includegraphics[width=\linewidth]{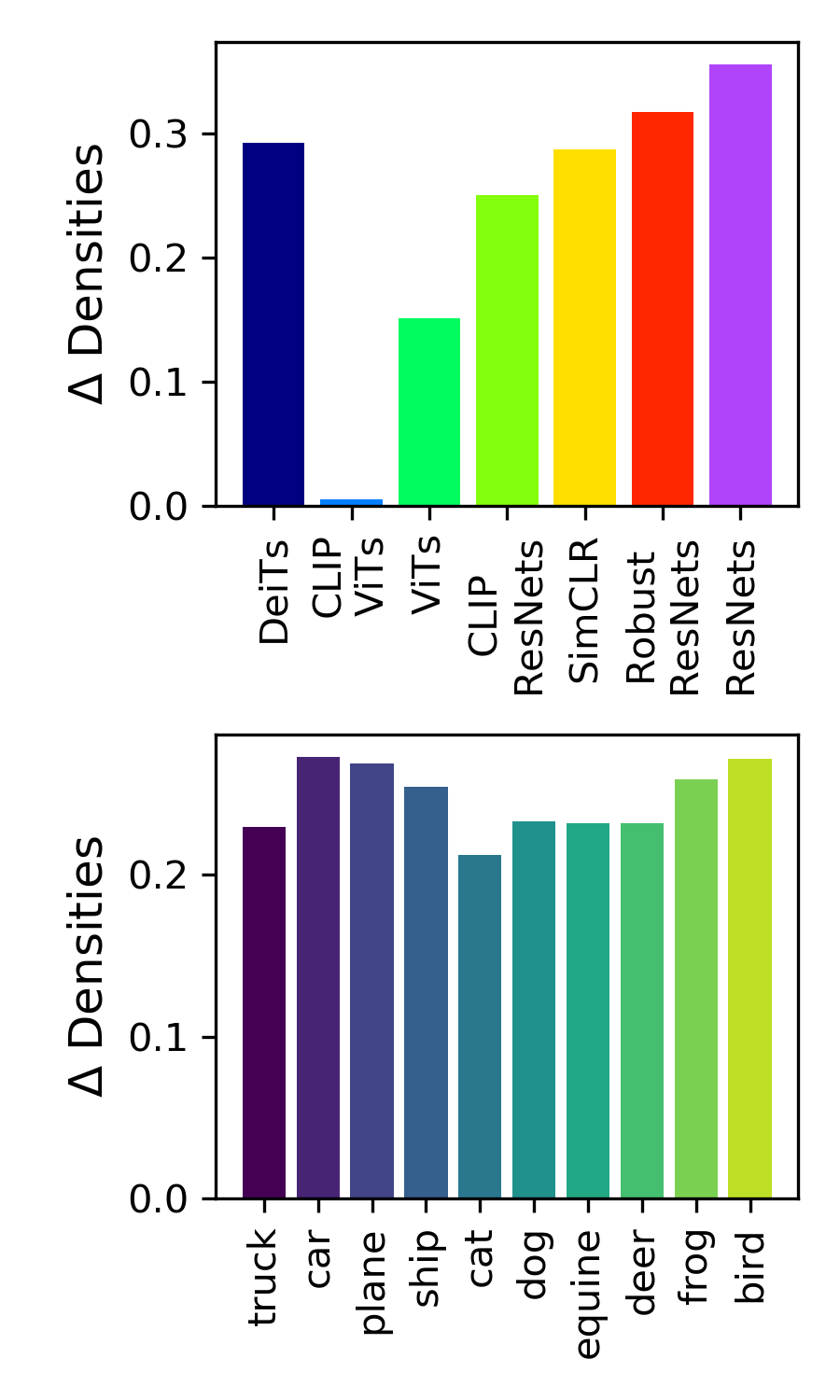}
    \end{minipage}
    \begin{minipage}{.195\textwidth}
    \centering
    \includegraphics[width=\linewidth]{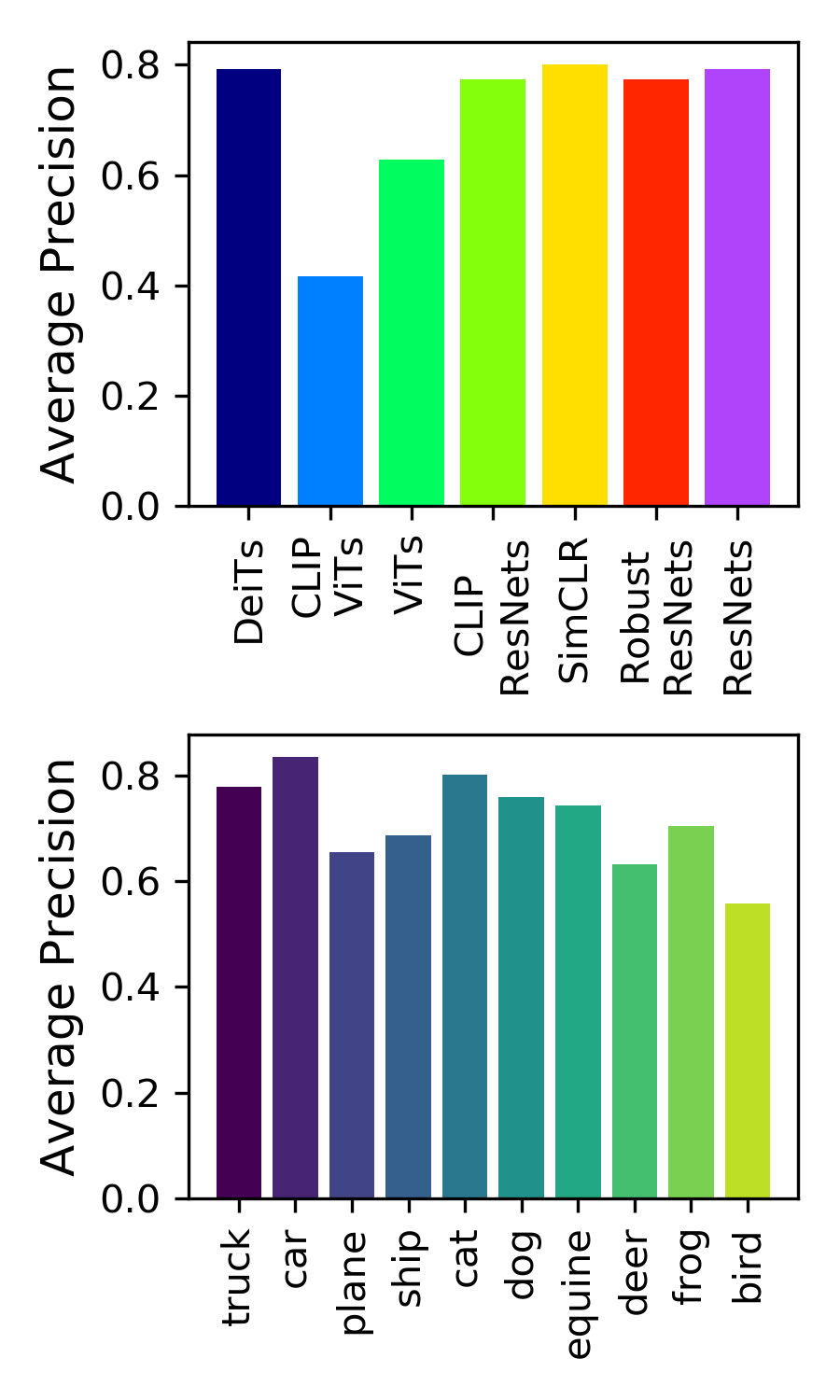}
    \end{minipage}
    \begin{minipage}{.195\textwidth}
    \centering
    \includegraphics[width=\linewidth]{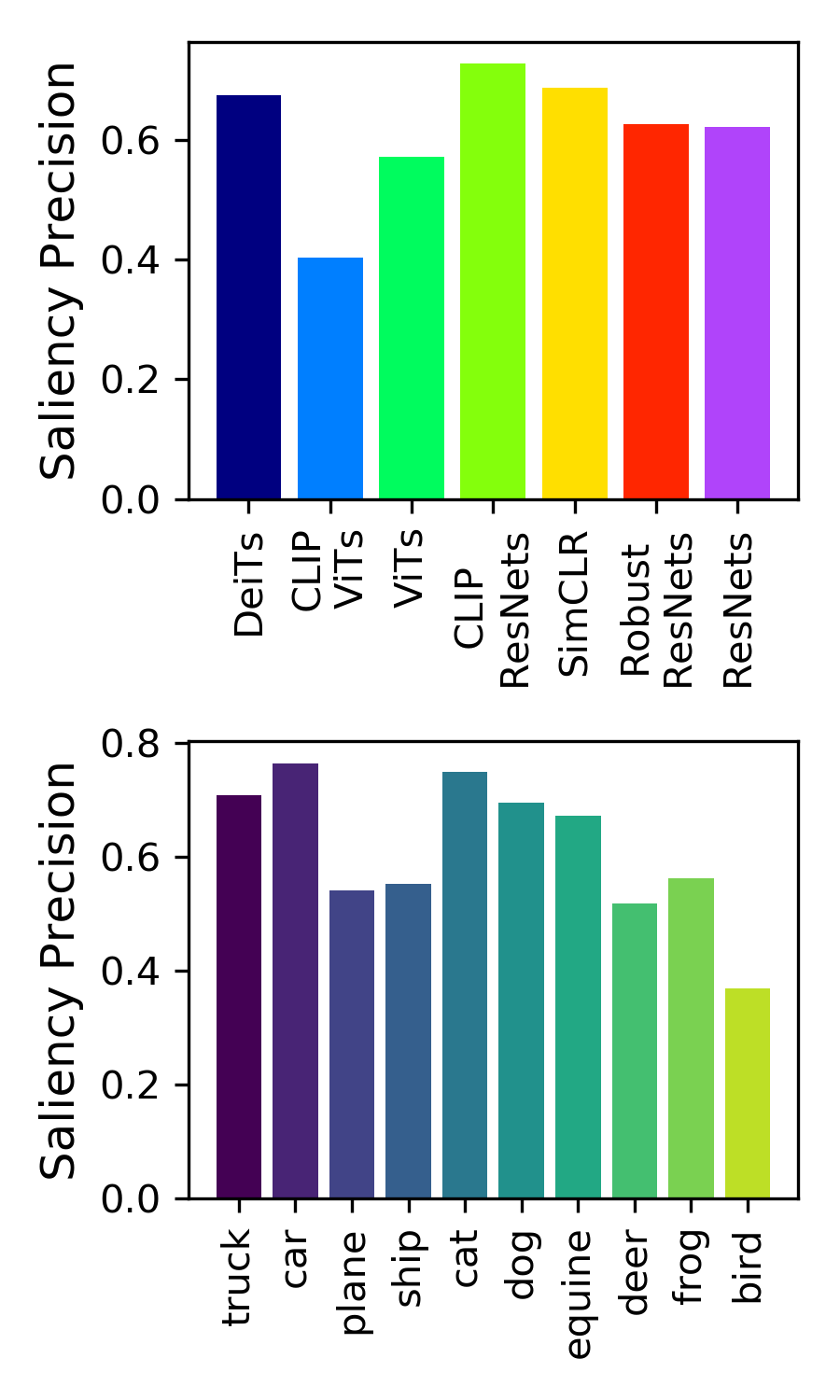}
    \end{minipage}
    \begin{minipage}{.195\textwidth}
    \centering
    \includegraphics[width=\linewidth]{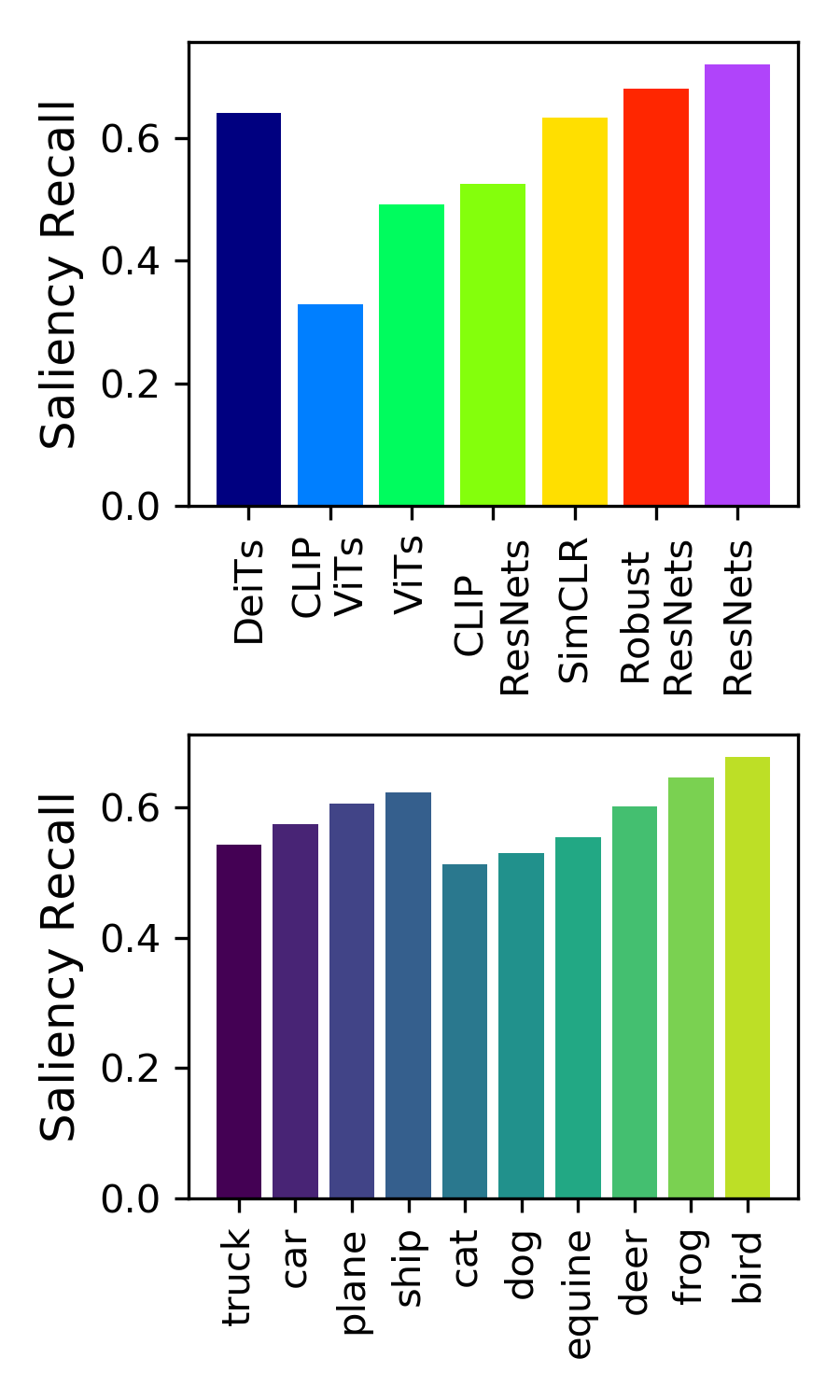}
    \end{minipage}
    \caption{Saliency alignment averaged over model categories (top) and object classes (bottom) for five alignment metrics.}
    \label{fig:all_sal_alignments}
\end{figure*}

To complement the noise analysis, we inspected saliency maps obtained via GradCAM \cite{gradcam}, which assigns a saliency score of 0 to 1 to each pixel. RIVAL10's segmentation masks allow for quantative assessment of the alignment of saliency to foregrounds. We inspected five metrics, defined as follows for a true binary object segmentation mask $\mathbf{m} \in \{0,1\}^d$ and a saliency map of equal shape $\mathbf{s} \in [0,1]^d$. Let $\mathbf{s}_\tau \in \{0,1\}^d$ be a binarized version of the saliency map, where a pixel of $\mathbf{s}_\tau$ is 1 only when its corresponding value in $\mathbf{s}$ is at least $\tau$. A standard metric in comparing segmentations is intersection over union (IOU), defined below.
$$\text{IOU} = \frac{\sum (\mathbf{m} \odot \mathbf{s}_{\tau=0.5})}{\sum(\mathbf{m})+\sum(\mathbf{s}_{\tau=0.5})} $$
Here, we assess the quality of the binarized saliency map as a segmentation mask of the foreground. We also found that inspecting the difference in average saliency for foreground and background pixels were useful, particularly in automatically discovering spurious background features. We define this metric, called $\Delta$ Densities, below.
$$\Delta\text{ Densities} = \frac{(\sum \mathbf{m} \odot \mathbf{s}) / \sum(\mathbf{m})}{(\sum (\mathbf{1} - \mathbf{m}) \odot \mathbf{s}) / \sum(\mathbf{1} -\mathbf{m})}$$
A third measure views saliency alignment as a binary classification task. Specifically, we compute average precision of a detector that uses pixel saliency as the discriminant score for classifying each pixel as foreground or background. Average precision combines recall and precision at all thresholds to give a general sense of discriminatory ability of some criteria. Formally, 
$$\text{Average Precision} = \sum_{n} (R_n-R_{n-1}) P_n$$
where $R_n, P_n$ refer to the precision and recall obtained at the $n^{th}$ threshold. 
Finally, we consider two additional metrics are analogs to precision and recall. Precision and recall typically hold meaning in binary classification tasks, though in our case, we wish to assess the alignment of saliency maps with continuous values (i.e. not true or false predictions). To this end, we define Saliency Precision and Saliency Recall as follows.

$$\text{Saliency Precision} = \frac{\sum \mathbf{s} \odot \mathbf{m}}{\sum \mathbf{s}}$$
Essentially, this amounts to a weighted precision, placing more importance on having highly salient pixels fall in the foreground. Another interpretation of this metric is the fraction of total saliency in the foreground, similar to \cite{rcls}.
$$\text{Saliency Recall} = \frac{\sum \mathbf{s}_{\tau=\tau^*} \odot \mathbf{m}}{\sum \mathbf{m}}$$
For Saliency Recall, we compute recall as normal on a binarized saliency map. However, the binarization threshold $\tau^*$ is chosen dynamically so to only retain the pixels that account for $75\%$ of total saliency. That is, $\frac{\sum\mathbf{s}_{\tau=\tau^*}}{\sum\mathbf{s}} = 0.75$. Intuitively, saliency recall captures the fraction of the segmentation mask that are among the more salient pixels. 

\subsection{Empirical Observations}
We present complete quantitative saliency alignment results in Figure \ref{fig:all_sal_alignments}. Generally, there is not a strong separation among models observed across all metrics. CLIP ViTs consistently score lower, with an average $\Delta$ Densities near zero. ViTs also generally have lower saliency alignment. Recall that at low noise levels, the transformer models had low relative foreground sensitivity. One may be inclined to argue that the saliency alignment analysis corroborates those results. However, we hesitate to make such assertions, as the results are not consistent across metrics, and key exceptions (such as the high alignment of DeiTs and Robust ResNets) exist. Our overall impression from the saliency analysis is that alignment GradCAMs to foregrounds may not always imply high relative sensitivity to foreground noise, suggesting that saliency maps alone may not capture the full story of a model's sensitivities.

Qualitatively, the GradCAMs for all transformers are much more patchy than ResNets, which usually yield GradCAMs with saliency organized in one or two clusters. We attribute this to the fundamental difference in how images are processed by ResNets, who employ significant spatial inductive biases, and transformers, who view images a set of patches that can attend to one another.

Looking to object classes, we see that the variance in alignment due to class observed for IOU is corroborated by average precision and saliency precision. When inspecting saliency recall, however, we see higher alignments for birds and ships. We believe this is an effect of the bias of Saliency Recall in favor of images with smaller foreground masks. Furthermore, high recall can still be consistent with poor foreground sensitivity, as the saliency map may cover much of both the foreground and background. 

\section{Attribute Ablation}

\begin{figure}
    \centering
    \includegraphics[width=0.95\linewidth]{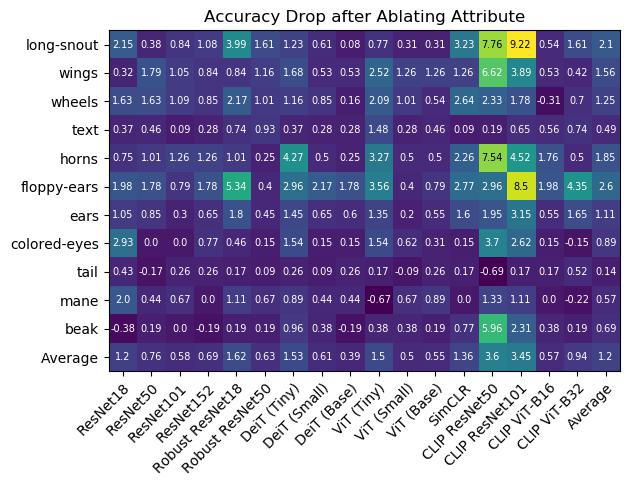}
    \caption{Degradation to model performance due to attribute ablation (via graying), as measured by accuracy.}
    \label{fig:attr_ablation}
\end{figure}

To assess sensitivity to attributes, we inspect the extreme of ablating the attributes entirely via graying. We do not consider attributes that cover the entire object, as ablating the attribute would remove the entire foreground. Overall, the removal of any individual attribute only slightly reduces model performance. The largest reduction occurs in CLIP ResNets, with an average drop in accuracy of roughly $3.5\%$. For most attribute-model pairs, accuracy drop is less than $1\%$. This suggests that attributes human deem informative in performing RIVAL10 classification are not very important to deep classifiers.

\section{Additional Qualitative Examples of Background Sensitivity}

\begin{figure*}
    \centering
    \includegraphics[width=\linewidth]{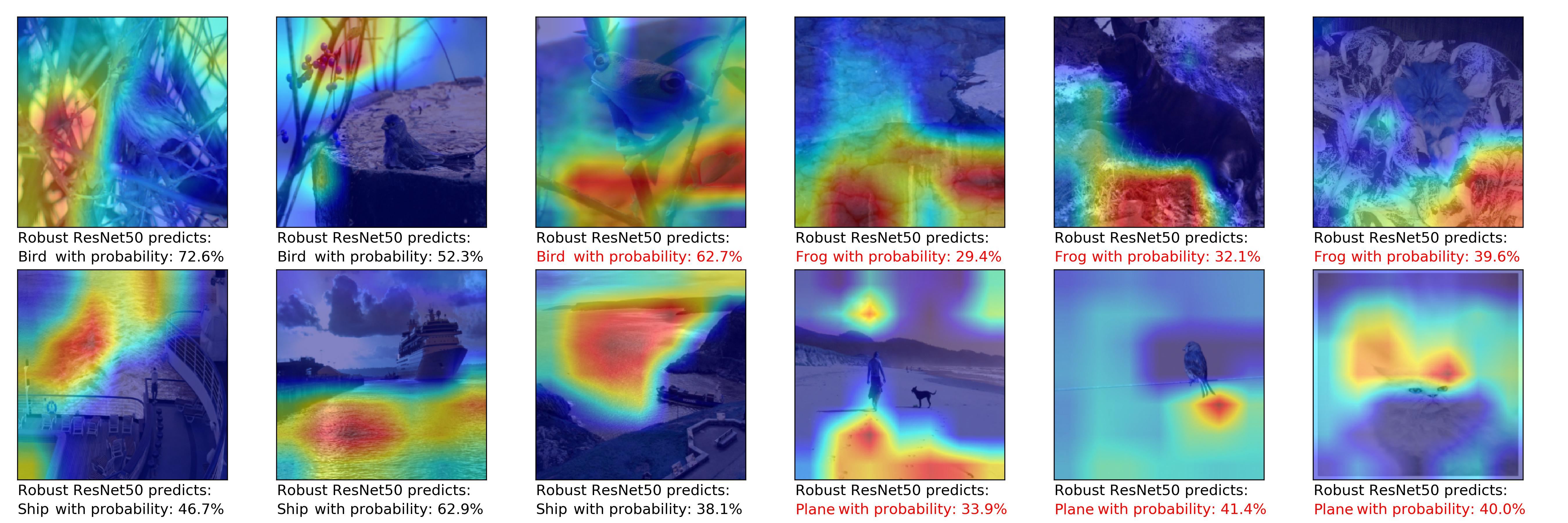}
    \caption{Additional examples of spurious features used by a Robust ResNet50 observed via sorting images by saliency alignment ($\Delta$ Densities). Misclassifications are in red text. Spurious features include branches, dry leaves, water, and sky.}
    \label{fig:robust50_gc}
\end{figure*}

\begin{figure*}
\centering
% \newsubfloat{
\begin{subfloat}
    \centering
    \begin{minipage}{0.33\textwidth}    
    \centering
    \includegraphics[width=0.99\linewidth]{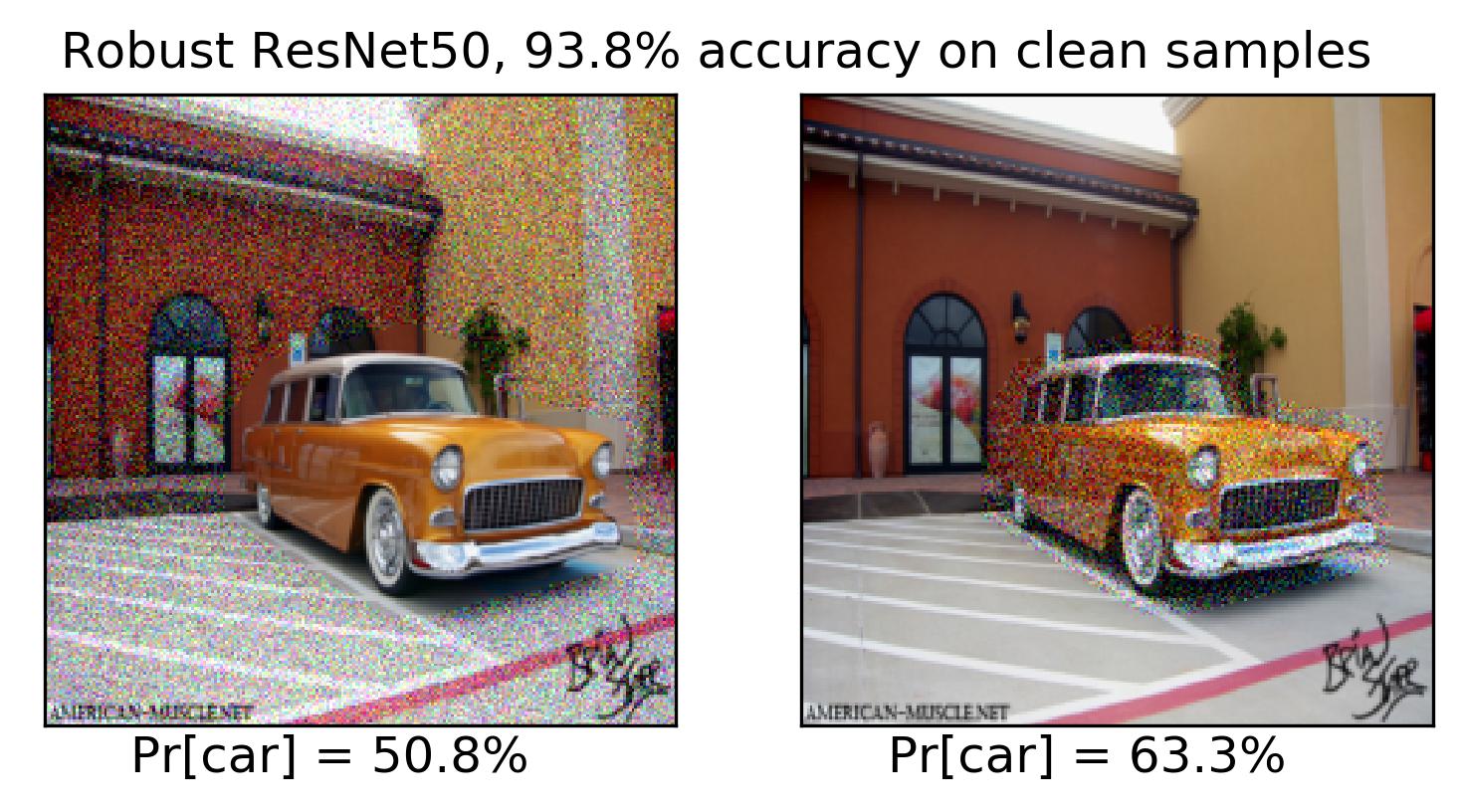}
    \end{minipage}
    \begin{minipage}{0.33\textwidth}    
    \centering
    \includegraphics[width=0.99\linewidth]{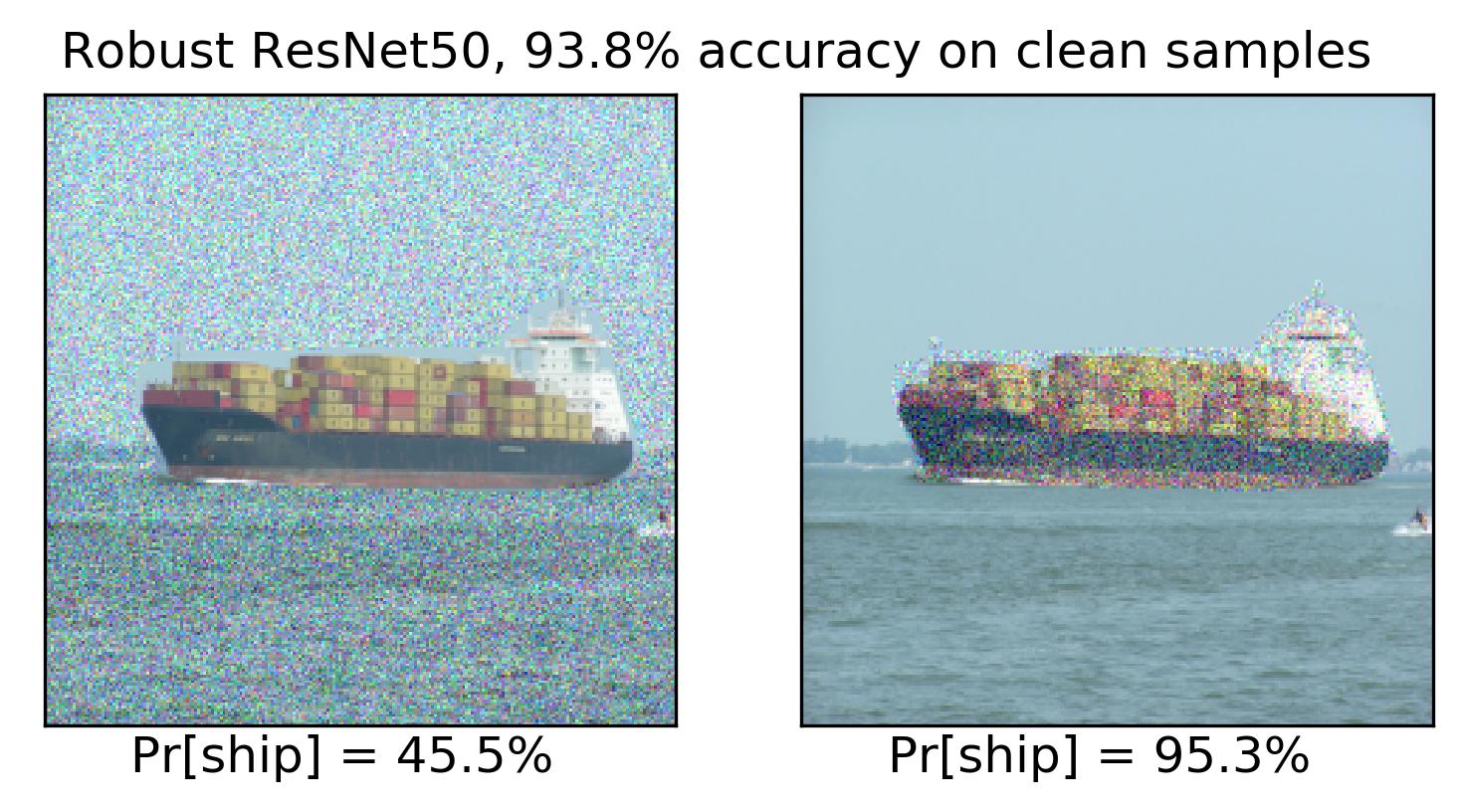}
    \end{minipage}
    \begin{minipage}{0.33\textwidth}    
    \centering
    \includegraphics[width=0.99\linewidth]{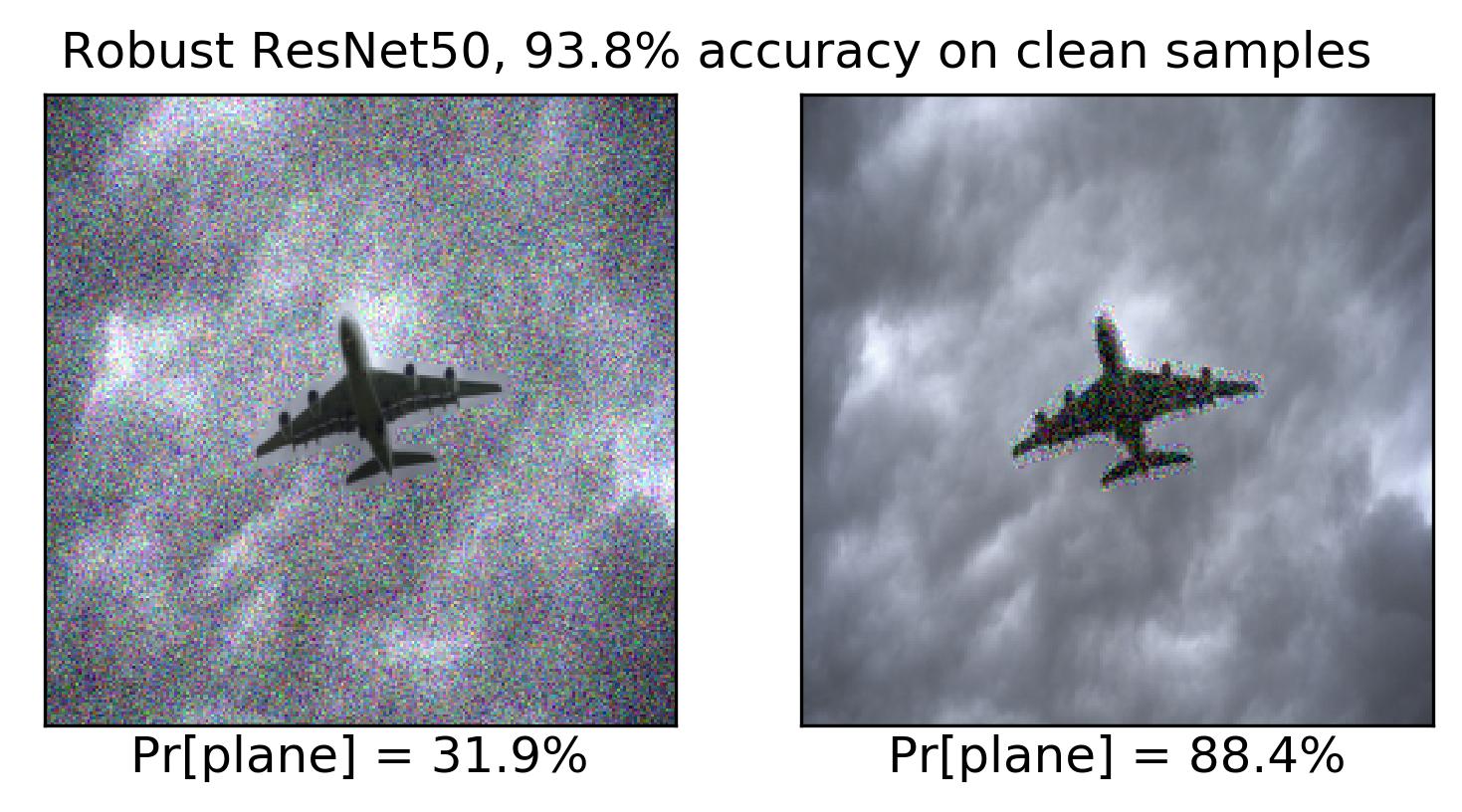}
    \end{minipage}
% }
\end{subfloat}
\centering
% \newsubfloat{
\begin{subfloat}
% \end{figure*}
% \begin{figure*}
    \centering
    \begin{minipage}{0.33\textwidth}    
    \centering
    \includegraphics[width=0.99\linewidth]{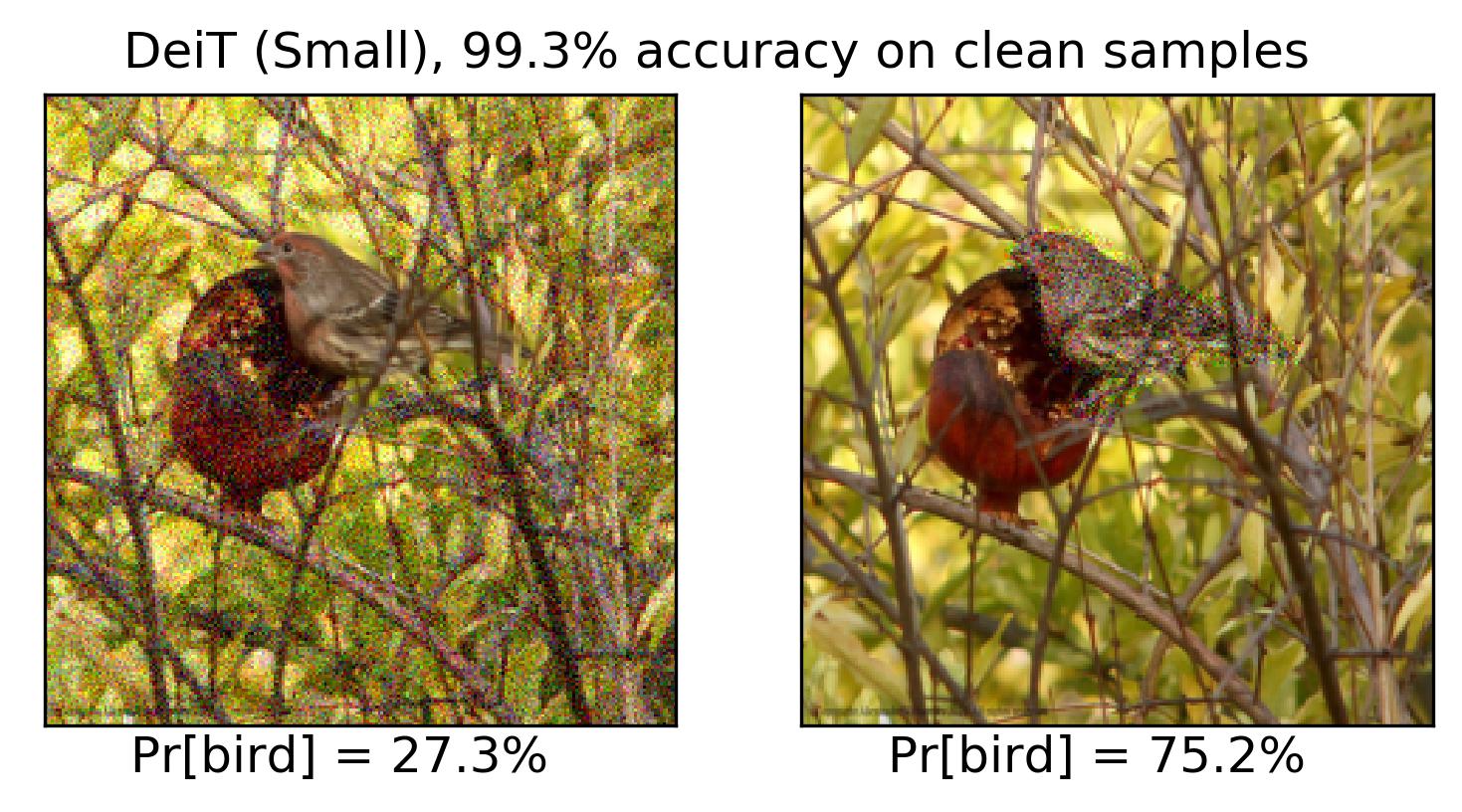}
    \end{minipage}
    \begin{minipage}{0.33\textwidth}    
    \centering
    \includegraphics[width=0.99\linewidth]{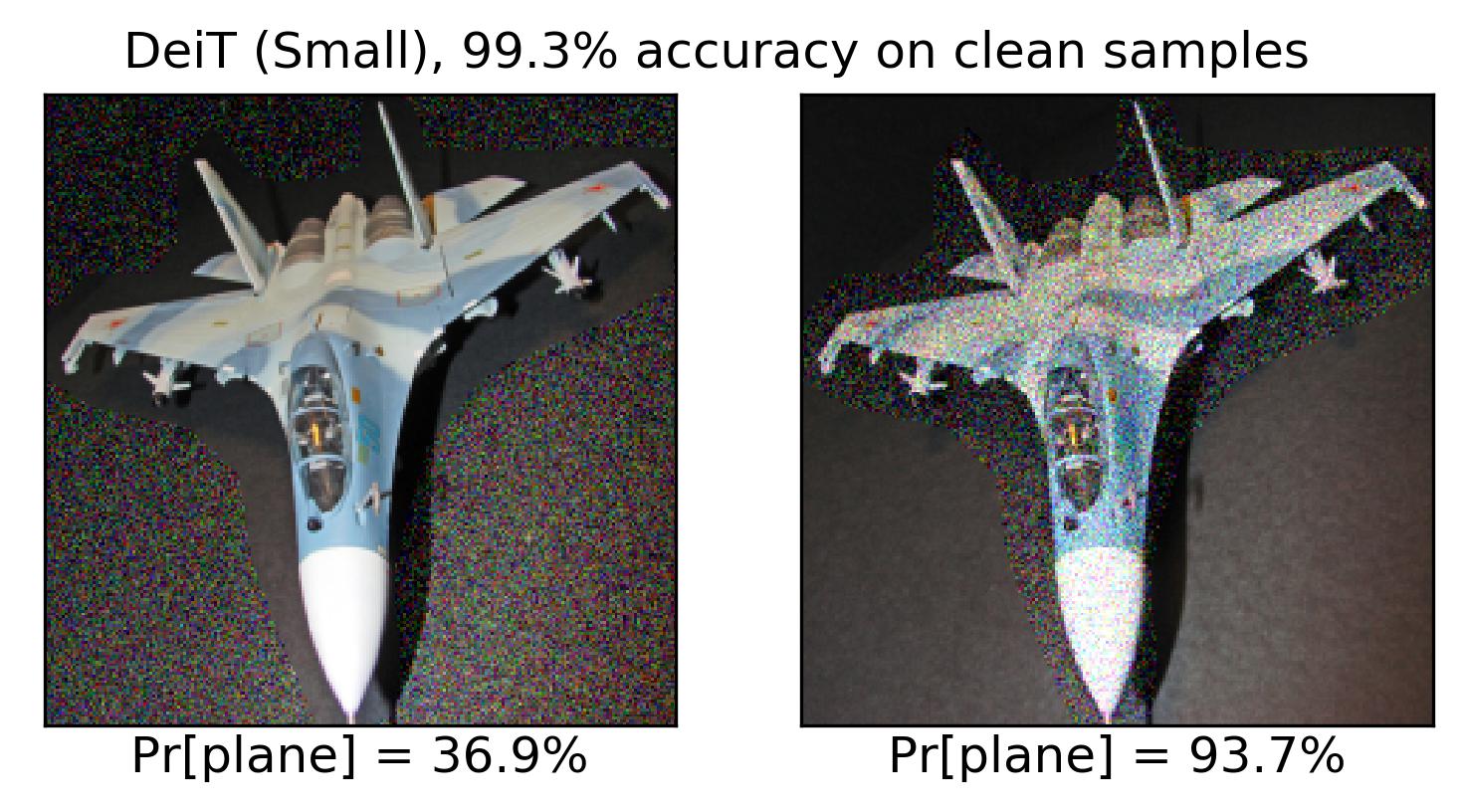}
    \end{minipage}
    \begin{minipage}{0.33\textwidth}    
    \centering
    \includegraphics[width=0.99\linewidth]{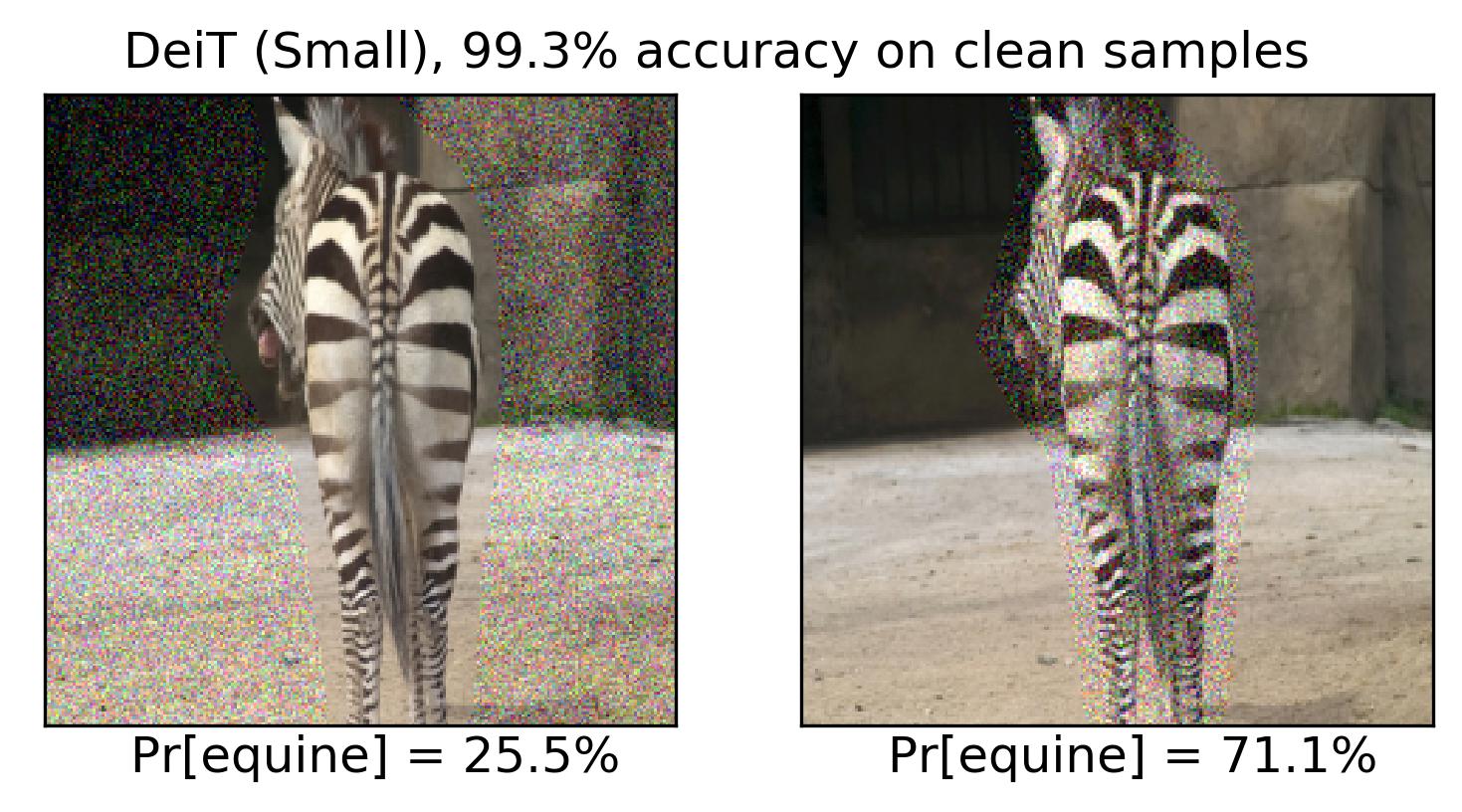}
    \end{minipage}
% }
\end{subfloat}
    \caption{Additional examples where background noise degrades performance of highly accurate models more than foreground noise. ({\bf top}): Robust ResNet50, ({\bf bottom}): DeiT (Small). Gaussian $\ell_\infty$ noise with standard deviation $\sigma = 0.12$ shown. Probabilities are averaged over ten trials.}
    \label{fig:noise_egs}
\end{figure*}

We provide additional examples qualitatively demonstrating instances where models have high background sensitivity. Figure \ref{fig:robust50_gc} shows GradCAMs where saliencies have worst alignment with foreground, as measured by $\Delta$ Densities, for a Robust ResNet50. In Figure \ref{fig:noise_egs}, we display instances where noise corruption reveals greater background sensitivity for Robust ResNet50 and DeiT (Small). 

\section{Additional Visualizations for Neural Node Attribution}

\begin{figure}[ht]
\vspace{-0.2cm}
\centering
\includegraphics[width=0.45\textwidth]{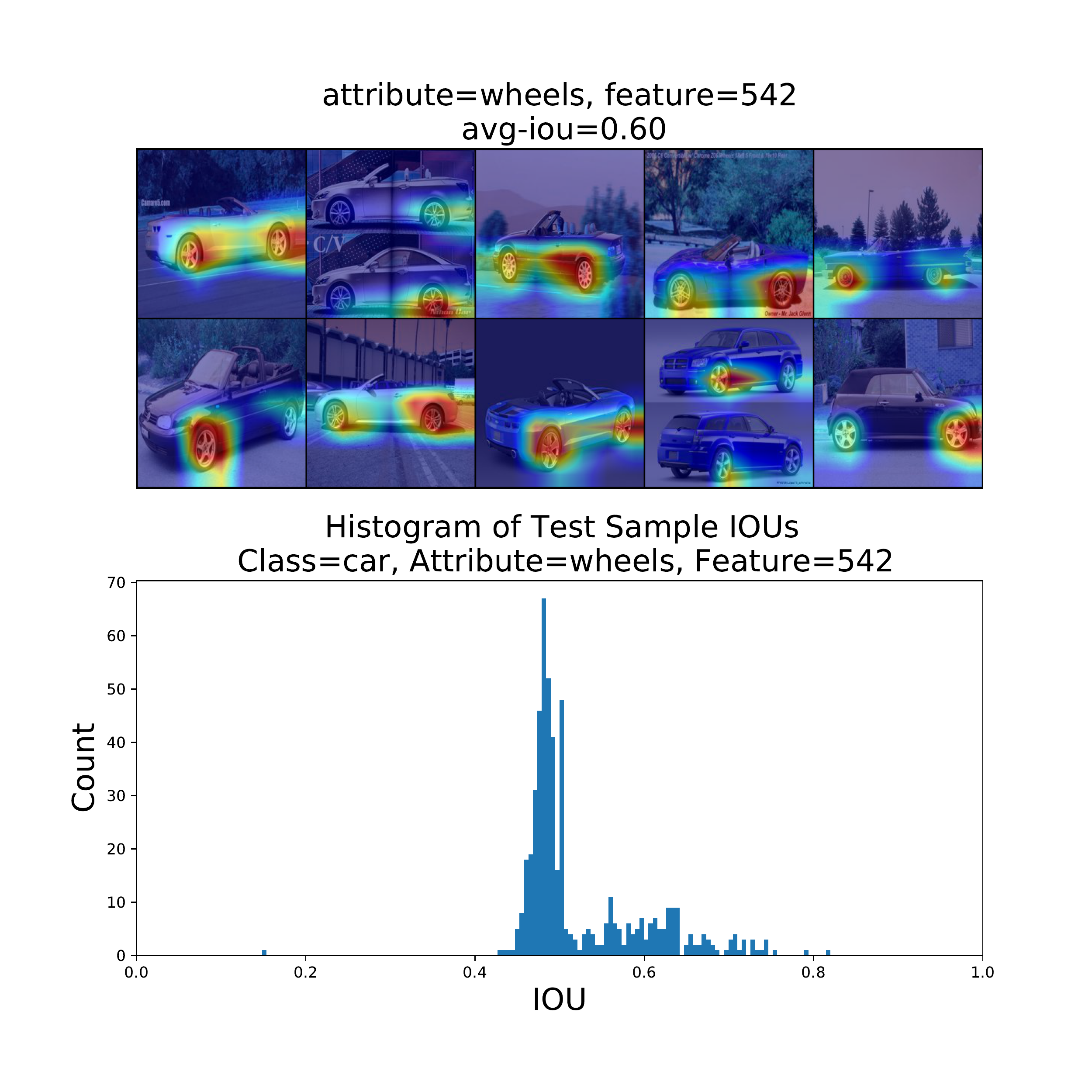}
\caption{(\textbf{Top}): Example GradCAMs on test images with respect to the top feature identified by IOU in training set.
(\textbf{Bottom}): Histograms of IOUs corresponding to this feature, attribute pair.}
\label{fig:top-gradcam-and-hist-2}
\vspace{-0.4cm}
\end{figure}

\begin{figure}[ht]
\vspace{-0.2cm}
\centering
\includegraphics[width=0.45\textwidth]{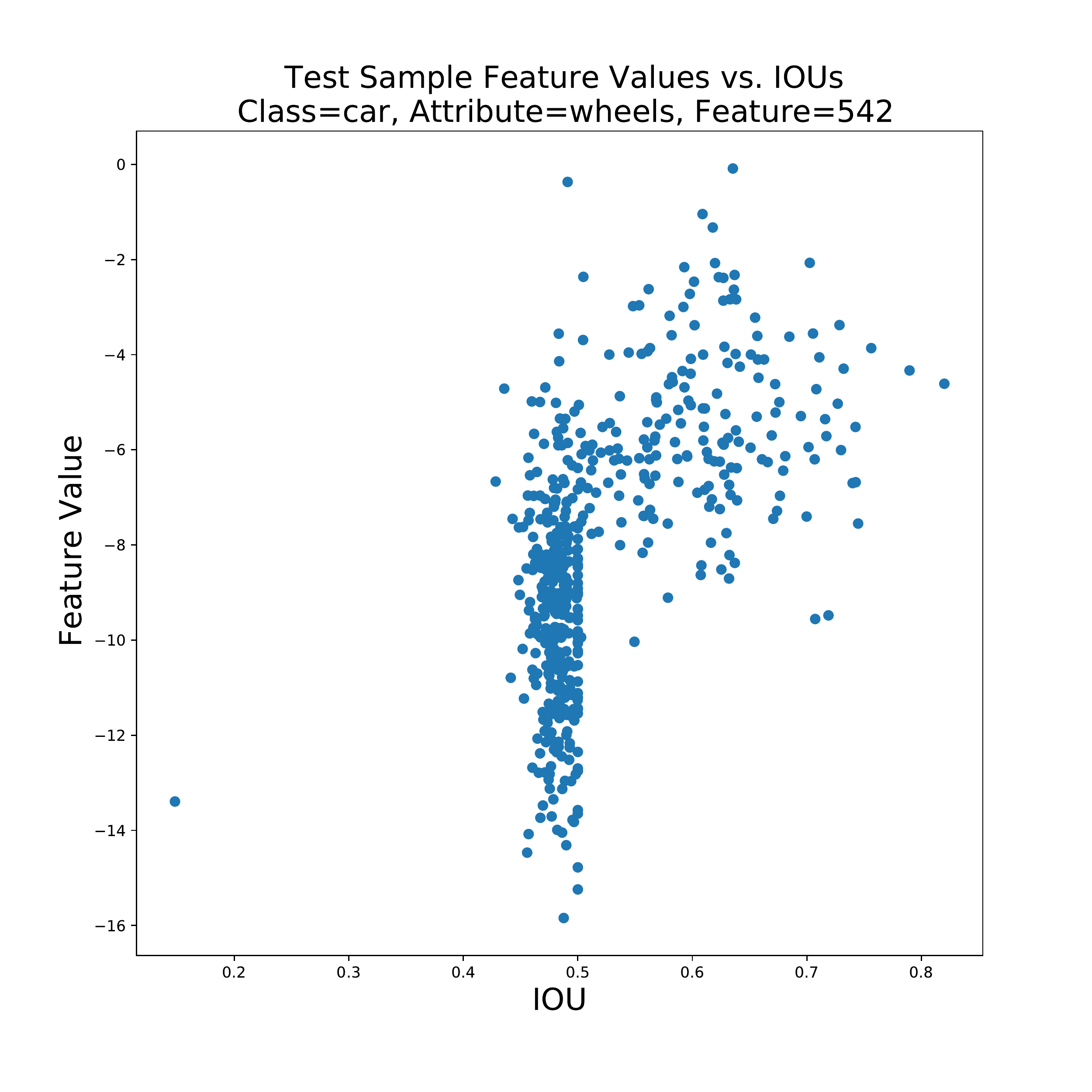}
\caption{Feature values vs IOU scores for class-attribute pair car and wheels.}
\label{fig:top-iou-feat-scatter-1}
\vspace{-0.4cm}
\end{figure}

\begin{figure}[ht]
\vspace{-0.2cm}
\centering
\includegraphics[width=0.45\textwidth]{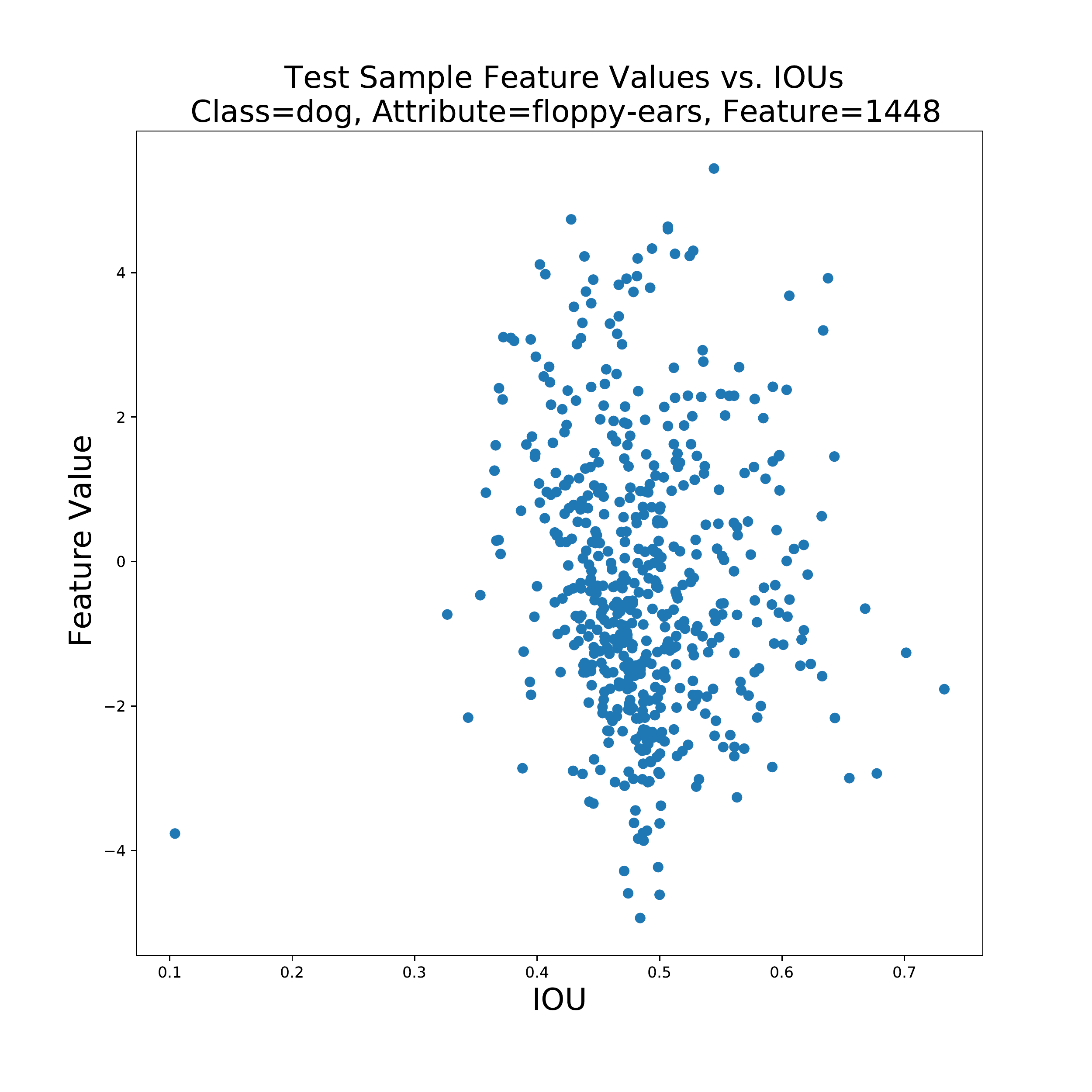}
\caption{Feature values vs IOU scores for class-attribute pair dog and floppy-ears.}
\label{fig:top-iou-feat-scatter-2}
\vspace{-0.4cm}
\end{figure}

We show GradCAMs and IOU histograms for another top feature attribute pair, cars and wheels, in Figure \ref{fig:top-gradcam-and-hist-2}. We observe qualitatively the same results as in the main text: IOU scores are high for this attribute on samples in this class. We also show scatterplots of IOUs vs. feature activations for this top pair as well as dogs and floppy-ears, the pair discussed in the main text, in Figures  \ref{fig:top-iou-feat-scatter-1} and \ref{fig:top-iou-feat-scatter-2} respectively. Interestingly, feature activation value and saliency alignment (as measured by IOU) do not seem to be strongly related.

\section{Attribute-specific Neural Node Attribution}

\begin{figure*}
    \centering
    \begin{minipage}{0.9\textwidth}
    \includegraphics[width=\linewidth]{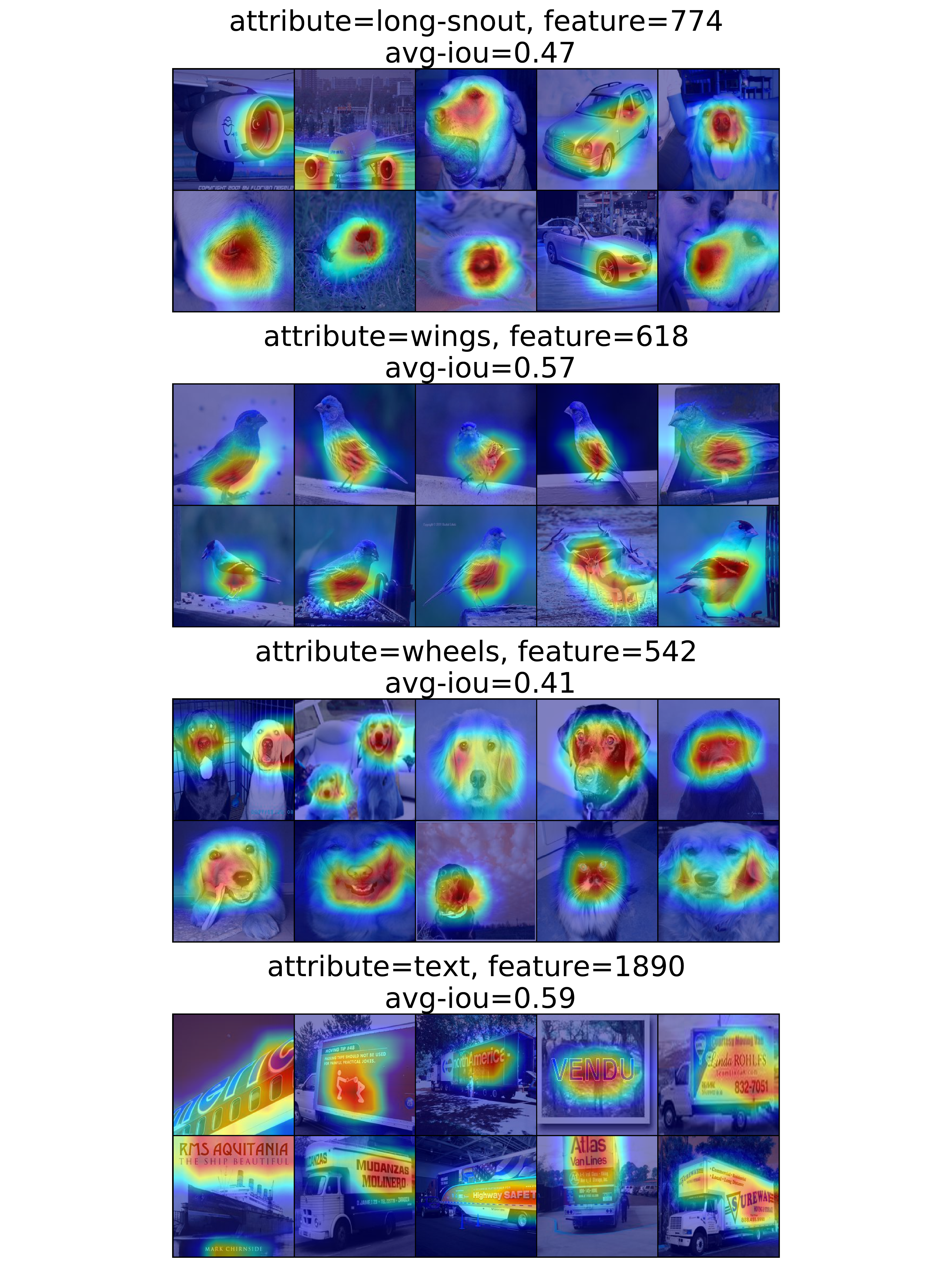}
    \caption{Saliency for top feature attribute pairs by IOU. First quarter of results shown here.}
    \label{fig:top-gradcams-all-1}
    \end{minipage}
\end{figure*}

\begin{figure*}
    \centering
    \begin{minipage}{0.9\textwidth}
    \includegraphics[width=\linewidth]{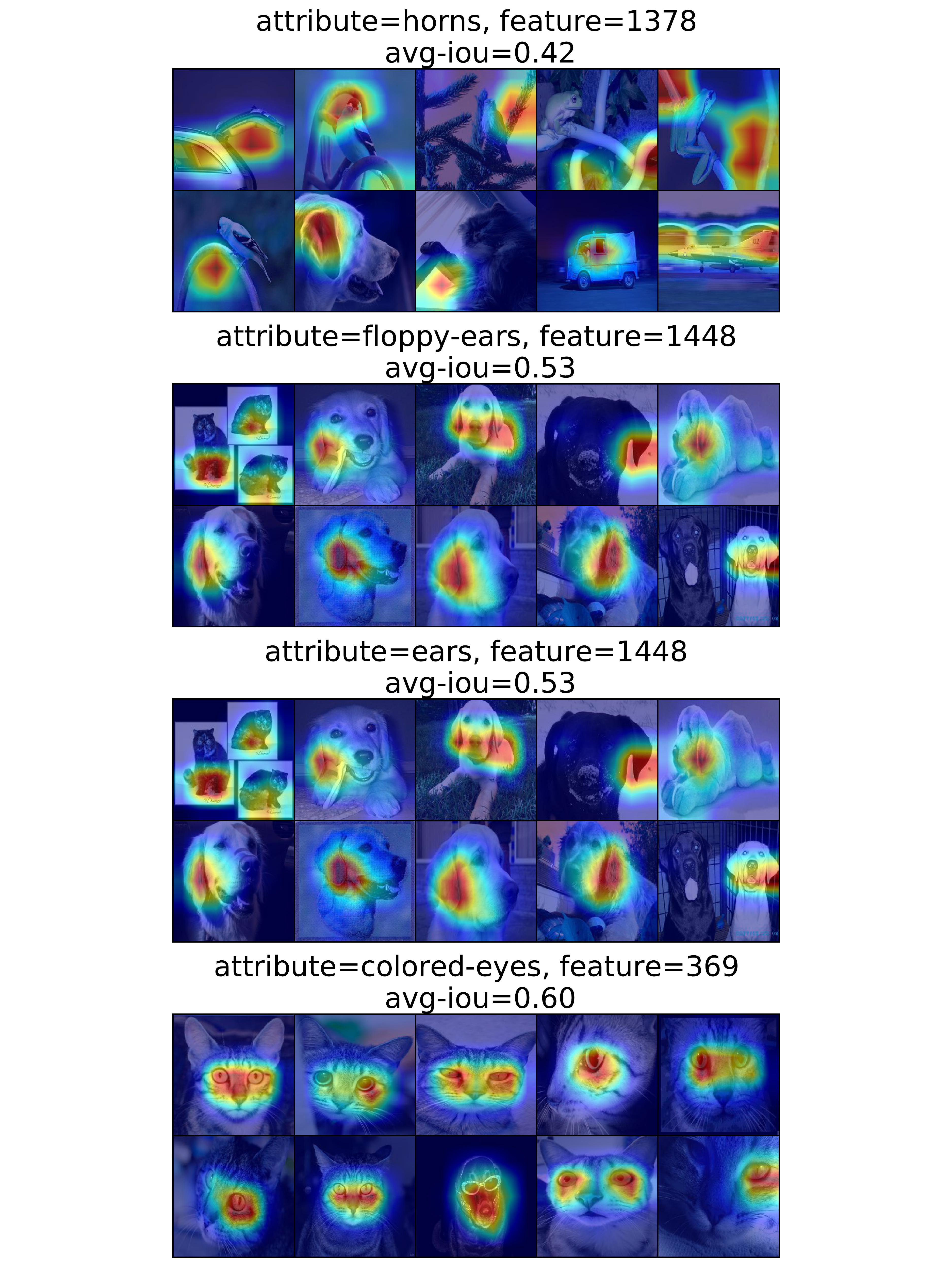}
    \caption{Saliency for top feature attribute pairs by IOU. Second quarter of results shown here.}
    \label{fig:top-gradcams-all-2}
    \end{minipage}
\end{figure*}

\begin{figure*}
    \centering
    \begin{minipage}{0.9\textwidth}
    \includegraphics[width=\linewidth]{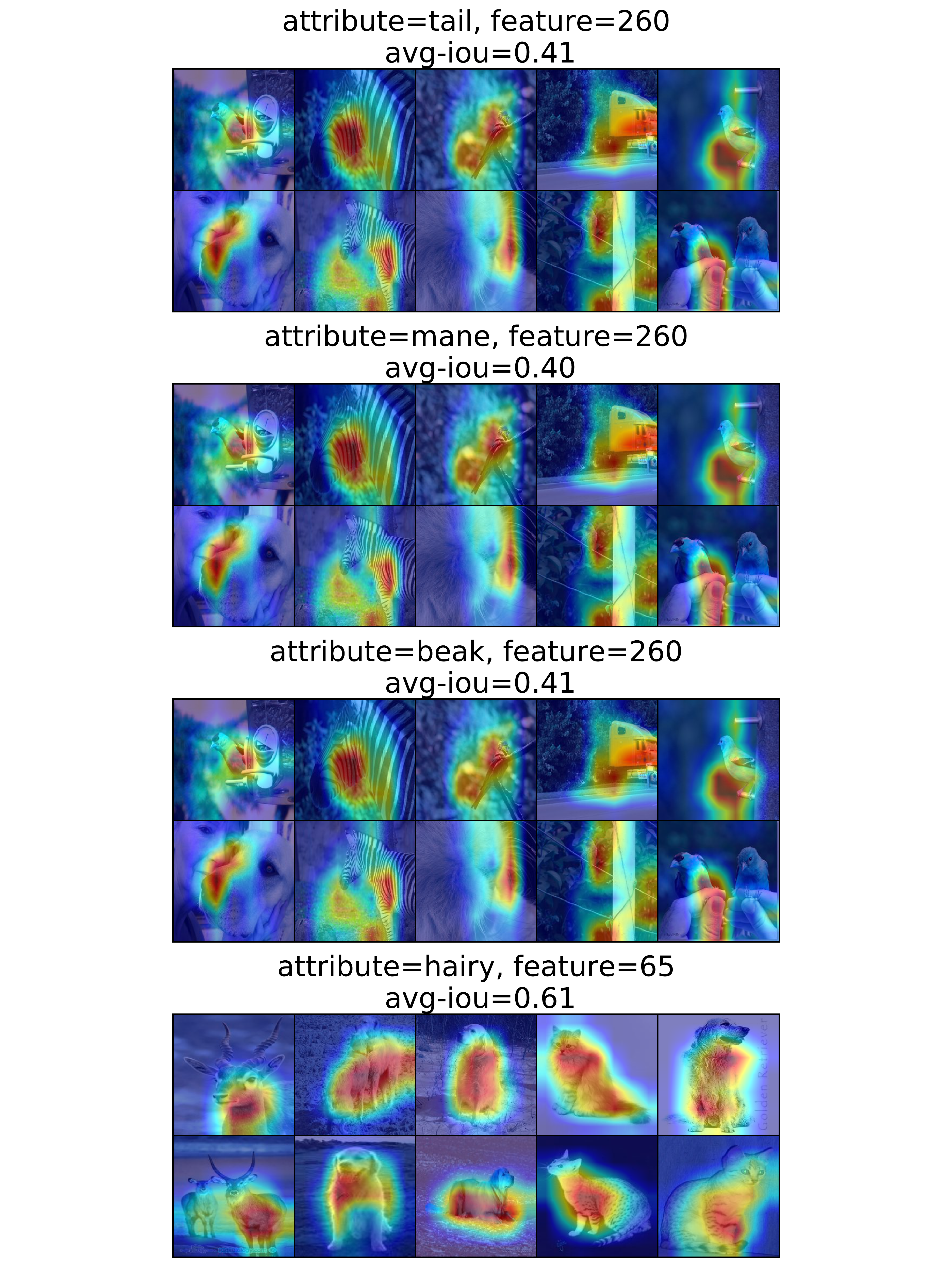}
    \caption{Saliency for top feature attribute pairs by IOU. Third quarter of results shown here.}
    \label{fig:top-gradcams-all-3}
    \end{minipage}
\end{figure*}

\begin{figure*}
    \centering
    \begin{minipage}{0.9\textwidth}
    \includegraphics[width=\linewidth]{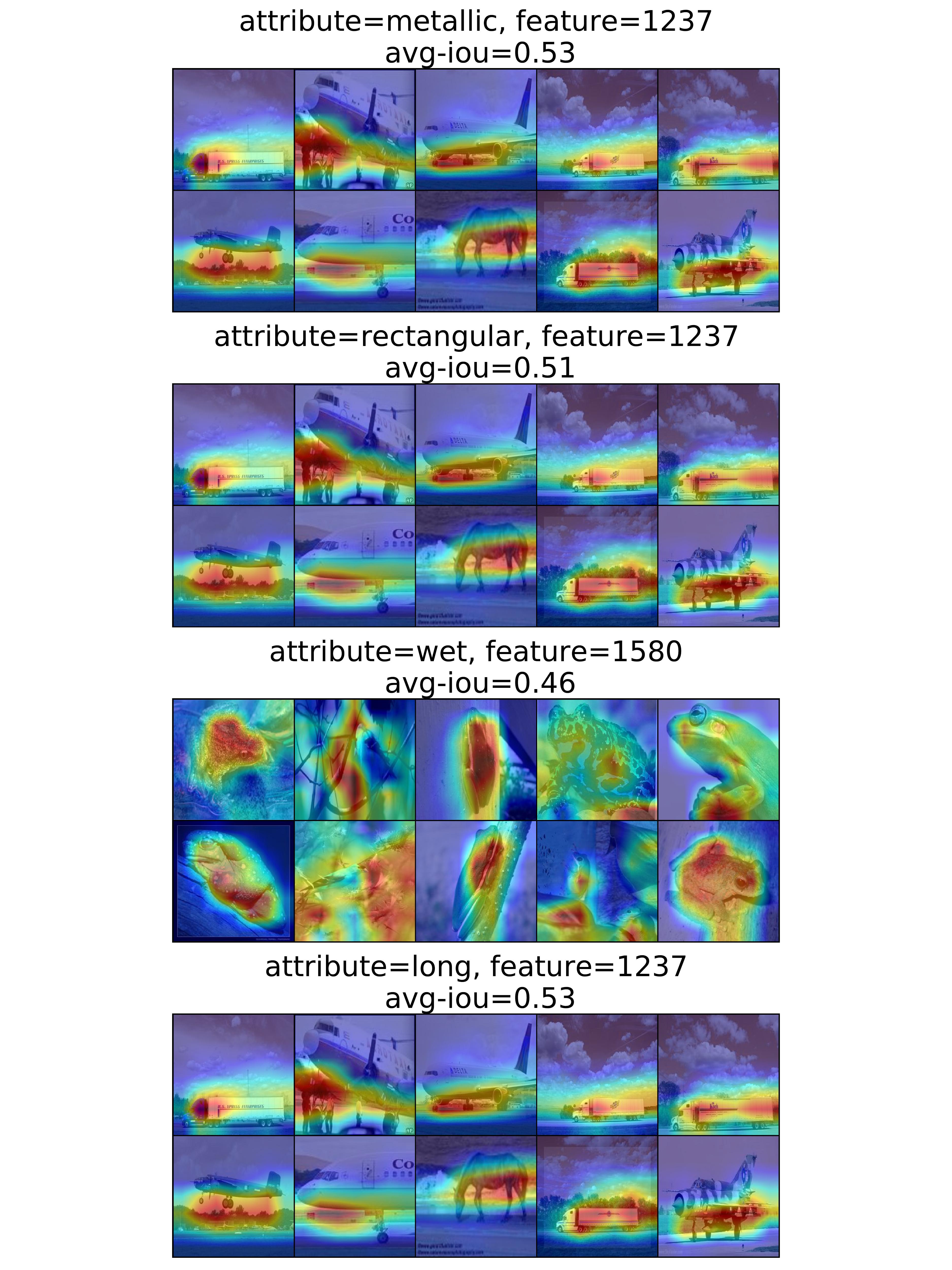}
    \caption{Saliency for top feature attribute pairs by IOU. Fourth quarter of results shown here.}
    \label{fig:top-gradcams-all-4}
    \end{minipage}
\end{figure*}

\begin{figure*}
    \centering
    \begin{minipage}{0.9\textwidth}
    \includegraphics[width=\linewidth]{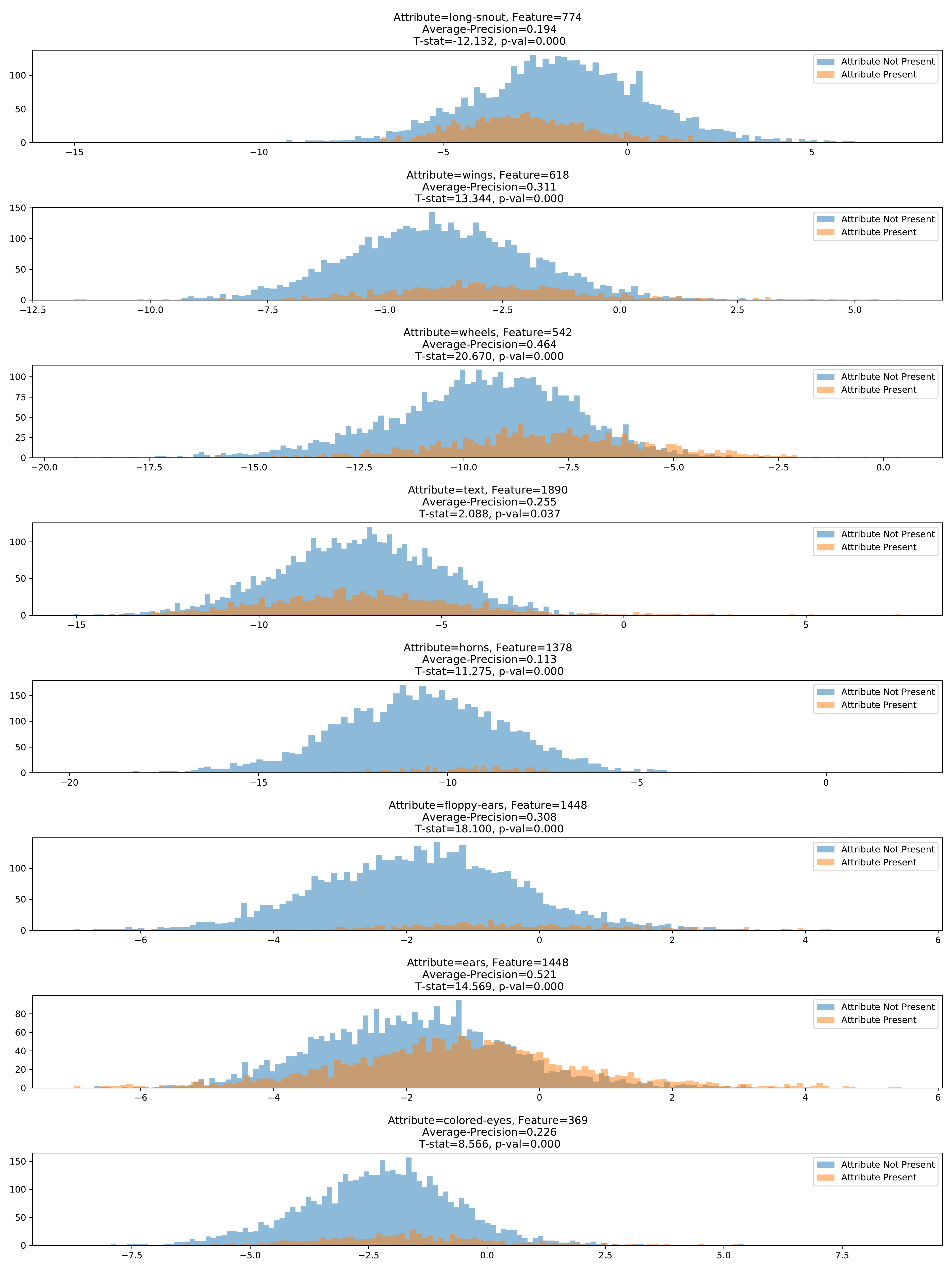}
    \caption{Top feature histograms for the top attribute feature pairs. First half shown here.}
    \label{fig:top-hist-all-1}
    \end{minipage}
\end{figure*}

\begin{figure*}
    \centering
    \begin{minipage}{0.9\textwidth}
    \includegraphics[width=\linewidth]{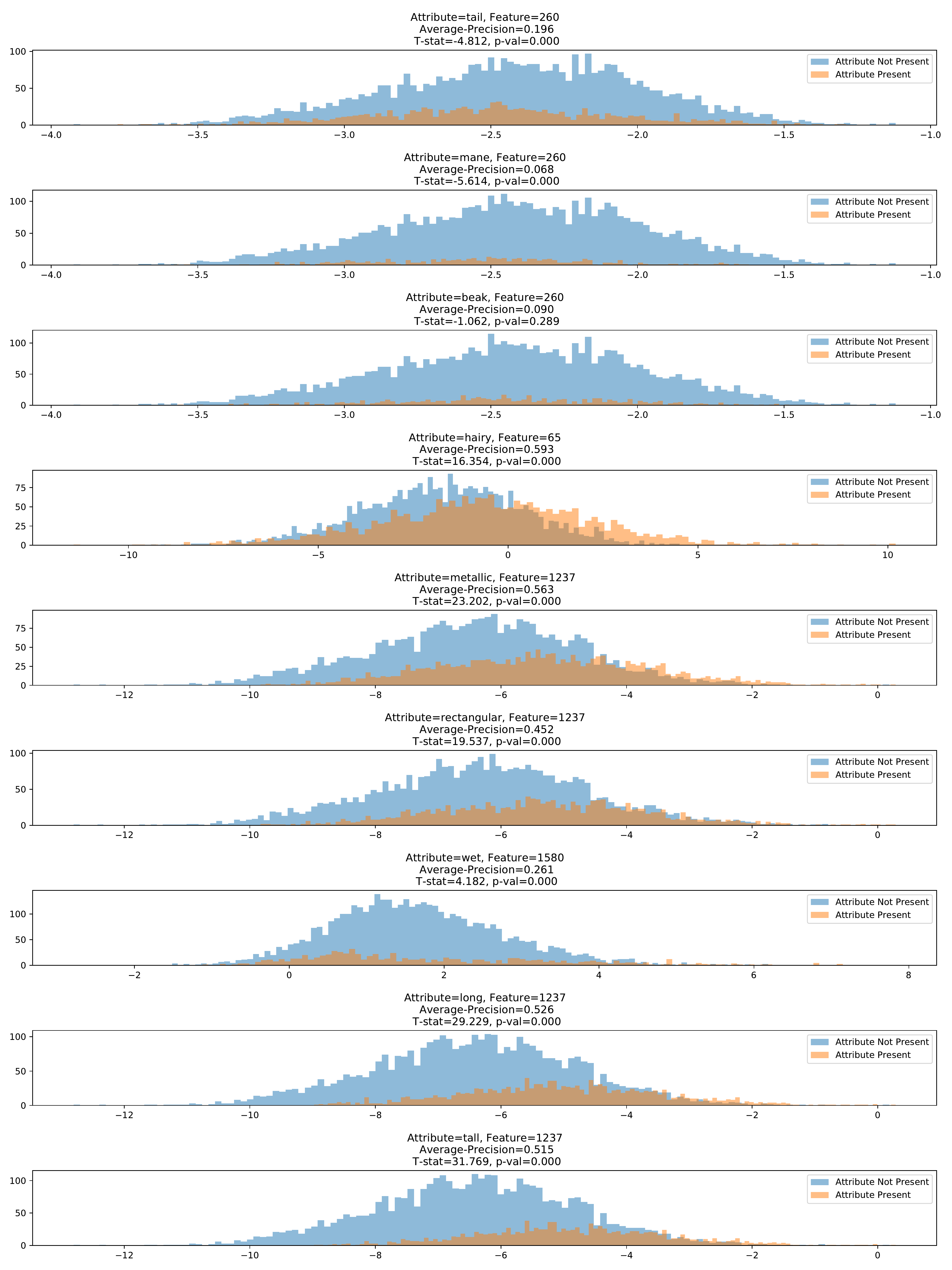}
    \caption{Top feature histograms for the top attribute feature pairs. Second half shown here.}
    \label{fig:top-hist-all-2}
    \end{minipage}
\end{figure*}

We report a variant of the neural node attribution section in the main text, where we do not filter by class. Instead, we focus the analysis on attributes. We use the same procedure to identify top feature attribute pairs as in the main text, \textit{except} for filtering by class. We show the complete saliency results for top feature attribute pairs for all attributes in Figures \ref{fig:top-gradcams-all-1}, \ref{fig:top-gradcams-all-2}, \ref{fig:top-gradcams-all-3}, \ref{fig:top-gradcams-all-4}. In addition, we show activation histograms for top feature attribute pairs identified by the method, colored them by the presence or not of the attribute in Figures \ref{fig:top-hist-all-1} and \ref{fig:top-hist-all-2}. We observe that the feature distributions do not separate for test samples with and without that attribute, \textit{despite} the reasonable quality of the GradCAMs. Note that we present GradCAMs on the top activating test images. The GradCAMs for top activating training images are even better, though this by design, as we choose feature-attribute pairs to maximize saliency alignment in training images.

This implies that filtering by class is necessary for the node attribution methods here discussed. When the same analysis is carried out irrespective of class, nodes cannot clearly be attributed. This result casts doubt on performing node attribution in \textit{class-free} fashion via saliency methods, though some authors argue that filtering by class reflects the actual practice of node attribution via saliency methods.

\section{Limitations}

The central challenge of our work is performing comparisons across diverse model types. In particular, the variance in general noise robustness poses as a major obstacle in employing our noise analysis. We believe that we have devised a normalization scheme to account for this, though there are likely other differences across models that could not be completely controlled against.

Moreover, our study only considers classification on ten relatively disparate classes. It is possible that as the classification task becomes more challenging, models rely less on short cuts out of necessity. However, it is also plausible that they make greater use of spurious features, as they seek any information that will help. Frankly, our study can not directly anticipate the outcome of repeating our analysis for a more difficult classification task. In future work, we may build on RIVAL10 to craft more finegrained classification tasks, perhaps leveraging attribute insertion and removal.

Lastly, we focus on only one saliency method throughout our analysis. It is possible that other saliency methods may produce maps that were more informative, or more in line with the results of our noise analysis. We chose GradCAM because of its popularity and did not include others because the saliency analysis was not the central focus of our work. 

\section{Statement of Potential Harms}

All AI technology has the potential to cause harm to others and this work is no exception. Our work targets improved robustness and interpretability of deep models, which authors believe may help reduce harm by permitting transparent explanation of model decisions.

\section{Code and Dataset License}

We plan to release our code and data under the MIT license to facilitate open and collaborative research. We have attached a zip file with the code to this submission.

\section{Statement of Offensive Content and Personally Identification Information (PII)}

We declare that our dataset has minimal risk of offensive content. The classes we choose for this dataset (e.g. airplane, car, truck..) are generally of a benign and non-offensive nature.

The images in our dataset were sourced from ImageNet. Therefore our dataset carries the same risks of PII as those in ImageNet, albeit restricted to the classes considered. For instance, although each selected class is not human-related, some images nevertheless contain images of humans. We could not verify that consent of these individuals to have their picture contained in a computer vision database. In future versions of the data, we plan to remove these images with face detectors.

No PII associated to Workers will be released.

%Large latexfigures
\clearpage
\newpage


\clearpage
\newpage

%%%%%%%%% REFERENCES

\clearpage
\newpage

% {\small
% \bibliographystyle{ieee_fullname}
% \bibliography{egbib}
% }

\end{appendix}

%% file: PaperForReview.bbl
\begin{thebibliography}{10}\itemsep=-1pt

\bibitem{counterfactual_gan}
Andreas~Geiger Axel~Sauer.
\newblock Counterfactual generative networks.
\newblock In {\em International Conference on Learning Representations (ICLR)},
  2021.

\bibitem{objectnet}
Andrei Barbu, David Mayo, Julian Alverio, William Luo, Christopher Wang, Dan
  Gutfreund, Joshua~B. Tenenbaum, and Boris Katz.
\newblock Objectnet: A large-scale bias-controlled dataset for pushing the
  limits of object recognition models.
\newblock In {\em NeurIPS}, 2019.

\bibitem{basu_fragile}
Samyadeep Basu, Phillip Pope, and Soheil Feizi.
\newblock Influence functions in deep learning are fragile.
\newblock {\em CoRR}, abs/2006.14651, 2020.

\bibitem{vits_about_same}
Philipp Benz, Soomin Ham, Chaoning Zhang, Adil Karjauv, and In~So Kweon.
\newblock Adversarial robustness comparison of vision transformer and mlp-mixer
  to cnns.
\newblock {\em arXiv preprint arXiv:2110.02797}, 2021.

\bibitem{vits_at_least_as_robust}
Srinadh Bhojanapalli, Ayan Chakrabarti, Daniel Glasner, Daliang Li, Thomas
  Unterthiner, and Andreas Veit.
\newblock Understanding robustness of transformers for image classification.
\newblock {\em CoRR}, abs/2103.14586, 2021.

\bibitem{chang2021towards}
Chun-Hao Chang, George~Alexandru Adam, and Anna Goldenberg.
\newblock Towards robust classification model by counterfactual and invariant
  data generation.
\newblock {\em arXiv preprint arXiv:2106.01127}, 2021.

\bibitem{SimCLR}
Ting Chen, Simon Kornblith, Mohammad Norouzi, and Geoffrey Hinton.
\newblock A simple framework for contrastive learning of visual
  representations, 2020.

\bibitem{Cordts2016Cityscapes}
Marius Cordts, Mohamed Omran, Sebastian Ramos, Timo Rehfeld, Markus Enzweiler,
  Rodrigo Benenson, Uwe Franke, Stefan Roth, and Bernt Schiele.
\newblock The cityscapes dataset for semantic urban scene understanding.
\newblock In {\em Proc. of the IEEE Conference on Computer Vision and Pattern
  Recognition (CVPR)}, 2016.

\bibitem{deng2009imagenet}
Jia Deng, Wei Dong, Richard Socher, Li-Jia Li, Kai Li, and Li Fei-Fei.
\newblock Imagenet: A large-scale hierarchical image database.
\newblock In {\em 2009 IEEE conference on computer vision and pattern
  recognition}, pages 248--255. Ieee, 2009.

\bibitem{vits}
Alexey Dosovitskiy, Lucas Beyer, Alexander Kolesnikov, Dirk Weissenborn,
  Xiaohua Zhai, Thomas Unterthiner, Mostafa Dehghani, Matthias Minderer, Georg
  Heigold, Sylvain Gelly, Jakob Uszkoreit, and Neil Houlsby.
\newblock An image is worth 16x16 words: Transformers for image recognition at
  scale.
\newblock {\em CoRR}, abs/2010.11929, 2020.

\bibitem{robustness}
Logan Engstrom, Andrew Ilyas, Hadi Salman, Shibani Santurkar, and Dimitris
  Tsipras.
\newblock Robustness (python library), 2019.

\bibitem{erion2021}
Gabriel~G. Erion, Joseph~D. Janizek, Pascal Sturmfels, Scott Lundberg, and
  Su{-}In Lee.
\newblock Learning explainable models using attribution priors.
\newblock {\em CoRR}, abs/1906.10670, 2019.

\bibitem{Everingham15}
M. Everingham, S.~M.~A. Eslami, L. Van~Gool, C.~K.~I. Williams, J. Winn, and A.
  Zisserman.
\newblock The pascal visual object classes challenge: A retrospective.
\newblock {\em International Journal of Computer Vision}, 111(1):98--136, Jan.
  2015.

\bibitem{Everingham10}
M. Everingham, L. Van~Gool, C.~K.~I. Williams, J. Winn, and A. Zisserman.
\newblock The pascal visual object classes (voc) challenge.
\newblock {\em International Journal of Computer Vision}, 88(2):303--338, June
  2010.

\bibitem{bolts}
William Falcon and Kyunghyun Cho.
\newblock A framework for contrastive self-supervised learning and designing a
  new approach.
\newblock {\em arXiv preprint arXiv:2009.00104}, 2020.

\bibitem{farhadi2009describing}
Ali Farhadi, Ian Endres, Derek Hoiem, and David Forsyth.
\newblock Describing objects by their attributes.
\newblock In {\em 2009 IEEE Conference on Computer Vision and Pattern
  Recognition}, pages 1778--1785. IEEE, 2009.

\bibitem{shortcuts}
Robert Geirhos, J{\"{o}}rn{-}Henrik Jacobsen, Claudio Michaelis, Richard~S.
  Zemel, Wieland Brendel, Matthias Bethge, and Felix~A. Wichmann.
\newblock Shortcut learning in deep neural networks.
\newblock {\em CoRR}, abs/2004.07780, 2020.

\bibitem{fragile_interpretability}
Amirata Ghorbani, Abubakar Abid, and James Zou.
\newblock Interpretation of neural networks is fragile.
\newblock {\em Proceedings of the AAAI Conference on Artificial Intelligence},
  33(01):3681--3688, Jul. 2019.

\bibitem{adv_mix}
Sven Gowal, Chongli Qin, Po{-}Sen Huang, A.~Taylan Cemgil, Krishnamurthy
  Dvijotham, Timothy~A. Mann, and Pushmeet Kohli.
\newblock Achieving robustness in the wild via adversarial mixing with
  disentangled representations.
\newblock {\em CoRR}, abs/1912.03192, 2019.

\bibitem{resnets}
Kaiming He, Xiangyu Zhang, Shaoqing Ren, and Jian Sun.
\newblock Deep residual learning for image recognition.
\newblock In {\em Proceedings of the IEEE Conference on Computer Vision and
  Pattern Recognition (CVPR)}, June 2016.

\bibitem{hendrycks2019robustness}
Dan Hendrycks and Thomas Dietterich.
\newblock Benchmarking neural network robustness to common corruptions and
  perturbations.
\newblock {\em Proceedings of the International Conference on Learning
  Representations}, 2019.

\bibitem{hendrycks2021nae}
Dan Hendrycks, Kevin Zhao, Steven Basart, Jacob Steinhardt, and Dawn Song.
\newblock Natural adversarial examples.
\newblock {\em CVPR}, 2021.

\bibitem{huang2017densely}
Gao Huang, Zhuang Liu, Laurens Van Der~Maaten, and Kilian~Q Weinberger.
\newblock Densely connected convolutional networks.
\newblock In {\em Proceedings of the IEEE conference on computer vision and
  pattern recognition}, pages 4700--4708, 2017.

\bibitem{fashionpedia}
Menglin Jia, Mengyun Shi, Mikhail Sirotenko, Yin Cui, Claire Cardie, Bharath
  Hariharan, Hartwig Adam, and Serge~J. Belongie.
\newblock Fashionpedia: Ontology, segmentation, and an attribute localization
  dataset.
\newblock {\em CoRR}, abs/2004.12276, 2020.

\bibitem{fereshte}
Fereshte Khani and Percy Liang.
\newblock Removing spurious features can hurt accuracy and affect groups
  disproportionately.
\newblock {\em CoRR}, abs/2012.04104, 2020.

\bibitem{kim2018interpretability}
Been Kim, Martin Wattenberg, Justin Gilmer, Carrie Cai, James Wexler, Fernanda
  Viegas, et~al.
\newblock Interpretability beyond feature attribution: Quantitative testing
  with concept activation vectors (tcav).
\newblock In {\em International conference on machine learning}, pages
  2668--2677. PMLR, 2018.

\bibitem{koh2020understanding}
Pang~Wei Koh and Percy Liang.
\newblock Understanding black-box predictions via influence functions, 2020.

\bibitem{cifar10}
Alex Krizhevsky, Vinod Nair, and Geoffrey Hinton.
\newblock Cifar-10 (canadian institute for advanced research).

\bibitem{lampert2009learning}
Christoph~H Lampert, Hannes Nickisch, and Stefan Harmeling.
\newblock Learning to detect unseen object classes by between-class attribute
  transfer.
\newblock In {\em 2009 IEEE Conference on Computer Vision and Pattern
  Recognition}, pages 951--958. IEEE, 2009.

\bibitem{coco}
Tsung-Yi Lin, Michael Maire, Serge Belongie, Lubomir Bourdev, Ross Girshick,
  James Hays, Pietro Perona, Deva Ramanan, C.~Lawrence Zitnick, and Piotr
  Dollár.
\newblock Microsoft coco: Common objects in context, 2014.
\newblock cite arxiv:1405.0312Comment: 1) updated annotation pipeline
  description and figures; 2) added new section describing datasets splits; 3)
  updated author list.

\bibitem{liu2015faceattributes}
Ziwei Liu, Ping Luo, Xiaogang Wang, and Xiaoou Tang.
\newblock Deep learning face attributes in the wild.
\newblock In {\em Proceedings of International Conference on Computer Vision
  (ICCV)}, December 2015.

\bibitem{pgd}
Aleksander Madry, Aleksandar Makelov, Ludwig Schmidt, Dimitris Tsipras, and
  Adrian Vladu.
\newblock Towards deep learning models resistant to adversarial attacks.
\newblock In {\em 6th International Conference on Learning Representations,
  {ICLR} 2018, Vancouver, BC, Canada, April 30 - May 3, 2018, Conference Track
  Proceedings}. OpenReview.net, 2018.

\bibitem{hci-amt}
Tanushree Mitra, C.J. Hutto, and Eric Gilbert.
\newblock Comparing person- and process-centric strategies for obtaining
  quality data on amazon mechanical turk.
\newblock In {\em Proceedings of the 33rd Annual ACM Conference on Human
  Factors in Computing Systems}, CHI '15, page 1345–1354, New York, NY, USA,
  2015. Association for Computing Machinery.

\bibitem{ftr_viz}
Anh~Mai Nguyen, Jason Yosinski, and Jeff Clune.
\newblock Multifaceted feature visualization: Uncovering the different types of
  features learned by each neuron in deep neural networks.
\newblock {\em CoRR}, abs/1602.03616, 2016.

\bibitem{olah2017feature}
Chris Olah, Alexander Mordvintsev, and Ludwig Schubert.
\newblock Feature visualization.
\newblock {\em Distill}, 2017.
\newblock https://distill.pub/2017/feature-visualization.

\bibitem{olah2018_buildingblocks}
Chris Olah, Arvind Satyanarayan, Ian Johnson, Shan Carter, Ludwig Schubert,
  Katherine Ye, and Alexander Mordvintsev.
\newblock The building blocks of interpretability.
\newblock {\em Distill}, 2018.
\newblock https://distill.pub/2018/building-blocks.

\bibitem{Ouyang_2015_ICCV}
Wanli Ouyang, Hongyang Li, Xingyu Zeng, and Xiaogang Wang.
\newblock Learning deep representation with large-scale attributes.
\newblock In {\em Proceedings of the IEEE International Conference on Computer
  Vision (ICCV)}, December 2015.

\bibitem{NEURIPS2019_9015}
Adam Paszke, Sam Gross, Francisco Massa, Adam Lerer, James Bradbury, Gregory
  Chanan, Trevor Killeen, Zeming Lin, Natalia Gimelshein, Luca Antiga, Alban
  Desmaison, Andreas Kopf, Edward Yang, Zachary DeVito, Martin Raison, Alykhan
  Tejani, Sasank Chilamkurthy, Benoit Steiner, Lu Fang, Junjie Bai, and Soumith
  Chintala.
\newblock Pytorch: An imperative style, high-performance deep learning library.
\newblock In H. Wallach, H. Larochelle, A. Beygelzimer, F. d\textquotesingle
  Alch\'{e}-Buc, E. Fox, and R. Garnett, editors, {\em Advances in Neural
  Information Processing Systems 32}, pages 8024--8035. Curran Associates,
  Inc., 2019.

\bibitem{vits_robust_attn_is_key}
Sayak Paul and Pin-Yu Chen.
\newblock Vision transformers are robust learners, 2021.

\bibitem{pham2021learning}
Khoi Pham, Kushal Kafle, Zhe Lin, Zhihong Ding, Scott Cohen, Quan Tran, and
  Abhinav Shrivastava.
\newblock Learning to predict visual attributes in the wild.
\newblock In {\em Proceedings of the IEEE/CVF Conference on Computer Vision and
  Pattern Recognition}, pages 13018--13028, 2021.

\bibitem{clip}
Alec Radford, Jong~Wook Kim, Chris Hallacy, Aditya Ramesh, Gabriel Goh,
  Sandhini Agarwal, Girish Sastry, Amanda Askell, Pamela Mishkin, Jack Clark,
  Gretchen Krueger, and Ilya Sutskever.
\newblock Learning transferable visual models from natural language
  supervision.
\newblock {\em CoRR}, abs/2103.00020, 2021.

\bibitem{lime}
Marco~T{\'{u}}lio Ribeiro, Sameer Singh, and Carlos Guestrin.
\newblock "why should {I} trust you?": Explaining the predictions of any
  classifier.
\newblock {\em CoRR}, abs/1602.04938, 2016.

\bibitem{russakovsky2010attribute}
Olga Russakovsky and Li Fei-Fei.
\newblock Attribute learning in large-scale datasets.
\newblock In {\em European Conference on Computer Vision}, pages 1--14.
  Springer, 2010.

\bibitem{waterbirds}
Shiori Sagawa, Pang~Wei Koh, Tatsunori~B. Hashimoto, and Percy Liang.
\newblock Distributionally robust neural networks for group shifts: On the
  importance of regularization for worst-case generalization.
\newblock {\em CoRR}, abs/1911.08731, 2019.

\bibitem{gradcam}
Ramprasaath~R. Selvaraju, Abhishek Das, Ramakrishna Vedantam, Michael Cogswell,
  Devi Parikh, and Dhruv Batra.
\newblock Grad-cam: Why did you say that? visual explanations from deep
  networks via gradient-based localization.
\newblock {\em CoRR}, abs/1610.02391, 2016.

\bibitem{vits_maybe_more_robust}
Rulin Shao, Zhouxing Shi, Jinfeng Yi, Pin{-}Yu Chen, and Cho{-}Jui Hsieh.
\newblock On the adversarial robustness of visual transformers.
\newblock {\em CoRR}, abs/2103.15670, 2021.

\bibitem{simonyan2014very}
Karen Simonyan and Andrew Zisserman.
\newblock Very deep convolutional networks for large-scale image recognition.
\newblock {\em arXiv preprint arXiv:1409.1556}, 2014.

\bibitem{singla2021salient}
Sahil Singla and Soheil Feizi.
\newblock Salient imagenet: How to discover spurious features in deep
  learning?, 2021.

\bibitem{barlow}
Sahil Singla, Besmira Nushi, Shital Shah, Ece Kamar, and Eric Horvitz.
\newblock Understanding failures of deep networks via robust feature
  extraction.
\newblock In {\em The IEEE Conference on Computer Vision and Pattern
  Recognition (CVPR)}, 2021.

\bibitem{smoothgrad}
Daniel Smilkov, Nikhil Thorat, Been Kim, Fernanda~B. Vi{\'{e}}gas, and Martin
  Wattenberg.
\newblock Smoothgrad: removing noise by adding noise.
\newblock {\em CoRR}, abs/1706.03825, 2017.

\bibitem{integrated-grads}
Mukund Sundararajan, Ankur Taly, and Qiqi Yan.
\newblock Axiomatic attribution for deep networks.
\newblock In Doina Precup and Yee~Whye Teh, editors, {\em Proceedings of the
  34th International Conference on Machine Learning}, volume~70 of {\em
  Proceedings of Machine Learning Research}, pages 3319--3328. PMLR, 06--11 Aug
  2017.

\bibitem{rcls}
Kamil Szyc, Tomasz Walkowiak, and Henryk Maciejewski.
\newblock Checking robustness of representations learned by deep neural
  networks.
\newblock In Yuxiao Dong, Nicolas Kourtellis, Barbara Hammer, and Jose~A.
  Lozano, editors, {\em Machine Learning and Knowledge Discovery in Databases.
  Applied Data Science Track}, pages 399--414, Cham, 2021. Springer
  International Publishing.

\bibitem{tan2019efficientnet}
Mingxing Tan and Quoc Le.
\newblock Efficientnet: Rethinking model scaling for convolutional neural
  networks.
\newblock In {\em International Conference on Machine Learning}, pages
  6105--6114. PMLR, 2019.

\bibitem{deit}
Hugo Touvron, Matthieu Cord, Matthijs Douze, Francisco Massa, Alexandre
  Sablayrolles, and Herv{\'{e}} J{\'{e}}gou.
\newblock Training data-efficient image transformers {\&} distillation through
  attention.
\newblock {\em CoRR}, abs/2012.12877, 2020.

\bibitem{counterfactual_theory}
Victor Veitch, Alexander D'Amour, Steve Yadlowsky, and Jacob Eisenstein.
\newblock Counterfactual invariance to spurious correlations: Why and how to
  pass stress tests.
\newblock {\em CoRR}, abs/2106.00545, 2021.

\bibitem{nlp_spur}
Tianlu Wang, Diyi Yang, and Xuezhi Wang.
\newblock Identifying and mitigating spurious correlations for improving
  robustness in nlp models, 2021.

\bibitem{WelinderEtal2010}
P. Welinder, S. Branson, T. Mita, C. Wah, F. Schroff, S. Belongie, and P.
  Perona.
\newblock {Caltech-UCSD Birds 200}.
\newblock Technical Report CNS-TR-2010-001, California Institute of Technology,
  2010.

\bibitem{timm_library}
Ross Wightman.
\newblock Pytorch image models.
\newblock \url{https://github.com/rwightman/pytorch-image-models}, 2019.

\bibitem{debuggable}
Eric Wong, Shibani Santurkar, and Aleksander Madry.
\newblock Leveraging sparse linear layers for debuggable deep networks.
\newblock In Marina Meila and Tong Zhang, editors, {\em Proceedings of the 38th
  International Conference on Machine Learning}, volume 139 of {\em Proceedings
  of Machine Learning Research}, pages 11205--11216. PMLR, 18--24 Jul 2021.

\bibitem{madry_noise_or_signal}
Kai~Yuanqing Xiao, Logan Engstrom, Andrew Ilyas, and Aleksander Madry.
\newblock Noise or signal: The role of image backgrounds in object recognition.
\newblock {\em CoRR}, abs/2006.09994, 2020.

\bibitem{zhai2021scaling}
Xiaohua Zhai, Alexander Kolesnikov, Neil Houlsby, and Lucas Beyer.
\newblock Scaling vision transformers.
\newblock {\em arXiv preprint arXiv:2106.04560}, 2021.

\bibitem{cam}
Bolei Zhou, Aditya Khosla, {\`{A}}gata Lapedriza, Aude Oliva, and Antonio
  Torralba.
\newblock Learning deep features for discriminative localization.
\newblock {\em CoRR}, abs/1512.04150, 2015.

\bibitem{zhou-2021}
Yilun Zhou, Serena Booth, Marco~T{\'{\i}}lio Ribeiro, and Julie Shah.
\newblock Do feature attribution methods correctly attribute features?
\newblock {\em CoRR}, abs/2104.14403, 2021.

\end{thebibliography}
